\documentclass{article}
\usepackage{microtype}
\usepackage{graphicx}
\usepackage{pifont}
\usepackage{booktabs} % for professional tables
\usepackage{import}
\usepackage{makecell}
\PassOptionsToPackage{dvipsnames}{xcolor}
\usepackage{natbib}

\usepackage{amsmath}
\usepackage{amssymb}
\usepackage{mathtools}
\usepackage{amsthm}

\usepackage{amsmath,amsfonts,bm}

\def\eqref#1{equation~\ref{#1}}
\def\1{\bm{1}}

\def\rw{{\textnormal{w}}}

\def\mP{{\bm{P}}}

\def\mX{{\bm{X}}}

\DeclareMathAlphabet{\mathsfit}{\encodingdefault}{\sfdefault}{m}{sl}
\SetMathAlphabet{\mathsfit}{bold}{\encodingdefault}{\sfdefault}{bx}{n}

\def\gC{{\mathcal{C}}}

\def\gG{{\mathcal{G}}}
\def\gH{{\mathcal{H}}}

\def\gX{{\mathcal{X}}}

\newcommand{\E}{\mathbb{E}}

\newcommand{\R}{\mathbb{R}}

\newcommand{\bA}{\mathbf{A}}
\newcommand{\bL}{\mathbf{L}}

\newcommand{\cV}{\mathcal{V}}
\newcommand{\cE}{\mathcal{E}}

\theoremstyle{plain}
\newtheorem{theorem}{Theorem}[section]
\newtheorem{proposition}[theorem]{Proposition}
\newtheorem{lemma}[theorem]{Lemma}
\newtheorem{corollary}[theorem]{Corollary}
\theoremstyle{definition}
\newtheorem{definition}[theorem]{Definition}

\theoremstyle{remark}
\newtheorem{remark}[theorem]{Remark}

\usepackage[textsize=tiny]{todonotes}

\newcommand{\V}{\mathcal{V}} % vertex set
\renewcommand{\E}{\mathcal{E}} % edge set

\newcommand{\RR}{\mathbb{R}}
\renewcommand{\rw}{\mathbf{M}}

\renewcommand{\P}{\mathbf{P}}

\usepackage{appendix}

\usepackage{booktabs}

\usepackage{multirow, makecell}
\usepackage[dvipsnames]{xcolor}
\definecolor{oxfordblue}{rgb}{0.0, 0.1, 0.5}
\usepackage[colorlinks=true, linkcolor=oxfordblue, urlcolor=oxfordblue, citecolor=oxfordblue]{hyperref}
\usepackage{url}
\usepackage{soul}
\usepackage{tikz}
\usepackage{wrapfig,lipsum,booktabs}

\usepackage{titletoc}
\renewcommand{\cite}[1]{\citep{#1}}
\usepackage{caption}
\usepackage{subcaption}
\usepackage{enumitem}
\usepackage{authblk}

\title{Revealing Decurve Flows for Generalized Graph Propagation}
\author[1]{Chen Lin \thanks{These authors contributed equally to this work.}}
\author[2]{Liheng Ma \textsuperscript{*}}
\author[1]{Yiyang Chen \textsuperscript{*}}
\author[3]{Wanli Ouyang}
\author[1]{Michael M. Bronstein}
\author[1]{Philip H.S. Torr}
\affil[1]{University of Oxford}
\affil[2]{McGill University}
\affil[3]{University of Sydney}
\begin{document}
\maketitle

\begin{abstract}
This study addresses the limitations of the traditional analysis of message-passing, central to graph learning, by defining {\em \textbf{generalized propagation}} with directed and weighted graphs. The significance manifest in two ways.
\textbf{Firstly}, we propose {\em Generalized Propagation Neural Networks} (\textbf{GPNNs}), a framework that unifies most propagation-based graph neural networks. By generating directed-weighted propagation graphs with adjacency function and connectivity function, GPNNs offer enhanced insights into attention mechanisms across various graph models. We delve into the trade-offs within the design space with empirical experiments and emphasize the crucial role of the adjacency function for model expressivity via theoretical analysis.
\textbf{Secondly}, we propose the {\em Continuous Unified Ricci Curvature} (\textbf{CURC}), an extension of celebrated {\em Ollivier-Ricci Curvature} for directed and weighted graphs. Theoretically, we demonstrate that CURC possesses continuity, scale invariance, and a lower bound connection with the Dirichlet isoperimetric constant validating bottleneck analysis for GPNNs.
We include a preliminary exploration of learned propagation patterns in datasets, a first in the field. We observe an intriguing ``{\em \textbf{decurve flow}}'' - a curvature reduction during training for models with learnable propagation, revealing the evolution of propagation over time and a deeper connection to over-smoothing and bottleneck trade-off.

\end{abstract}

\section{Introduction}
\label{sec:intro}
Propagation of vertex features is central to various graph models, including message-passing models (MPNNs) and graph transformers (GTs). 

MPNNs pass features by the graph structure of incoming data, inherently capturing characteristics of the underlying graph structure ~\cite{hamilton2017InductiveRepresentationLearning, velickovic2018GraphAttentionNetworks, monti2017GeometricDeepLearning}. Propagation of MPNNs is carefully inspected by analyzing its expressivity and geometric properties based on undirected and unweighted graphs. 
%Despite their empirical achievements on various datasets, propagation guided by adjacency directly brought limitations in two aspects. 
In detail, the expressive power of message-passing is bounded by the first-order Weisfeiler-Lehman test (1-WL)\cite{xu2019HowPowerfulAre,chen2019EquivalenceGraphIsomorphism}. Furthermore, MPNNs suffer from over-squashing \cite{alon2020bottleneck} and over-smoothing \cite{chen2020measuring, oono2019graph} phenomena due to the undesirable geometric properties of the graph structures (eg. molecule, knowledge graph) like bottlenecks and complete graphs. 

On the other hand, graph transformers determine the propagation of features with ``attention'' ~\cite{vaswani2023attention}.
Pioneering methods directly apply self-attention on graphs, suffering a lack of structural information~\cite{dwivedi2021GeneralizationTransformerNetworks} and can be potentially improved by injecting additional positional/structural encodings ~\cite{kreuzer2021RethinkingGraphTransformers, rampasek2022RecipeGeneralPowerful, chen2022StructureAwareTransformerGraph,ma2023graph}. Although GTs have shown superior performance on various benchmarks and diverse heuristics, a clear understanding of how they gain their power remains elusive \cite{cai2023ConnectionMPNNGraph}. Specifically, GTs are considered propagating messages with a undirected-unweighted complete graph ~\cite{shirzad2023exphormer}. However, a random walk on complete graph would soon converges to stable distribution which implies severe over-smoothing ~\cite{giraldo2022understanding}, contradicting the empirical success of GTs.  

In this study, we demonstrate that the limitation of existing analysis on message propagation largely stems from the traditional focus on defining propagation within the realm of undirected and unweighted graphs. Challenging conventional boundaries, we conceptualize the \textbf{generalize propagation} with directed and weighted graphs. The significance of this generalization manifest in two ways. 

% \wl{include better modeling of attention}
Firstly, we introduce  {\em Generalized Propagation Networks} (\textbf{GPNNs}), which unify GTs, MPNNs, and their variants by adopting directed, weighted propagation graphs. Central to GPNNs is the {\em adjacency function}, mapping adjacency matrices to vertex pair embeddings, and the {\em connectivity function}, converting these embeddings and vertex features into scalar values. 
This framework advances the analysis of propagation in graph models by its inherent ability to model ``attention'', which is neglected with undirected and unweighted modeling.
For completeness, we show that GPNNs' expressiveness is determined by their adjacency function theoretically.
Furthermore, we provide a taxonomy (see Appendix~\ref{appendix: taxonomy}) for the models that belongs to GPNN family and showcase the high-level design choice in literature. Through empirical evaluations within the design space, we show the impact of adjacency and connectivity function and pinpointing the most effective configurations as well as ineffective ones.

Secondly, we propose the {\em Continuous Unified Ricci Curvature} (\textbf{CURC}), extending the celebrated {\em Ollivier-Ricci (OR) Curvature} to directed and weighted graphs ~\cite{ollivier2009ricci}. CURC is defined by Wasserstein distance between mean transition probabilities centered at each vertex and an asymmetric distance for vertex pairs. During the derivation of CURC, we give a stronger version of \textbf{Kantorovich-Rubinstein duality}, which not only serves as the theoretical basis for our design of \textbf{reciprocal edge weight}, but also enables a wider class of valid distance function to create different graph curvatures based on Riemannian Geometry. Furthermore,  three benign properties of CURC are demonstrated, which are particularly designed for generalized propagation graphs: (i) 
{\em continuity}, which enables smooth variation of the curvature with respect to the propagation graphs; (ii) unity with Ollivier-Ricci Curvature when the input graph is undirected and unweighted, when $(u,v)$ and $(v,u)$ coexist and both $1.0$; and (iii) a lower bound  linking CURC with the Dirichlet Isoperimetric Constant, validating the use of CURC in analyzing bottlenecks on graphs.

We analyze the geometric property of the generalized propagation for models with learnable propagation during training, which has been carried out for undirected-unweighted propagation with OR curvature by ~\cite{topping2022UnderstandingOversquashingBottlenecks}. 
First in the field, we spotted the intriguing ``\textbf{\em decurve flow}'' phenomenon -- a curvature reduction propagation graph evolution observed in multiple models, including the stat-of-the-art, during early stage in training. 
In contrast to rewiring~\cite{topping2022UnderstandingOversquashingBottlenecks}, which increases curvature and reduces bottlenecks, generalized propagation graph of inspected models starts from a distribution of high curvatures and gradually decurves. This suggests a possible explanation of GTs performance: starting from a extreme curved state, GTs solves over-smoothing vs trade-off with respect to data.

\begin{figure}[t]
    \centering
    \begin{minipage}{0.2\textwidth}
        \centering
        \includegraphics[width=\textwidth]
        {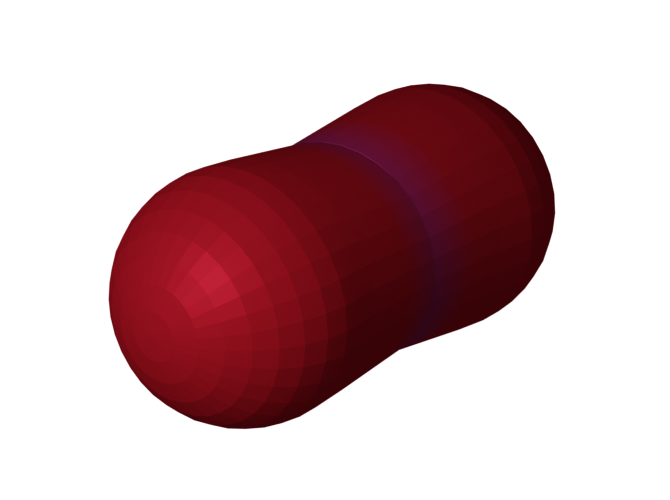}
    \end{minipage}
    \begin{minipage}{0.2\textwidth}
        \includegraphics[width=\textwidth]
        {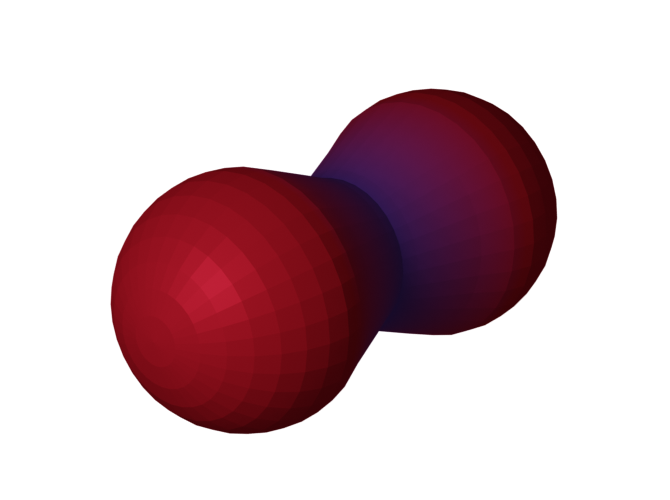}
    \end{minipage}
    \begin{minipage}{0.2\textwidth}
        \includegraphics[width=\textwidth]
        {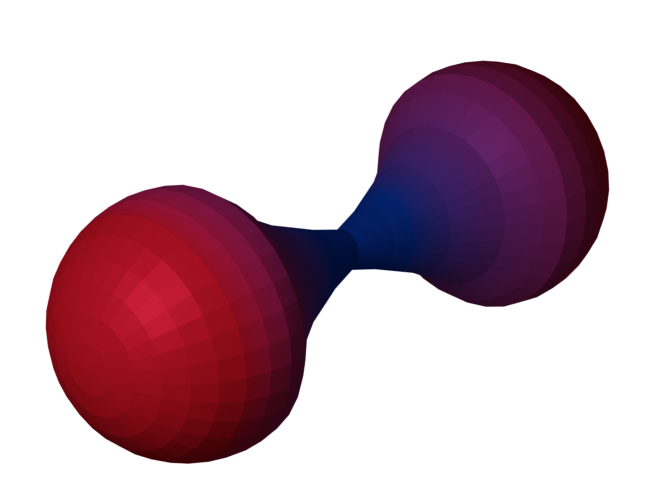}
    \end{minipage}
      \setlength{\belowcaptionskip}{-10pt}
    \caption{Geometric Flow Depicting Curvature Decrease. \\ 
    This visual serves as an analogy for the curvature decrease observed in
    {\em Generalized Propagation Neural Networks} (\textbf{GPNNs}) with learnable propagation. The proposed {\em Continuous Uniform Ricci Curvature} (\textbf{CURC}) are applied to characterize this ``{\em \textbf{decurve flow}}'' observed across various models.
    %While intensifying bottlenecks, the ``decurve flow'' mitigates over-smoothing across various models. 
    Dark-blue and red represent regions of negative and positive curvature, respectively}
    \label{fig:enter-label}
\end{figure}

% \section{Directed and Weighted Propagation}
% \label{sec:preliminary}
% \import{sections/}{preliminary.tex}

\section{Directed and Weighted Propagation}
\label{sec:framework}
This section challenges the convention by defining message propagation on \textbf{directed-weighted} graphs. We show that directed-weighted propagation inherently provides insight to the  ``attention'' mechanism in MPNNs. This analysis is extended towards a general graph model framework - GPNN, characterizing various propagation based methods. We finish this section with a discussion on its \textbf{design space} and an exhibition of GPNNs \textbf{expressivity}.

\paragraph{Preliminary} In this work, we concern machine learning algorithms on undirected graphs  $\mathcal{G} = (\mathcal{V}, \mathcal{E})$ characterized by two sets, namely the vertices $\mathcal{V}=\{1,\hdots, n \}$ and the edges  $\mathcal{E} \subseteq \mathcal{V}\times \mathcal{V}$. Given the indexes of the vertices, the structure of the graph can also be represented by its adjacency matrix $\mathbf{A}$, where $a_{uv} = 1$ iff $(u,v) \in \mathcal{E}$ and zero otherwise. Each vertex and edge could be associated with attributes $x_u$ and $e_{uv}$. 
%We use $\mathbf{X} \in \mathbb{R}^{n \times d}$ and $\mathbf{E} \in \mathbb{R}^{n^2 \times d}$ to represent the collections of those features. 
The intermediate hidden variables passing between layers are denoted as $h$. A graph network model usually contains multiple layers marked as $L$. 

\paragraph{Generalized Propagation} We consider message propagation on weighted-directed graph $\mathcal{G}_p = (\mathcal{V}, \mathcal{E}, \mathcal{\omega})$, where $\omega$ denotes the connectivity: $\forall (u,v) \in \mathcal{E} \rightarrow \omega_{uv}  \in \mathbb{R}$. The connectivity $\omega$ models the amount of information that is passed along with a specific edge (eg. ``attention'').
In fact,  the concept of weighted propagation is common and can be introduced to MPNNs by refining its definition (see Appendix~\ref{appendix: background-MPNN})
\begin{equation} \label{eq:mpnn_refine}
\begin{aligned}
\omega^{l+1}_{vu} & = C_l(h_v^l, h_u^l, e_{vu}) \\
m_v^{l+1} & =\sum_{u \in N(v)} \omega^l_{vu} \cdot M_l\left(h_u^l \right)
\end{aligned}
\end{equation}
where $C_l$ is a scalar function that models the connectivity. This formulation is standard in most MPNNs \cite{kipf2017SemiSupervisedClassificationGraph, monti2017GeometricDeepLearning, velickovic2018GraphAttentionNetworks, bresson2018ResidualGatedGraph}. Previous analyses over message passing are based on the summation over neighbors, overlooking the important affect of connectivity. 

\paragraph{GPNNs} To obtain a general formulation, we make extentions: (1) summation over all vertices. (2) an \textbf{adjacency function} $F$, defined as a permutation equivariant mapping from the adjacency matrix $\mathbf{A}$ and edge feature $\mathbf{E}$ (if available) to its embedding $f_{vu}$. (3) the \textbf{connectivity function} which depend on adjacency feature $f_{vu}$ and along with vertex features. The resulting formulation contains four steps for each layer,
\begin{equation} \label{eq: gpnn}
\begin{aligned}
f^l_{vu} &= [F_l(\mathbf{A}, \mathbf{E})]_{vu} \\
% f^{l+1}_{vu} & =  C_l(h_{v}, h_w,f^{l}_{vu})  \\ % edge enhance?
\omega^{l+1}_{vu} & = C_l( h^l_{v}, h^l_w, f^l_{vu})  \\
m_v^{l+1} & =\sum_{u \in \mathcal{V}} \omega^{l+1}_{vu} \cdot M_l\left(h_u^l\right) \\ %, f^l_{vu}
h_v^{l+1} & =U_l\left(h_v^l, m_v^{l+1}\right)
\end{aligned}    
\end{equation}
where $[\cdot]_{ij}$ indicates the $(i,j)$ element of the input matrix/tensor.

\setlength{\abovecaptionskip}{0pt}
\begin{figure}[t!]
    \centering
    \hspace*{-0.2cm}
    \includegraphics[width=0.6\textwidth]{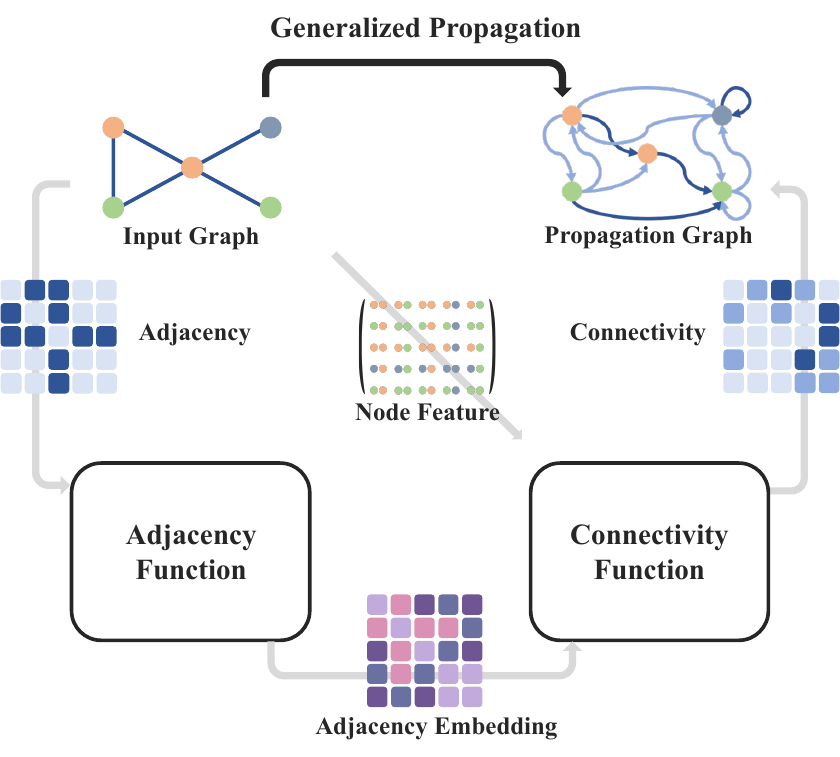}
    \setlength{\belowcaptionskip}{-10pt}
    \caption{The demonstration of GPNN framework}
    \label{fig:GPNN_framework}
\end{figure}

\subsection{GPNN Design Space} \label{sec: gpnn design space}
In Appendix~\ref{appendix: casting} we show existing methods belong to GPNNs by casting their formulation with GPNNs' adjacency and connectivity function.
GPNNs cover a wide range of propagation-based models, making them excellent guidelines for taxonomy (see Appendix~\ref{appendix: taxonomy}). Here we highlight several important design choices.
\vspace{-10pt}
\paragraph{Multi-head v.s. Single-head} pinpointing the question of whether multiple propagation graphs should be used in each GPNN layer. 
When multi-head is adopted, $\omega_{uv}^l$, $m_v^l$ and $h_v^l$ will be specific to each head as $\omega_{uv}^{lh}$, $m_v^{lh}$ and $h_v^{lh}$. 
The multi-head approach is adopted commonly with attention mechanisms.
\vspace{-10pt}
\paragraph{Local v.s. Non-Local v.s. Global Propagation} can be identified based on the constraints imposed by the input graph.
Local propagation constraints the propagation graph to be a weighted version of the input graph, 
which is typical for MPNNs. 
% which is the case for MPNNs.
Global propagation poses no constraint to the propagation graph,
characterizing most GTs.
Non-local propagation, making the propagation graph a super graph of the input graph,
% includes additional but not fully-connected edges 
includes additional edges but does not form a complete graph
to balance cost and expressiveness,
incorporating techniques like graph-rewiring, polynomial spectral GNNs, diffusion-enhanced GNNs, and efficient GTs or MPNNs with virtual nodes. 
\vspace{-10pt}
\paragraph{Static v.s. Dynamic Propagation} \label{sec:static vs dynamic} 
\emph{static} mode denotes that all layers in GPNNs are sharing the same propagation graph. 
In contrast, the \emph{dynamic} mode indicates that different propagation graphs are used in different layers. 
This is typically implemented by evolving the connectivity $\omega_{vu}^l$ across layers. 
\vspace{-10pt}
\paragraph{Feature-independent v.s. Feature-dependent Propagation}
% \wl{meaningless}
The propagation graph can be generated in two ways: depending on the node features, referred to as \textit{feature-dependent (Feat. Dep.)} propagation, 
commonly seen in models with attention or gating mechanisms.
Otherwise, the propagation is referred to as \textit{feature-independent (Feat. Ind.)} propagation.
The adjacency function $F_l$, as the essential component in GPNNs, 
interprets the topological information contained in the adjacency matrix $\mathbf{A}$ for the connectivity function.
The adjacency functions can be as simple as performing normalization on adjacency matrix $\bA$ or the Laplacian matrix $\bL$, as seen in many MPNNs.
More complicated designs, such as computing power series of $\bA$($\bL$) (polynomial spectral/diffusion GNNs~\cite{defferrard2017ConvolutionalNeuralNetworks, gasteiger2019DiffusionImprovesGraph, frasca2020SIGNScalableInception}, GRIT~\cite{ma2023graph})
or performing eigendecomposition on $\bA$($\bL$)~\cite{kreuzer2021RethinkingGraphTransformers, hussain2022GlobalSelfAttentionReplacement}, can explore richer topological information and facilitate connectivity function to generate more expressive propagation graphs.

\subsection{Expressiveness of GPNNs}
\label{sec:expressiveness}
GPNNs subsume models with various expressiveness. 
However, by abstracting away the exact adjacency and mild assumption over connectivity function (MLPs), we deliver two propositions by inducting a common routine for propagation-based models expressivity assessment.
To reach the upper-bounded expressiveness for both propositions, a sufficient number of heads,  MLPs with a sufficient layer and width for the connectivity function, and an update function are required.
First, we consider the GPNNs with static propagation for a sufficient number of heads and layers, referred to as \textit{static GPNNs}. 

\begin{proposition} \textbf{(Static GPNNs)} \label{homogeneous-gpnn-maintex}
For a static GPNN model with a fixed adjacency feature $[f_{uv}]_{u,v\in \cV}$ and sufficient heads and layers, the expressiveness is upper-bounded by the color refinement iteration
\begin{equation}\label{GDWL_to_proof}
    \mathcal{X}^{t+1}_{\gG}(v) = \text{hash} \{\{ (\mathcal{X}^t_{\gG}(u), f_{vu}): u \in \V \}\}
\end{equation}\end{proposition}
The proof the proposition~\ref{homogeneous-gpnn-maintex} is provided in Appendix~\ref{homogeneous-gpnn}. This expressiveness result could be induced by existing expressivity proof in propagation-based GNNs. This proposition assured an expressiveness upper bound for models in GPNN families by solely inspecting their adjacency function.

Generally, we consider the unusual case when multiple adjacency function $F_l$ is applied, rotating among layers, eg. ~\cite{rampasek2022RecipeGeneralPowerful}. 

\begin{proposition} \textbf{(Dynamic GPNNs)}
% Eq. \ref{eq:mpnn_refine}}
Suppose a Layer-recurrent GPNN model M has a layer-dependent adjacency embedding 
repeats every $p$ layer: $f^l_{vu} = f^{l+p}_{vu}, \forall v,u \in \cV$. 
With sufficient heads and layers, by stacking of such repetition, the expressiveness of M is upper-bounded by the color refinement iteration
\begin{equation} \label{combined_coloring}
    \mathcal{X}^{t+1}_{\gG}(v) = \text{hash} \{\{  \left(\mathcal{X}^t_{\gG}(u), \left(f^1_{vu}, f^2_{vu}, \ldots, f^p_{vu} \right) \right): u \in \V \}\}
\end{equation}
\end{proposition}
We refer readers to Appendix~\ref{layer-recurrent-gpnn-proof} for detailed proof. This proposition justified 
those approaches
introducing multiple adjacency functions for different layers:
the overall expressiveness is not weaker than the ones using any one of the adjacency functions alone.
This proposition provides a strong theoretical foundation for the usage of \textit{dynamic propagation} designs in GPNNs.

\section{Continuous Unified Ricci Curvature}
\label{sec:curvature}
In the previous section, we generalize message propagation with directed-weighted graphs and carry out a general framework that characterizes various propagation-based graph models. In this section, we introduce the {\em Continuous Unified Ricci Curvature} (CURC) designed explicitly for analyzing GPNNs' weighted-directed propagation graph. Notably, we propose a stronger version of Kantorovich-Rubinstein duality, which serves as the theoretical basis for our design of reciprocal edge weight and enables a broader class of valid distance functions. Later, we discuss the connection between optimal transport and information propagation on graphs. Finally, we finish by analyzing the properties of CURC, including a lower bound connecting the Dirichlet isoperimetric constant and CURC. 

\newcommand{\bD}{\mathbf{D}}

\subsection{Preliminaries for CURC}
% TODO: 1. align notation, 考虑之前的定义与CURC要用的定义的区别，在这里解释清楚 2. 解释notation为何不通 (convention) 3. Yau et al. 和 ozawa 他们怎么做的，每个概念的联系，我们如何改变 得到什么效果

To facilitate comprehension of readers and maintain the notation consistency with earlier works on curvatures, 
we utilize function notation to describe CURC. 
For example, we interpret $\omega$ as a function $\omega: \V \times \V \mapsto \R^{\geq 0}$ where $\omega(v,u)$ is equivalent to $\omega_{vu}$ in Eq.~\ref{eq:mpnn_refine}. 
To avoid confusion with the notations in previous sections, we use $x, y$ instead of $v, u$ specifically in for CURC.
Note that, for CURC, we only consider the absolute value of $\omega_{vu}$.
We assume all propagation graphs $\gG=(\V, \E, \omega)$ in this section to be strongly-connected weighted-directed finite graphs with $n$ vertices, except otherwise specified. Let the \textbf{random walk matrix} $W \in \R^{n \times n}$ be defined by $[W]_{xy} := d_{x}^{-1} \omega(x, y)$, where $d_x:= \sum_{y \in \V} \omega(x, y)$. In the following discussions, we abuse the notation of $W$ to keep in line with other function-form variables, where we denote $[W]_{xy}$ as $W(x, y)$. We construct CURC based on the curvature proposed by \citet{ozawa2020geometric} and Lin-Liu-Yau Ricci Curvature ~\citet{lin2011ricci}. 

\subsection{Construction of CURC} 
According to the \textit{Perron-Fronbenius Theorem} \cite{Kirkwood2020ThePT}, for a finite strongly-connected weighted-directed graph $\gG = (\V, \E, \omega)$, there exists a strictly positive left eigenvector $\textbf{v}_{pf} \in \R^{1 \times n}$ of the corresponding random walk matrix $W$ (\textit {Perron-Frobenius left eigenvector}). By normalizing $\textbf{v}_{pf}$, we state the following concept of \textbf{Perron measure}, which also corresponds to the stationary distribution for the random walk matrix $W$.

\begin{definition} \textbf{(Perron measure)} \label{perron measure}
%condition, 为啥这里能用
For $\gG = (\V, \E, \omega)$, let $\textbf{v}_{pf} \in \R^{1 \times n}$ be its Perron-Frobenius left eigenvector. The \textit{Perron measure} $\mathfrak{m}: \V \mapsto (0, 1]$ is defined by
    $$
        \mathfrak{m}(x) := (\frac{\textbf{v}_{pf}}{\|\textbf{v}_{pf}\|})_{x}.
    $$\\
\end{definition}
\vspace{-20pt}
\begin{definition} \textbf{(Mean transition probability kernel)} \label{mtpk}
    Let $\mathfrak{m}: \V \mapsto (0, 1]$ be the perron measure defined on graph $\gG = (\V, \E, \omega)$. The \textit{mean transition probability kernel} $\mu: \V \times \V \mapsto [0, 1]$ is defined by
    \begin{equation}\label{mean transition kernel}
        \mu(x, y):= \begin{cases} \frac{1}{2}[W(x, y) + \frac{\mathfrak{m}(y)}{\mathfrak{m}(x)} W(y, x)] & \text{if $y \neq x$} \\ 0 &\text{if $y = x$}\end{cases} \\
    \end{equation}
    
    where $\mu_x(y) := \mu(x, y)$ and $\sum_{y \in \V} \mu_x(y) = 1$ for fixed $x \in \V$.
\end{definition}
Here, we can easily induce that  $\sum_{y \in \V} \frac{\mathfrak{m}(y)}{\mathfrak{m}(x)} W(y, x) = 1$ by checking $\sum_{y \in \V} W(x, y)=1$ and  $\sum_{y \in \V} \frac{\mathfrak{m}(y)}{\mathfrak{m}(x)} W(y, x) = 1$. Intuitively,  $\sum_y \frac{\mathfrak{m}(y)}{\mathfrak{m}(x)} W(y, x)$  contains the information of random walks from $y$ to $x$. 
% 在介绍CURC的定义之前，我们先介绍KR duality, 对于一般的ricci curvature，KR-duality给出了一个使用linear programming 计算的方法。我们给出了一种KR-duality with looser condition on the distance, 指出其实在我们设计像ricci curvature这样的关于flow的curvature的distance function时，我们只需要满足triangle inequality即可，这也是我们设计reciprocal distance function的根据
% 这里的distance function definite是指d(x, y) = 0 iff x = y
We aim to define curvature for weighted-directed graphs with Wasserstein distance by optimal transpor. We will use this mean transition probability kernel to construct initial mass distribution for the optimal transport \cite{ozawa2020geometric}. In order to reduce the optimal transport problem to linear programming, K-R duality is essential. Conventionally, we require a distance function to be non-negative, definite, symmetric and satisfies the triangle inequality.  
In the following, we propose a version of K-R duality that requires a weaker condition on the distance function of graph $\gG$, excluding the symmetry restriction.

\begin{definition}\textbf{(Coupling)} \label{Coupling}
    Suppose $\mu$ and $\nu$ to be two probability distribution on finite sets $\mathcal{X}$ and $\mathcal{Y}$ respectively. Let $\Pi\left(\mu, \nu\right)$ denotes the set of \textit{couplings} between $\mu$ and $\nu$. We say $\pi: \mathcal{X} \times \mathcal{Y} \mapsto [0, 1]$ $\in \Pi\left(\mu, \nu\right)$ is a \textit{coupling} if
    $$
    \sum_{y \in V} \pi(x, y)=\mu(x), \quad \sum_{x \in V} \pi(x, y)=\nu(y).
    $$
\end{definition}

\begin{theorem} \textbf{(K-R duality)}\label{KR duality}
    Let $\gG = (\V, \E, \omega)$ be a graph with (asymmetric) distance function $d: \V \times \V \mapsto \R^{\geq 0}$ satisfying triangle inequality and admits $d(x, y) = 0 \text{ if and only if } x = y$. Then for probability measure $\mu, \nu: \V \mapsto [0, 1]$, the \textbf{Kantorovich-Rubinstein duality} holds. Namely,
    \begin{equation}\label{KR duality equation}
    \inf _{\pi \in \Pi\left(\mu, \nu\right)} \sum_{x, y \in \V} d(x, y) \pi(x, y) = \sup _{f \in \operatorname{Lip}_1(\V)} \sum f(z)\left(\mu -\nu\right),
    \end{equation}
    where $\pi \in \Pi\left(\mu, \nu\right)$ is a coupling between $\mu, \nu$ and $f:\V \mapsto \R\in\operatorname{Lip}_1(\V)$, if for all $x, y \in \V$, $f(y) - f(x) \leq d(x, y)$.
\end{theorem}

Proof of Theorem \ref{KR duality} is given in Appendix \ref{KR duality proof}. Additionally, for computational intensive scenario, we offer two lower-bound estimations of CURC under distinct assumptions in Appendix \ref{lower bound 1} and \ref{lower bound 2}, with computational costs of $\mathcal{O}(n^3)$ and $\mathcal{O}(n^4)$, respectively.
Here, we define the \textbf{reciprocal edge weight} to construct such \textbf{asymmetric distance function}. 

\begin{definition} \textbf{(Reciprocal edge weight)} \label{reciprocal edge weight}
%TODO: Reciprocal edge weight, 记作r
    Let $\varepsilon$ be a small positive real value, then the \textit{$\varepsilon$-masked reciprocal edge weight} $\mathfrak{r}^{\varepsilon}: \V \times \V \mapsto \R^{>0} \cup \{\infty\}$ is defined by 
    $$
    \mathfrak{r}^{\varepsilon}(x, y) := \begin{cases} \frac{1}{\omega(x, y)} & \text{if } \omega(x, y) \geq \varepsilon \\ \frac{1}{\varepsilon} & \text{if } \omega(x, y) < \varepsilon \end{cases}
    $$
%TODO: 结合GPNN的asymmetric distance function, enable our application propagation on asymmetric weight
\end{definition}
\begin{definition} \textbf{(Reciprocal distance function)} \label{reciprocal distance function}
    The \textit{$\varepsilon$-masked reciprocal distance function} $d^{\varepsilon}: \V \times \V \mapsto \R^{> 0} \cup \{\infty\}$ is defined by
    $$
        d^{\varepsilon}:= \text{shortest path with $\mathfrak{r}^{\varepsilon}$ as edge length.}
    $$ %TODO: weighted冗余
\end{definition}
In particular, $d^{\varepsilon}$ is a possibly \textbf{asymmetric distance function} on $\V \times \V$, which is consistent with directed-weighted graph propagation in GPNNs.
% 过度
Now we are ready to drive the definition of CURC. Attention should be placed to handle the singularity issue of reciprocal distance. We use an extra definition wtih $\kappa^\epsilon_{CURC}$ to define $\kappa_{CURC}$ with $\epsilon$ limit.
\begin{definition} \textbf{(CURC)} \label{CURC}
    Let $\gG=(\V, \E, \omega)$ be a graph equipped with the \textit{$\varepsilon$-masked weighted reciprocal distance function} $d^{\varepsilon}$. For distinct $x, y \in \V$, we define the \textit{$\varepsilon$-masked Continuous Unified Ricci Curvature} by
    $$
        \kappa_{\mathrm{CURC}}^{\varepsilon}(x, y):=1-\frac{\mathcal{W}_1^{\varepsilon}\left(\mu_x, \mu_y\right)}{d^{\varepsilon}(x, y)} ,
    $$
    where the Wasserstein distance $\mathcal{W}_1^{\varepsilon}$ is based on $d^{\varepsilon}$. 
    The corresponding \textit{Continuous Unified Ricci Curvature (CURC)} is defined by
    $$
        \kappa_{\mathrm{CURC}}(x, y):=\lim _{\varepsilon \rightarrow 0} \kappa_{\mathrm{CURC}}^{\varepsilon}(x, y).
    $$
\end{definition}
\vspace{-0pt}
\subsection{Optimal Transport of Information}
Here, we share intuition over optimal transport of information. Existing applicaiton of optimal transportation in a message-passing on graphs $G = (\V, \E)$ is investigated by ~\cite{topping2022UnderstandingOversquashingBottlenecks}, utilizing Ollivier-Ricci (OR) curvature. For vertices $x, y$, OR-curvature, represented as $\kappa_{OR}(x, y)$, evaluates the ratio of Wasserstein to graph distance, signifying the cost of moving uniform mass across edges. In this framework, information from each vertex $x$ diminishes as it spreads to neighboring vertices due to the nested aggregation function. Such diminishing impact is akin to the cost in Wasserstein distance. By normalizing this, we assert OR-curvature as a metric to measure the difficulty of information transport between vertices. This motivates us to develop CURC based on optimal transport.
% Information flow in generalized propagation inherently differs from message-passing with undirected-unweighted graph for the following reasons:
% \vspace{-0pt}
% \begin{itemize}[itemsep=0pt, parsep=0pt]
%     \item The distribution of information emanating from each vertex is uneven.
%     \item Traversing different edges causes different extent of diminishing effect on information flow.
%     \item Weighted-directed graphs exhibit distinct geometric and spectral properties when compared to unweighted-undirected graphs.
% \end{itemize}
% \vspace{-0pt}
% To address these distinctions and accurately model information propagation on weighted-directed graphs, we introduce the \textit{Continuous Unified Ricci Curvature} as an extension to the Ollivier-ricci curvature, specifically designed for our strongly-connected weighted-directed propagation graphs. 
% }

\subsection{{Properties of CURC}}

\begin{proposition}\label{properties of CURC}
The \textit{Continuous Unified Ricci Curvature} $\kappa_\mathrm{CURC}$ admits the following properties: 
    \vspace{-0pt}
    \begin{itemize}[itemsep=0pt, parsep=0pt]
        \item (Unity) For connected unweighted-undirected graph $\gG=(\V, \E)$, for any vertex pair $x, y \in \V$, we have $\kappa_\mathrm{CURC}(x, y) = \kappa_{OR}(x, y)$, where $\kappa_{OR}$ is the Ollivier-ricci curvature.
        \item (Continuity) If we perceive $\kappa_{\mathrm{CURC}}(x, y)$ as a function of $\omega$, then $\kappa_{\mathrm{CURC}}(x, y)$ is continuous w.r.t. $\omega$ entry-wise.
        \item (Scale invariance) For strongly-connected weighted-directed graph $\gG=(\V, \E, \omega)$, when all edge weights $\omega$ are scaled by an arbitrary positive constant $\lambda$, the value of  $\kappa_\mathrm{CURC}(x, y)$ for any vertex pair $x, y \in \V$ is invariant. 
    \end{itemize}
\end{proposition}
\vspace{-0pt}

Detailed proof Proposition \ref{properties of CURC} is given in Appendix~\ref{Curvature property}. The Unity and Continuity properties collectively establish CURC as a continuous extension of the canonical Ollivier-Ricci curvature, as originally introduced in \citet{ollivier2009ricci}. Scale invariance is a crucial property for CURC being a measure of bottlenecks on the propagation graph. Intuitively, uniformly scaling of edge weight does not change the dynamics of information on the propagation graph, leading to a robust bottlenecks or oversquashing measure on the graph.

Furthermore, we establish how CURC can be perceived as a measure of bottlenecks phenomenon with the help of the so-called \textit{Dirichlet isoperimetric constant}, which is the direct extension of \textit{Cheeger constant} into strongly-connected weighted-directed graphs. \textit{Cheeger constant} has been widely used to measure bottlenecks of GNN messsage-passing on undirected-unweighted graphs due to its connection with community clustering. In \citet{topping2022UnderstandingOversquashingBottlenecks}, they established the connection between Balanced Forman curvature with Cheeger constant using spectral properties. Likewise, we derive similar property between CURC and Dirichlet isoperimetric constant to conclude our theoretical discussion of CURC and its connection to bottlenecks.

\begin{definition} \textbf{(Boundary Perron-measure)} \label{Boundary Perron-measure}
    For a non-empty $\Omega \subset \V$, its \textit{Boundary Perron-measure} is defined as
    $$\mathfrak{m}(\partial \Omega):=\sum_{y \in \Omega} \sum_{z \in V \backslash \Omega} \mathfrak{m}_{y z},$$
    where $\mathfrak{m}_{y z}:= \mathfrak{m}(y) \mu(y, z)$ and $\mathfrak{m}(\Omega) = \sum_{x \in \Omega} \mathfrak{m}(x)$.
\end{definition}

\begin{definition} \textbf{(Dirichlet isoperimetric constant)} \label{DIC}
    The \textit{Dirichlet isoperimetric constant} $\mathcal{I}_{\mathcal{V}}^D$ on a non-empty set $\V$ is defined by
    $$\mathcal{I}_{\mathcal{V}}^D:=\inf _{\Omega \subset \V} \frac{\mathfrak{m}(\partial \Omega)}{\mathfrak{m}(\Omega)}.$$
\end{definition}

%TODO: 讲清楚Dirichlet isoperimetric constant是cheeger constant的direct extension （在weighted directed grpah中。With definition 4.6, 我们可以引出一个关于CURC在weighted directed graph上关于DIC的性质。

\begin{theorem}\label{weighted cheeger constant}
Let $\gG = (\V, \E, \omega)$ and $E_R(x):=\left\{y \in \V \mid d(x, y) \geq R\right\}$. Fix $x \in \V$, we assume $\inf_{y \in \V \backslash \{x\}} \kappa_{\mathrm{CURC}}(x, y) \geq K$ for some $K \in \R$ and $-\sum_{y \in \V} \mu(x, y) d(x, y) \geq \Lambda$ for some $\Lambda \in (-\infty, 0)$. For $D > 0$, we further assume that for all $y \in \V$, $d(x, y)\leq D$. Then the \textit{Dirichlet isoperimetric constant}  admits the following lower bound:
$$
\mathcal{I}_{E_R(x)}^D \geq \frac{K R+\Lambda}{D}.
$$
\end{theorem}

Proof of Theorem \ref{weighted cheeger constant} is given in Appendix \ref{weighted cheeger constant proof}. Same as its counterpart in undirected-unweighted graphs, a small Dirichlet isopermetric constant indicates higher probability of bottlenecks on propagation graph. For a fixed vertex $x$, a large value of $\kappa_{\mathrm{CURC}}(x, y)$ for $y \in \V$ implie a large value of $K$, which results in a larger lower bound for the Dirichlet ioperimetric constant $\mathcal{I}_{E_R(x)}^D$ according to Theorem \ref{weighted cheeger constant}. Combined with Proposition \ref{properties of CURC}, CURC constitues a proper instrument for measuring the bottleneck phenomenon on propagation graphs within our GPNN framework. For better visualization, we present the CURC distributions of our GPNN models in the later experiments.

\begin{figure}[ht!]
    \centering
    \begin{minipage}{0.2\textwidth}
        \centering
        \includegraphics[width=\textwidth]{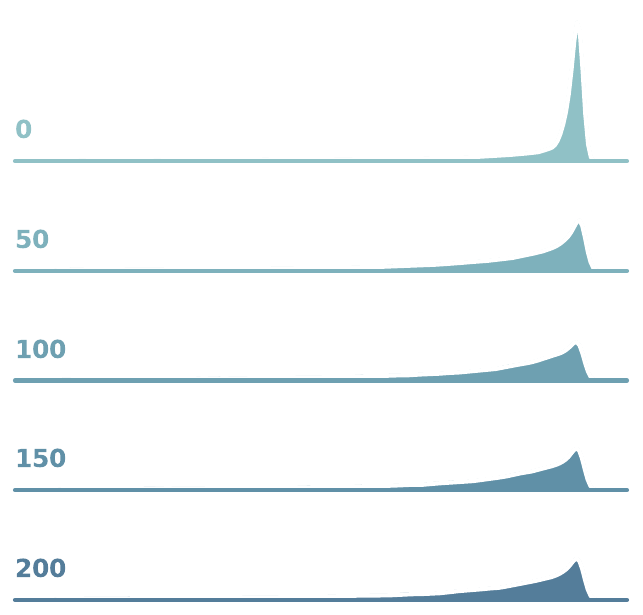}
        {\scriptsize (a) SAN}
    \end{minipage}%
    \begin{minipage}{0.2\textwidth}
       \centering 
        \includegraphics[width=\textwidth]{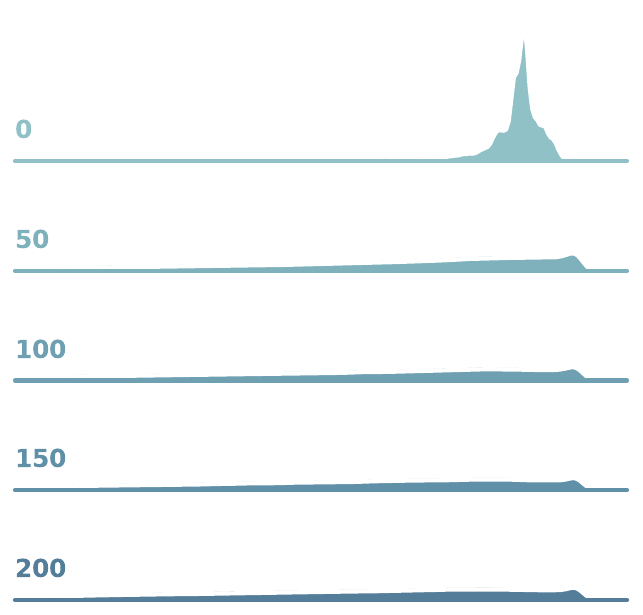}
        {\scriptsize (b) Graphormer}
        \end{minipage}
    \begin{minipage}{0.2\textwidth}
        \centering
        \includegraphics[width=\textwidth]{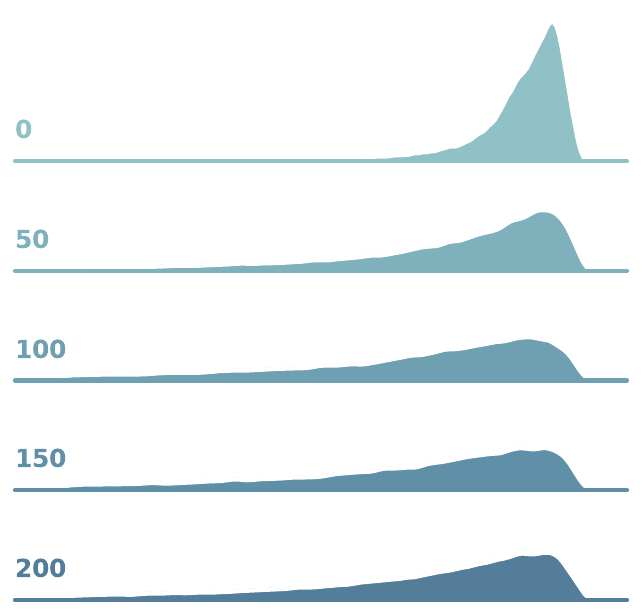}
        {\scriptsize (c) GRIT}
    \end{minipage}%
    \begin{minipage}{0.2\textwidth}
       \centering 
        \includegraphics[width=\textwidth]{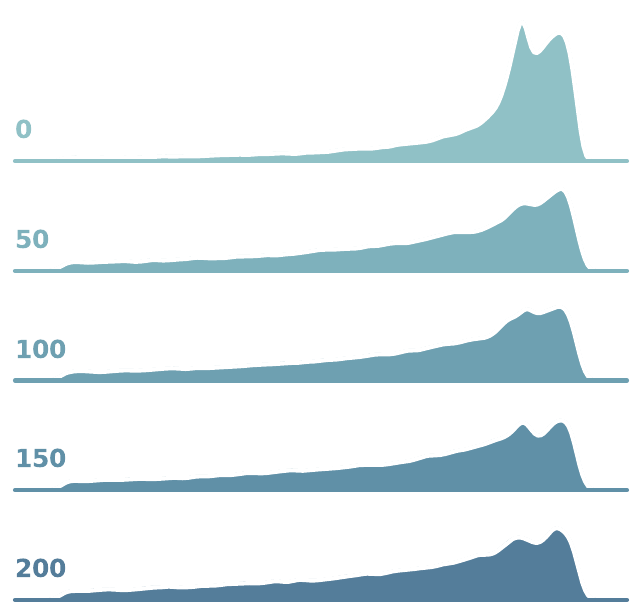}
        {\scriptsize (d) GPNN-PE}
    \end{minipage}
     \vskip 0.1in
    \caption{Visualization of the curvature distribution characterized on ZINC for 4 models with dynamic propagation learned from data. The CURC distribution of 0 to 200 epochs are placed vertically for each model. A tendency of curvature shifting to their left side is observed across all models. This ubiquitous {\em \textbf{decurve flow}} phenomenon suggests a change of bottlenecks by the directed-unweighted propagation graphs analysis central to this work.}
    \label{fig:trend_curc}
\end{figure}

\begin{figure*}[h!]
    \centering
    % Row 1
    \begin{minipage}[b]{0.33\textwidth}
        \centering
        \includegraphics[width=\textwidth]{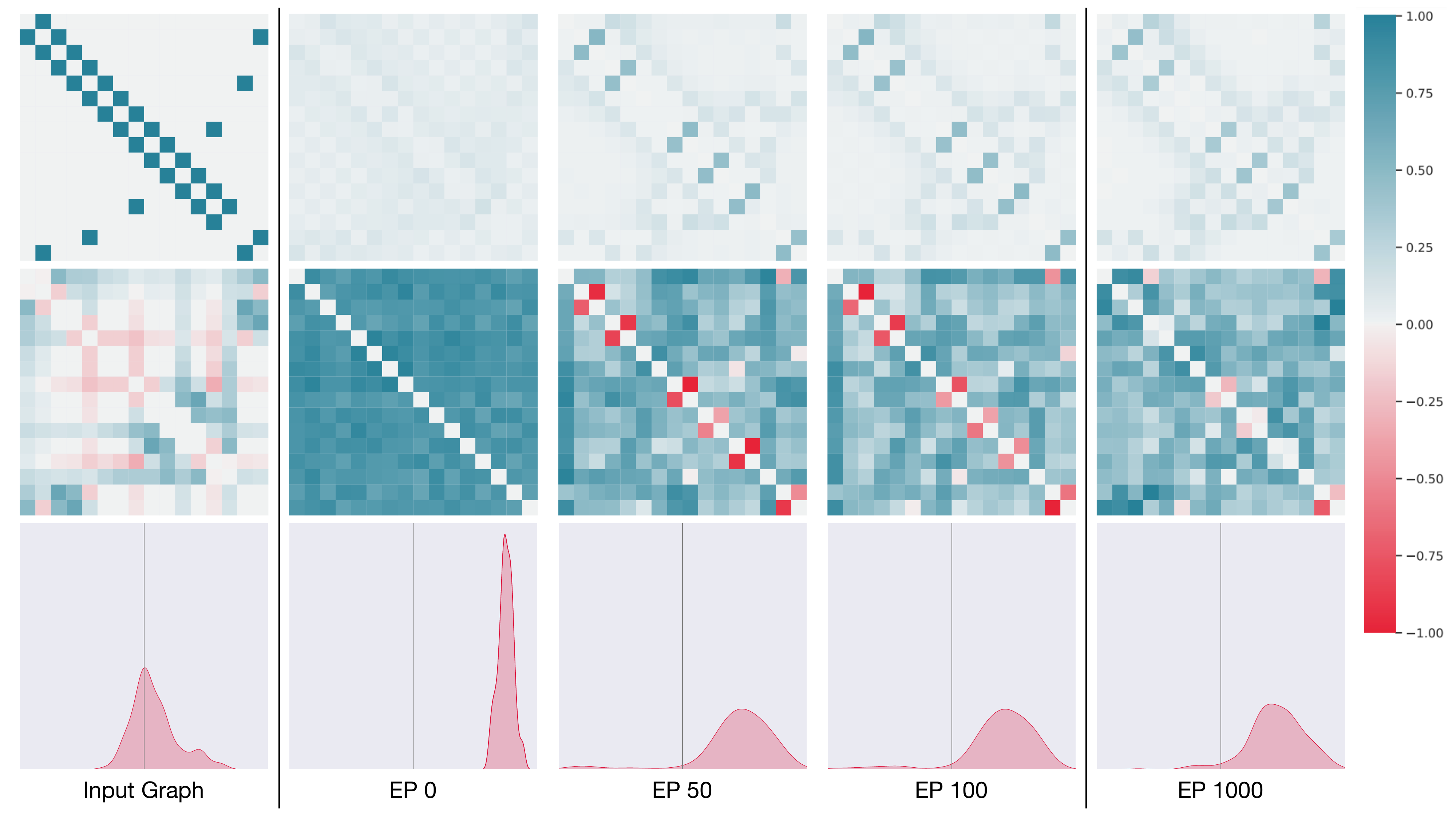}
        % \caption{Visualization of Attention and CURC for Graph-1 in ZINC}
        \label{fig:curc-zinc-1}
    \end{minipage}%
    \begin{minipage}[b]{0.33\textwidth}
        \centering
        \includegraphics[width=\textwidth]{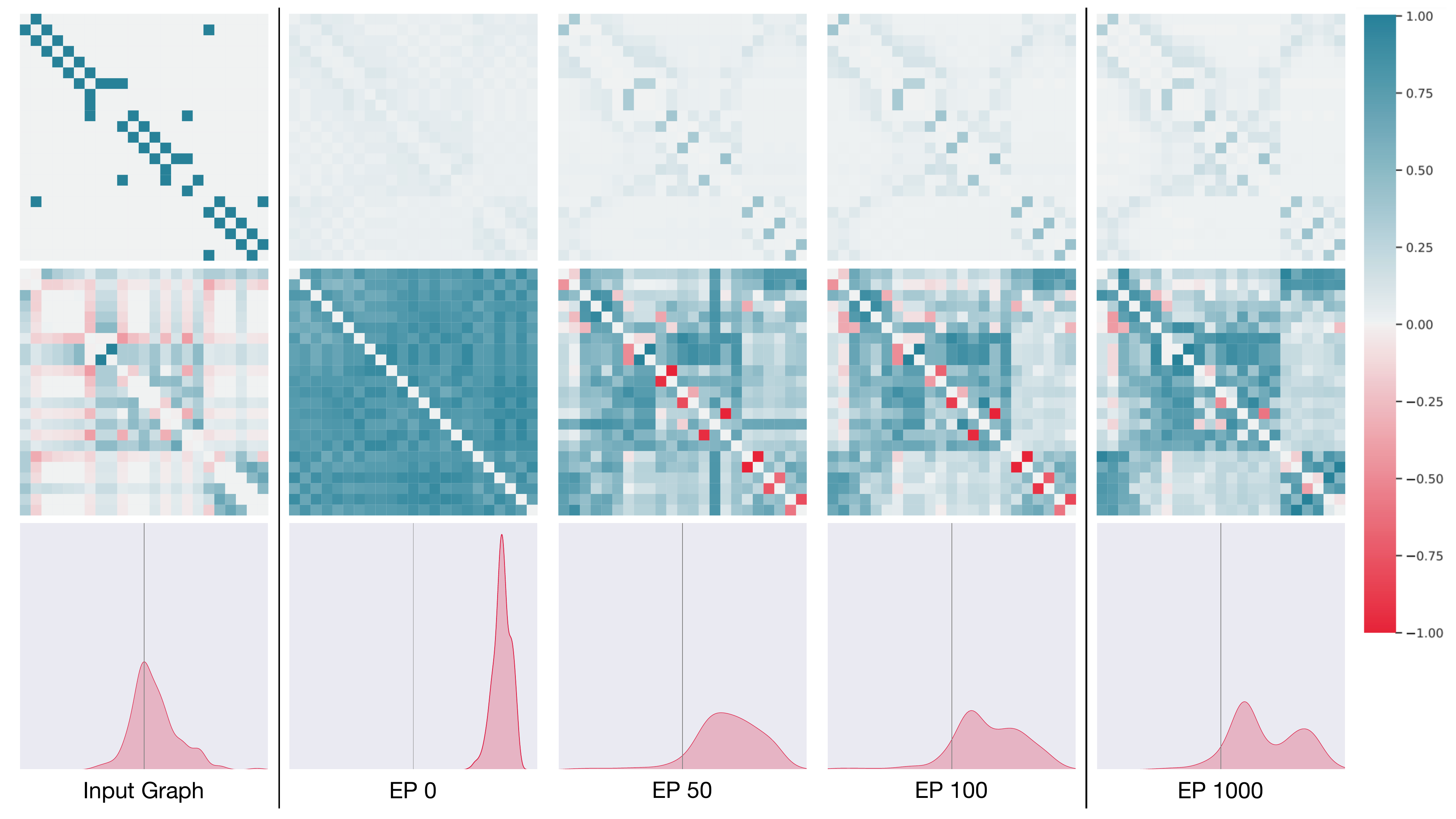}
        % \caption{Visualization of Attention and CURC for Graph-2 in ZINC}
        \label{fig:curc-zinc-2}
    \end{minipage}%
    \begin{minipage}[b]{0.33\textwidth}
        \centering
        \includegraphics[width=\textwidth]{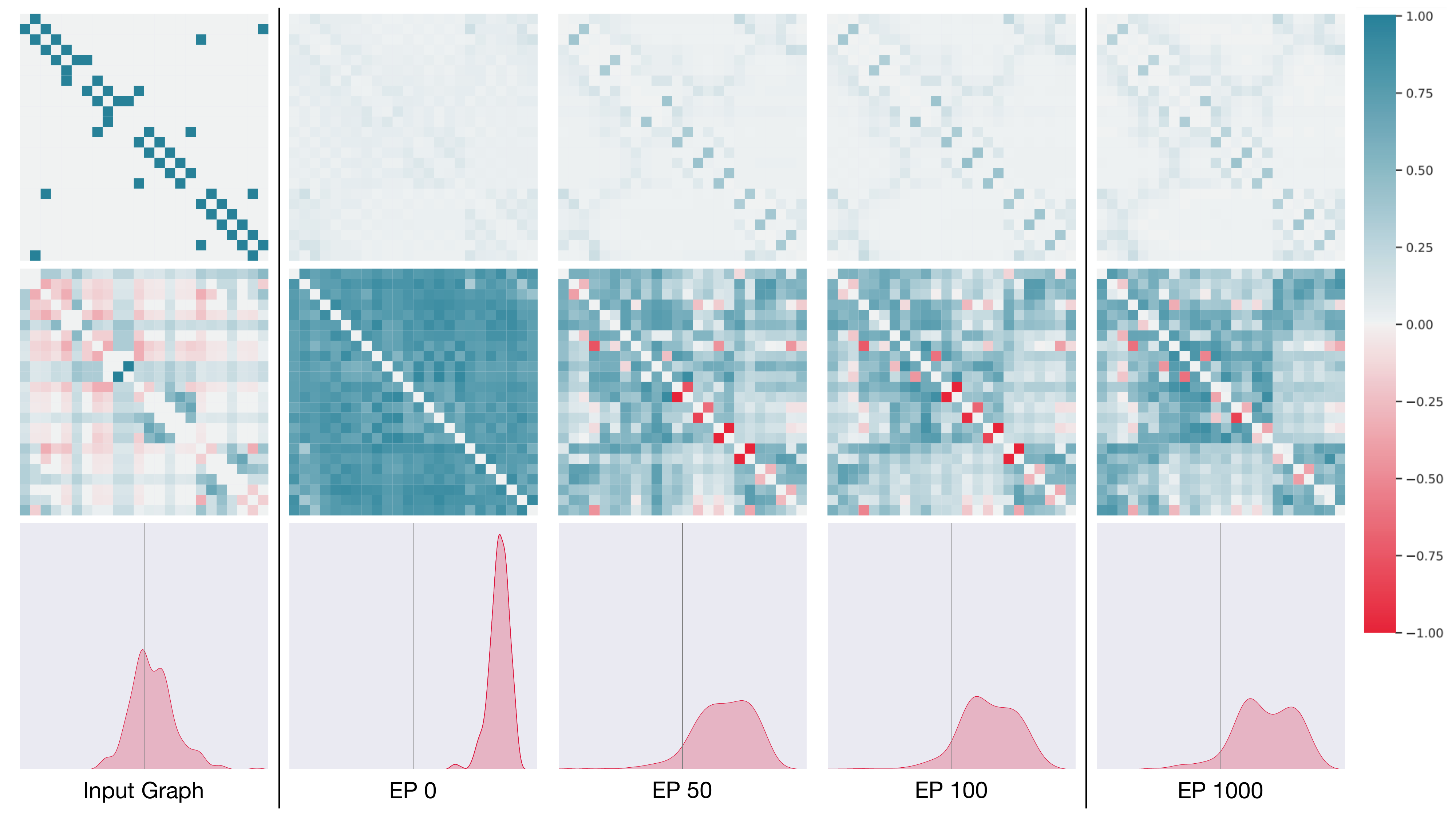}
        % \caption{Visualization of Attention and CURC for Graph-3 in ZINC}
        \label{fig:curc-zinc-3}
    \end{minipage}
    
    % Row 2
    \begin{minipage}[b]{0.33\textwidth}
        \centering
        \includegraphics[width=\textwidth]{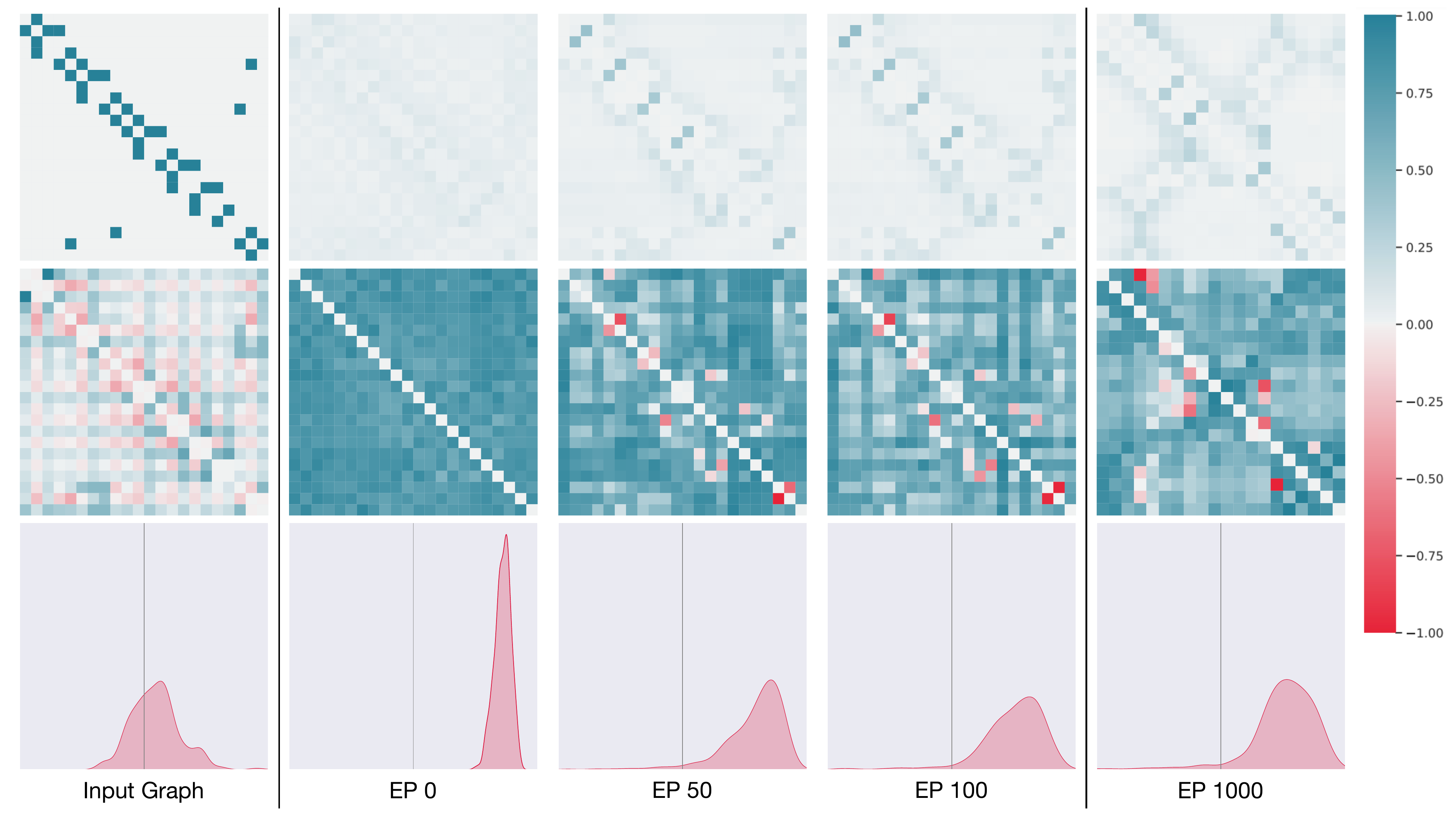}
        % \caption{Visualization of Attention and CURC for Graph-4 in ZINC}
        \label{fig:curc-zinc-4}
    \end{minipage}%
    \begin{minipage}[b]{0.33\textwidth}
        \centering
        \includegraphics[width=\textwidth]{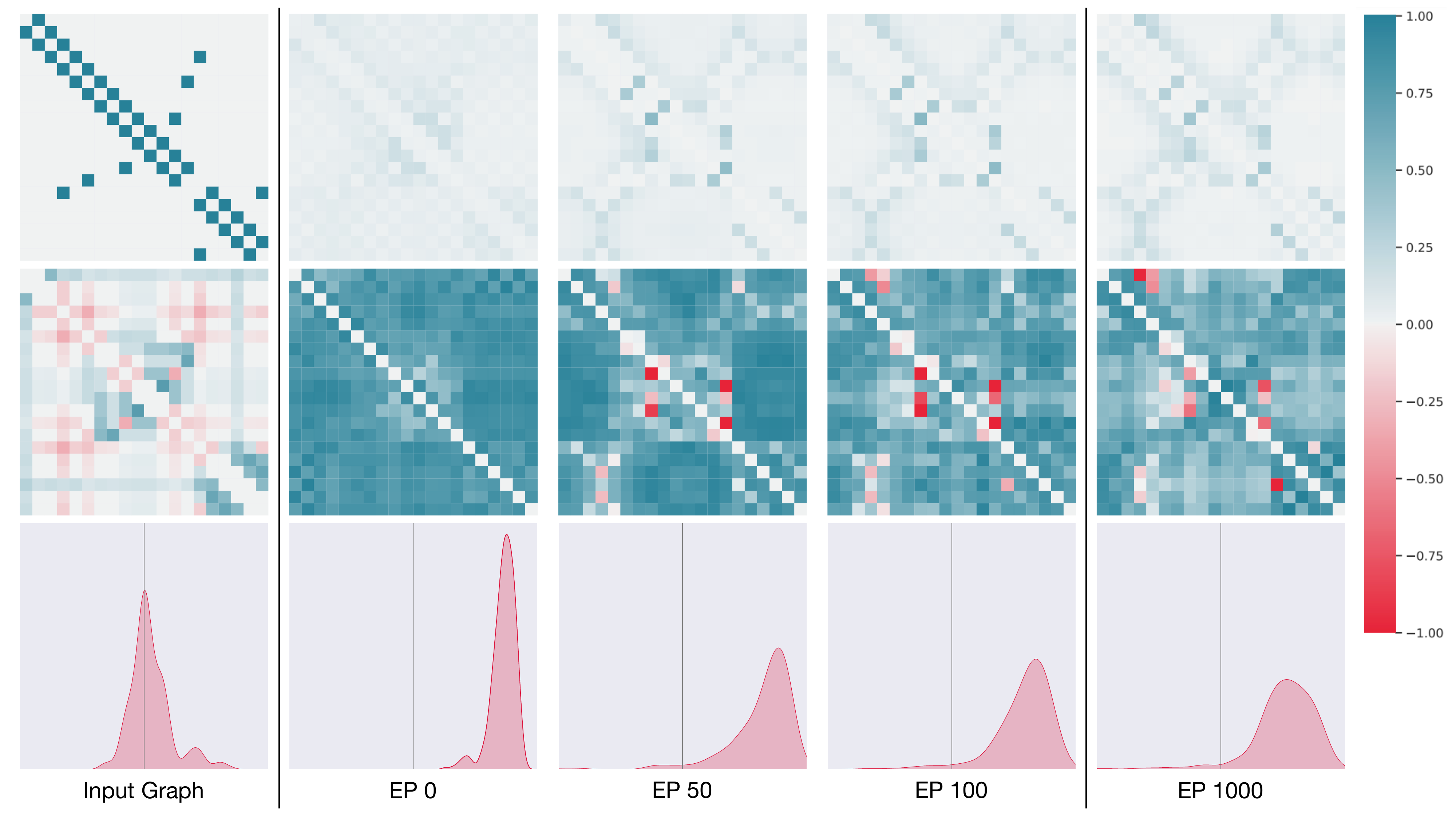}
        % \caption{Visualization of Attention and CURC for Graph-5 in ZINC}
        \label{fig:curc-zinc-5}
    \end{minipage}%
    \begin{minipage}[b]{0.33\textwidth}
        \centering
        \includegraphics[width=\textwidth]{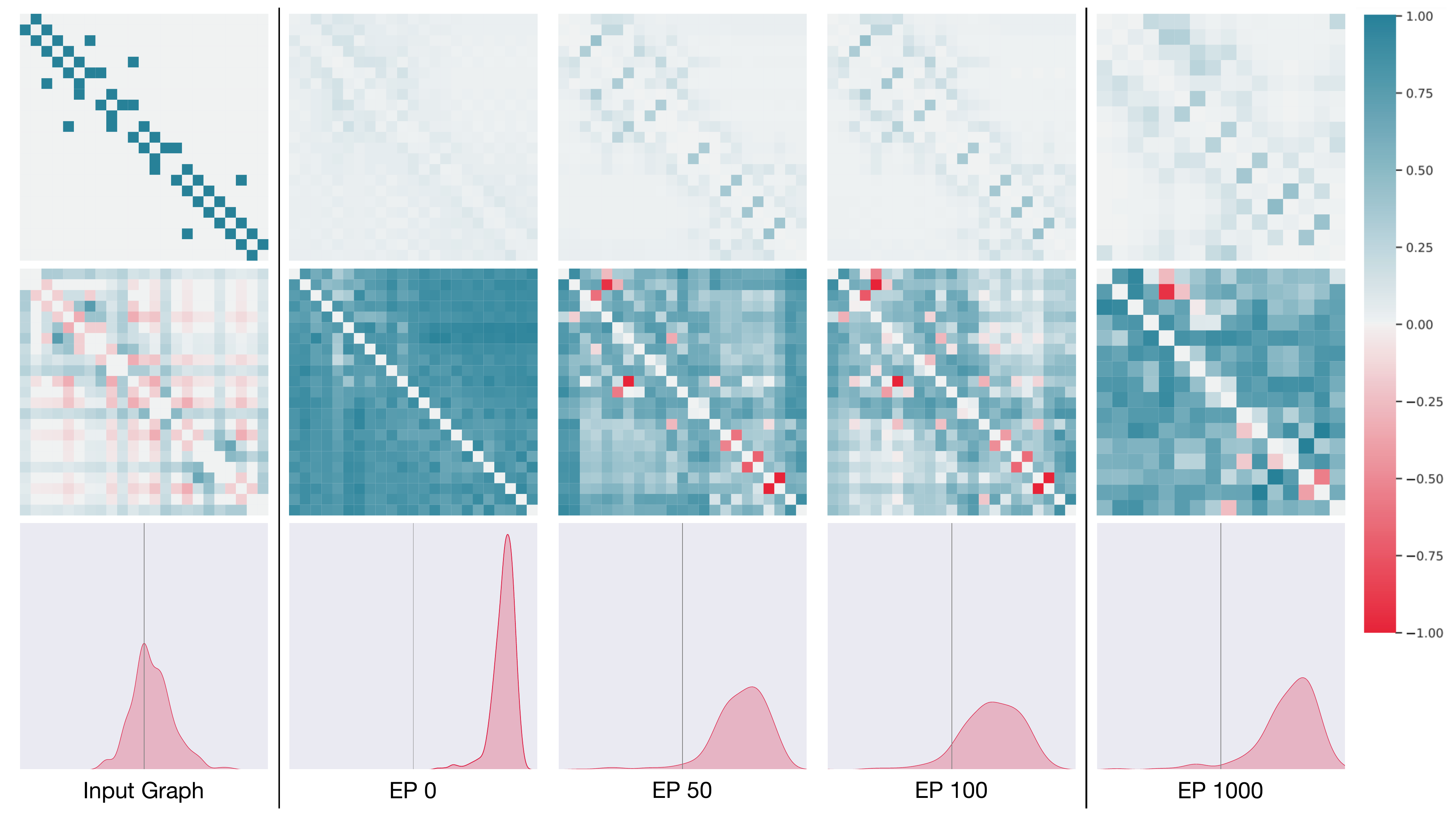}
        % \caption{Visualization of Attention and CURC for Graph-6 in ZINC}
        \label{fig:curc-zinc-6}
    \end{minipage}

    \caption{Visualization of Generalized Propagation and its Curvature for 6 graphs in ZINC at . 1st row: the visualization of propagation graph connection $\omega_{uv}$; 2nd row: the visualization of CURC $\kappa_{CURC}(u,v)$; 
    3rd row: the distributions of CURC; In the left most column, we visualize the input graph. Ep-0, Ep-50, Ep-100, Ep-1000 are listed afterwords.}
    \label{fig:curc_vs_attn}
\end{figure*}

\section{``Decurve Flow'' Phenomenon}
\label{sec:decirve_flow}
% \subsection{Analyzing Geometric Properties of Propagation Graphs with CURC}

% As discussed in the previous sections,
% with the proposed CURC, we are able to analyze the geometric properties of the propagation graphs of arbitrary GPNN variants.
% % the propagation graphs are the central of GPNNs.
% % With the tool CURC, we can analyze the geometric properties of the propagation graphs.
% Among many GPNN variants,
% we are especially interested in analyzing the propagation graphs of GTs,
% with which one might reveal the optimal propagation graphs.
% This can facilitate the understanding the graph learning and guide the design of other GPNN variants.
The proposed Continuous Unified Ricci Curvature (CURC) allows for in-depth analysis of geometric properties in GPNN variants, particularly focusing on Graph Transformers (GTs). This analysis aids in identifying optimal propagation graph structures, enhancing our understanding of graph learning and informing the design of new GPNN models.

\subsection{Analyzing the Learning Dynamic on CURC}

To better understand the behavior of GTs and exploring potential learning patterns,
we analyze the trend in CURC distributions of the propagation graphs during training.

We consider three classical graph transformers, Graphormer~\cite{ying2021TransformersReallyPerform}, SANs~\cite{kreuzer2021rethinking}, GRIT~\cite{ma2023graph} as well as a simplified, \textit{feat. ind.} variant of GRIT, namely GPNN-PE (details in Appx.~\ref{appendix:gpnn-pe}).

We first visualize 
the kernel density estimate (KDE) plots of CURC distributions on sampled graphs from ZINC datasets on 0, 50, 100, 150 and 200 training epochs (as shown in Fig.~\ref{fig:trend_curc}).
Based on the visualization, we reveal that these models all exhibit similar learning patterns:
the propagation graphs at the initial stage resemble a smoothed complete graph
% with edges of close weights,
resulting in a strongly right-skewed CURC distribution with nearly all positive curvatures.
As the training epochs increase,
the learned propagation graphs gradually shift towards the left, indicating the geometric patterns learned from the graphs.

We name this phenomenon as \textbf{\textit{decurve flow}}.
In fact, \textbf{\textit{Decurve flow}} can be connected to the recent study on the relationship between over-smoothing and over-squashing~\cite{giraldo2022understanding}.
\citet{giraldo2022understanding} reveal that small curvature values might cause the over-squashing problem,
whereas, 
over-large curvatures might, on the other hand, lead to over-smoothing issues.
Correspondingly,
we conjecture that,
even though start with random initialized complete propagation graphs, 
GTs learn to diminish the CURC to alleviate over-smoothing.
In other words, there might exist an optimal propagation graph, 
balancing the over-squashing and over-smoothing, 
which is potentially identifiable via further in-depth analysis on the CURC distributions.

\subsection{In-depth exploration on propagation matrices, CURC maps and CURC distributions}

To further understand \textbf{\textit{decurve flow}}, 
we go beyond the distribution and 
visualize propagation matrices (1st row), the corresponding CURC maps (2nd row) as well as the CURC distributions (3rd row) of GPNN-PE on 6 graphs from ZINC (shown in Fig~\ref{fig:curc_vs_attn}).
From the visualization, we observe that the CURC maps of GPNN-PE, from all large positive values, gradually learn small and even negative CURC values for certain node pairs.
These observations match our conjecture:
a well-optimized propagation matrix might deliberately diminish the information transport between two nodes, in order to prevent over-smoothing.

\section{Navigating GPNN Design}
\label{sec:experiment}
\newcommand{\first}[1]{\textcolor{SeaGreen}{#1}}
\newcommand{\second}[1]{\textcolor{BurntOrange}{#1}}
\newcommand{\third}[1]{\textcolor{Periwinkle}{#1}}
\newcommand{\cmark}{\ding{51}}%
\newcommand{\xmark}{\ding{55}}
 
GPNN framework unifies GTs, MPNNs and their variants by two key components, \textit{adjacency function} and \textit{connectivity function}.
Here, we first conduct a series of in-depth exploratory experiments on ZINC datasets~\cite{dwivedi2020BenchmarkingGraphNeural} to 
explore their design spaces 
and reveal the effective configurations as well as ineffective ones.
Then we conduct thorough experiments on five datasets from ~\cite{dwivedi2020BenchmarkingGraphNeural} and two from ~\cite{dwivedi2022LongRangeGraph} to properly justify our hypothesis on the configurations.

\subsection{Designs of Connectivity Functions}
\label{sec:exp_connect}

In Sec.\ref{sec: gpnn design space},
we discuss the taxonomy of the connectivity functions in our GPNN framework.
Here, we explore the impact of each potential design on the connectivity functions 
and identify the essential configurations.
Here, we conduct a thorough comparison experiment on ZINC-12K~\cite{dwivedi2020BenchmarkingGraphNeural} for different variants of GPNNs with different configurations:
(1) \textit{feat. dep.} \cmark v.s., \textit{feat. ind.} \xmark;
(2) \textit{dynamic} \cmark v.s. \textit{static} \xmark; 
(3) \textit{multi-head} \cmark v.s. \textit{single head} \xmark
; as well as 
(4)  \textit{global} \cmark v.s. \textit{local}\xmark;
The experimental results are shown in Table~\ref{tab:exp_connect}.

Empirically, compared to the full configuration (all \cmark),
changing \textit{feat. dep.} to \textit{feat. ind.} only leads to a statistically insignificant performance drop.
This empirical finding matches our theoretical analysis in Sec.~\ref{sec:expressiveness}:
the adjacency function is the dominant component of the expressive power on distinguishing non-isomorphism graphs, whereas the \textit{feat. ind.} counterpart is much less essential.
This drives us to conduct further ablation study based on this variant.
% Based on it, we use this variant, denoted as \textbf{GPNN-PE}(details in Appendix.??,\textcolor{blue}{Here, reference does not work} \mlh{to add in casting}), as the base model to further study the other potential configurations.
From the empirical results, 
changing \textit{global} to \textit{local}, \textit{dynamic} to \textit{static} and \textit{multi-head} to \textit{single-head} will demonstrate different levels of decline in efficacy.
The comparisons against two existing MPNNs demonstrate similar findings and also hint the importantness of the choices of adjacency functions. 

\begin{table*}[h!]
    \centering
    \caption{The Exploration of Connectivity Function}
    \vskip 0.12in
    {\scriptsize
    \centering
    % \resizebox{2*\columnwidth}{!}{
    \begin{tabular}{ccccccl}
    \toprule
         \textbf{ZINC} (\textbf{MAE}$\downarrow$) &  \makecell[c]{Local\xmark\\Global\cmark} &   \makecell[c]{Static\xmark\\Dynamic\cmark}&    \makecell[c]{Single \xmark \\ Multi. \cmark \\ Head. } &  \makecell[c]{Feat. \\ Ind. \xmark \\ Cond. \cmark}& Adj. Func. & Model \\
       % \cmidrule{2-6} 
       \midrule
 {${0.059 \pm 0.002}$}& \cmark &  \cmark & \cmark & \cmark  & 21-RRWP* & GRIT\\ \midrule
{${0.060 \pm 0.003}$} &   
\cmark & \cmark & \cmark &  \xmark & 21-RRWP  & GPNN-PE \\
{${0.064 \pm 0.002}$} 
 & \cmark & \xmark & \cmark & \xmark & 21-RRWP  & \quad + static  \\
{$0.066 \pm 0.005$} & \cmark &  
\cmark &  \xmark & \xmark & 21-RRWP & \quad + 1-head\\
{${0.068 \pm 0.003}$} &   
\xmark & \cmark & \cmark &  \xmark & 21-RRWP &  \quad + local \\ \midrule
$0.070 \pm 0.004$ & \xmark & \xmark&  \xmark & \xmark  & RWSE+1-RRWP & (GINE+RWSE)\\ 
$0.526 \pm 0.051$ & \xmark & \xmark& \xmark & \xmark & 1-RRWP &  (GIN) \\ 
       \bottomrule
 \multicolumn{5}{l}{
   *: 21-RRWP will naturally include RWSE with 21-order.}
    \end{tabular}
   % }
   }
\\
    \label{tab:exp_connect}
\end{table*}

\begin{table}[h!]
\vskip -0.1in
    \centering
    \caption{Adjacency Function;
    For absolute PE $\hat{\P}_i$,we reatin the usage as absolute PE $\P_{i,i}$ as well as concatenate them $\P_{i,j}=[\hat{\P}_i, \hat{\P}_j]$ to construct relative PE}
    \vskip 0.12in
    {\small
    % \resizebox{2*\columnwidth}{!}{
    \begin{tabular}{llc}
    \toprule
      & Adjacency Function    & \textbf{ZINC} (\textbf{MAE}$\downarrow$) \\
       % \cmidrule{2-6} 
       \midrule
 & 21-RRWP  &  {$\mathbf{0.060 \pm 0.003}$}  \\
 & SPD &  {$\mathbf{0.072 \pm 0.004}$} \\
 & pair-LapPE &  {$\mathbf{0.151 \pm 0.006}$}  \\
 & 1-RRWP &  {$\mathbf{0.125 \pm 0.006}$}  \\
 % & \makecell[c]{self. id.  
 % ($\approx$ DeepSet})&  $\mathbf{0.133 \pm 0.010}$  \\
       \bottomrule
    \end{tabular}
   % }
   }
\\
    \label{tab:exp_adj}
    \vskip -0.2in
\end{table}

\subsection{Designs of Adjacency Function}
\label{sec:exp_adj}

Following the previous exploration, 
we would like to verify the essentiality of the adjacency function on the empirical results.
Specifically, using GPNN-PE as the platform for comparison,
we conduct experiments on ZINC, comparing 4 choices of adjacency function designs: 21-order RRWP (more details in Appendix~\ref{rrwp})), paired Laplacian Positional encoding (pair-LapPE), 1-order RRWP (equivalent to random matrix plus self-identification), and shortest-path distance~\cite{ying2021TransformersReallyPerform} (as shown in Tab.~\ref{tab:exp_adj}).
% For pair-LapPE, the original absolute LapPE~\cite{dwivedi2021GeneralizationTransformerNetworks, kreuzer2021RethinkingGraphTransformers} will not only be injected into node attributes but also be paired and concatenated as relative PE for connectivity function.
The experimental results are shown in Table~\ref{tab:exp_adj}.
According to the empirical results, 
GPNN-PEs with different choices of adjacency functions demonstrate huge performance differences,
revealing the importance of the design choices on the adjacency functions.

\subsection{Benchmarking GPNN-PE}

To verify our findings in the previous exploratory experiments,
we would like to further benchmark our highlighted GPNN variant, GPNN-PE, in comparison to other typical graph models.
% we would like to verify our identification of the design choices as well as position the corresponding variant among various existing GPNN-variants.
Thus, we further evaluate GPNN-PE on 
% on five datasets from the Benchmarking GNNs work ~\cite{dwivedi2020BenchmarkingGraphNeural} and
two datasets from the Long-Range Graph Benchmark~(LRGB)~\cite{dwivedi2022LongRangeGraph} (shown in Tab.~\ref{tab:lrgb}). 
The empirical results further justify our theoretical findings: GPNN-PE reaches a comparable performance to GRIT and outperforms other GTs and MPNNs with weaker adjacency functions.
More experimental results and further details concerning the experimental setup can be found in Appendix~\ref{appendix:experiment_details}.

% Thus, we further evaluate GPNN-PE on 
% on five datasets from the Benchmarking GNNs work ~\cite{dwivedi2020BenchmarkingGraphNeural} and
% These datasets are among the most widely used graph benchmarks and cover diverse graph learning tasks,
% including node classification, graph classification, and graph regression, with a focus on graph structure and long-range dependencies. Further details concerning the experimental setup can be found in Appendix~\ref{appendix:experiment_details}.

\newcommand{\ourmethod}{GPNN-PE }

\begin{table}[h!]
% \begin{wraptable}{R}{7cm}
\vspace{-4mm}
\centering
\caption{Test performance on LRGB~\cite{dwivedi2022LongRangeGraph}. 
% Baselines' performance from \cite{rampasek2022RecipeGeneralPowerful}.
    Shown is the mean$\pm$s.d. of 4 runs. Highlighted are the top \first{first}, \second{second}, and \third{third} results.
    \#~Param $\sim 500K$. 
    % for both datasets.
    }~\label{tab:lrgb}
    \vskip 0.12in
    \scalebox{0.9}{
    \setlength{\tabcolsep}{2pt}{\small
    \begin{tabular}{lcc}
    \toprule
       \textbf{Model}  &  \textbf{Peptides-func} & \textbf{Peptides-struct} \\
       \cmidrule{2-3} 
       &  \textbf{AP}$\uparrow$ & \textbf{MAE}$\downarrow$ \\
       \midrule
       GCN  & $0.5930 \pm 0.0023$  & $0.3496 \pm 0.0013$ \\
GINE & $0.5498 \pm 0.0079$ & $0.3547 \pm 0.0045$ \\
GatedGCN & $0.5864 \pm 0.0035$ & $0.3420 \pm 0.0013$ \\
GatedGCN+RWSE & $0.6069 \pm 0.0035$ & $0.3357 \pm 0.0006$ \\
\midrule 
Transf.+LapPE & $0.6326 \pm 0.0126$ & {${0.2529 \pm 0.0016}$} \\
SAN+LapPE & $0.6384 \pm 0.0121$  & $0.2683 \pm 0.0043$ \\
SAN+RWSE& {${0.6439\pm 0.0075}$}  & $0.2545\pm 0.0012$ \\
GPS & \third{$\mathbf{0.6535 \pm 0.0041}$} & \third{$\mathbf{0.2500 \pm 0.0012}$}\\
GRIT & \first{$\mathbf{0.6988 \pm 0.0082}$} & \second{$\mathbf{0.2460 \pm 0.0012}$}
\\
\midrule
% PE-Prop & $\mathbf{0.6444}$ & $\mathbf{0.2464}$ \\
GPNN-PE & \second{$\mathbf{0.6954 \pm 0.0023}$} & \second{$\mathbf{0.2474 \pm 0.0010}$} \\
% \quad + static & \second{$\mathbf{0.6955 \pm 0.0057}$} & \first{$\mathbf{0.2454 \pm 0.0003}$} \\
% \quad + 1-head & {$0.6874 \pm 0.0161$} & \third{$\mathbf{0.2473 \pm 0.0013}$} \\
       \bottomrule
    \end{tabular}}}
    \label{lrgb}
% \end{wraptable}
\end{table}

% \section{Related Work}
% \label{sec:related_work}
% \import{sections/}{literature.tex}

\section{Conclusion and Future Work}
\label{sec:conclusion}
This study addresses the limitations of the traditional analysis of message-passing, central to graph learning, by defining generalized propagation with directed and weighted graphs. The introduced Generalized Propagation Networks (GPNNs) unifies Graph Transformers (GTs), Message Passing Neural Networks (MPNNs), and their variants through the use of directed, weighted propagation graphs. This approach aims to advances the field by incorporating "attention" mechanisms into graph analysis, which have been largely overlooked in traditional, undirected, and unweighted graph models. Our findings demonstrate that the expressiveness of GPNNs is fundamentally tied to the design of the adjacency function, thereby highlighting its theoretical and practical importance. Additionally, we have presented a taxonomy for GPNN-related models, illuminating the prevalent design choices and their implications through empirical analysis within this framework.
Furthermore, we proposed the Continuous Unified Ricci Curvature (CURC), an extension of the Ollivier-Ricci (OR) Curvature, tailored for directed and weighted graphs. The CURC exhibits key properties such as continuity, unity with OR curvature under specific conditions, and a novel lower bound relationship with the Dirichlet Isoperimetric Constant, facilitating enhanced analysis of graph bottlenecks. By CURC, we reveal the {\em \textbf{``decurve flow''}} phenomenon, calling for future research with propagation analysis on weighted-directed graphs.

\newpage
\bibliography{ref}
\bibliographystyle{unsrtnat}

%%%%%%%%%%%%%%%%%%%%%%%%%%%%%%%%%%%%%%%%%%%%%%%%%%%%%%%%%%%%%%%%%%%%%%%%%%%%%%%
%%%%%%%%%%%%%%%%%%%%%%%%%%%%%%%%%%%%%%%%%%%%%%%%%%%%%%%%%%%%%%%%%%%%%%%%%%%%%%%
% APPENDIX
%%%%%%%%%%%%%%%%%%%%%%%%%%%%%%%%%%%%%%%%%%%%%%%%%%%%%%%%%%%%%%%%%%%%%%%%%%%%%%%
%%%%%%%%%%%%%%%%%%%%%%%%%%%%%%%%%%%%%%%%%%%%%%%%%%%%%%%%%%%%%%%%%%%%%%%%%%%%%%%
\newpage
\appendix
\onecolumn
% \input{others/exp_variants}
% \input{others/notation_and_mpnn}

% \newpage
% \import{appendix/}{appendix_background.tex}
% \import{appendix/}{appendix_expressiveness.tex}
% \import{appendix/}{appendix_gpnns_casting.tex}
% \import{appendix/}{appendix_exp.tex}
% \import{appendix/}{appendix.tex}
% \import{appendix/}{curvature.tex}

\clearpage
\appendix
\appendixpage
\startcontents[section]
\printcontents[section]{l}{0}{\setcounter{tocdepth}{2}}

\clearpage
% \section{Appendix}
\clearpage
\section{Detailed Proofs for CURC}
\setcounter{equation}{0}
\renewcommand{\theequation}{\arabic{equation}}

\subsection{Kantorovich-Rubinstein duality}

Kantorovich-Rubinstein duality is an important result in the field of optimal transportation, which establishes the connection between optimal transportation problem and linear programming problem. The most common form of duality is stated in the context of Polish metric space. While in the setting up of $\kappa_{\operatorname{CURC}}$, the distance function as weighted shortest distance is not necessarily symmetric, which fails to define a metric space. Luckily, the duality still holds under weaker assumption without symmetry assumption. Here, we give a short proof of Kantorovich-Rubinstein duality for the sake of completeness.

\begin{definition} \textbf{(L-Lipschitz)}
    Let $d: \mathcal{X} \times \mathcal{X} \mapsto \R^{\geq0}$ be an asymmetric definite distance function on $\mathcal{X}$, we say $f: \mathcal{X} \mapsto \R$ is L-Lipschitz w.r.t. $d$ if
    $$
    \forall x, y \in \mathcal{X}, f(y)-f(x) \leq L d(x, y).
    $$
\end{definition}

\begin{definition} \textbf{(Support)}
    Let $\mathcal{X}$ be a finite set and $\mu: \mathcal{X} \mapsto [0, 1]$ be the corresponding probability measure, we define the \textit{support} of $\mu$ to be 
    $$
        \operatorname{supp}(\mu) = \{x \in \mathcal{X} : \mu(x) > 0 \}.
    $$
\end{definition}

\begin{definition}\textbf{(Coupling)} 
    Suppose $\mu$ and $\nu$ to be two probability distribution on finite sets $\mathcal{X}$ and $\mathcal{Y}$ respectively. Let $\Pi\left(\mu, \nu\right)$ denotes the set of \textit{couplings} between $\mu$ and $\nu$. We say $\pi: \mathcal{X} \times \mathcal{Y} \mapsto [0, 1]$ $\in \Pi\left(\mu, \nu\right)$ is a \textit{coupling} if
    $$
    \sum_{y \in V} \pi(x, y)=\mu(x), \quad \sum_{x \in V} \pi(x, y)=\nu(y).
    $$
\end{definition}

\begin{definition}\textbf{(C-convexity)}\label{c-convexity}
    Let $\mathcal{X}$ and $\mathcal{Y}$ be two sets and $c: \mathcal{X} \times \mathcal{Y} \mapsto \R \cup \{+\infty\}$. A function $f: \mathcal{X} \mapsto \R \cup \{+\infty\}$ is \textit{$c$-convex} if it is not identically $+\infty$, and there exists $\psi: \mathcal{Y} \mapsto \R \cup \{+\infty\}$ such that 
    $$
    \forall x \in \mathcal{X}, f(x) = \sup_{y \in \mathcal{Y}} (\psi(y) - c(x, y)).
    $$
    Then its corresponding \textit{$c$-transform} is the function $\psi^{c}$ defined by 
    $$
    \forall y \in \mathcal{Y}, f^{c}(y) = \inf_{x \in \mathcal{X}} (\psi(x) + c(x, y)).
    $$
\end{definition}

\begin{lemma}\label{asymmetric distance}
    Let $f: \mathcal{X} \mapsto \R$ be a function defined on a set $\mathcal{X}$. Let $d: \mathcal{X} \times \mathcal{X} \mapsto \R^{\geq 0}$ to be a distance function on $\mathcal{X}$ that satisfies the following properties:
    \begin{itemize}
        \item $\forall x \in \mathcal{X}$, $d(x, x) = 0$
        \item $\forall x \neq y \in \mathcal{X}, d(x, y) > 0$
        \item $\forall x, y, z \in \mathcal{X}$, $d(x, z) + d(z, y) \geq d(x, y)$
    \end{itemize}
    Then function $f$ is \textit{$d$-convex} $\iff$ $f$ is 1-Lipschitz w.r.t. distance function $d$.
    \end{lemma}
    \begin{proof}
    We first suppose $f$ is \textit{$d$-convex}, we want to show that $\forall x, y \in \mathcal{X}, f(x) - f(y) \leq d(y, x)$. By the definition of \textit{$c$-convex}, $\exists$ function $\psi: \mathcal{X} \mapsto \R$, such that $f(x) = \sup_{z \in \mathcal{X}} [\psi(z) - d(x, z)]$ and $f(x) = \sup_{z \in \mathcal{X}} [\psi(z) - d(y, z)]$. Suppose $z_0 = \operatorname{argsup}_{z \in \mathcal{X}}[\psi(z) - d(x, z)]$
    \begin{align*}
    f(x) - f(y) &= \sup_{z \in \mathcal{X}}[\psi(z) - d(x, z)] - \sup_{z \in \mathcal{X}}[\psi(z) - d(y, z)] \\
    &\leq [\psi(z_0) - d(x, z_0)] - [\psi(z_0) - d(y, z_0] \\
    &= d(y, z_0) - d(x, z_0) \\
    &\leq d(y, x) \text{ by triangular inequality}
    \end{align*}
    Now, suppose $f$ is 1-Lipschitz w.r.t. distance function $d$. We note that $f^c(y) = \inf_{x \in \mathcal{X}}[f(x) + d(x, y)]$. By 1-Lipschitz, we have that 
    \begin{align*}
        &f(y) - f(x) \leq d(x, y) \\
        \Rightarrow&f(x) - f(y) \geq -d(x, y) \\ 
        \Rightarrow&f(x) \geq f(y) - d(x, y) \\ 
        \Rightarrow&f(x) \geq \sup_{y \in \mathcal{X}} [f(y) - d(x, y)]
    \end{align*}
    by taking $x=y$ in the supremum and $d(x, x) = 0$, we have the following equality:
    $$
    f(x) = \sup_{y \in \mathcal{X}} [f(y) - d(x, y)]
    $$
    
    By the exact same argument, we can derive a bonus property that
    $$
    f^c(y) = \inf_{x \in \mathcal{X}}[f(x) + d(x, y)] = f(y).
    $$
    Therefore, on a set $\mathcal{X}$ with asymmetric distance function that satisfies triangle inequality, function $f$ is \textit{$d$-convexity} $\iff$ $f$ is 1-Lipschitz w.r.t. distance function $d$. And its \textit{$c$-transform} is itself.
\end{proof}

\begin{theorem} \textbf{(K-R duality)} \label{KR duality proof}
    Let $\V$ be a discrete finite set equipped with a possibly asymmetric distance function $d: \V \times \V \mapsto \R^{\geq 0}$ that is definite and satisfies triangle inequality. Suppose $\mu$ and $\nu$ to be two probability measure on $\V$. Then we have the Kantorovich--Rubinstein duality:
    \begin{equation}\label{duality}
        \inf _{\pi \in \Pi\left(\mu, \nu\right)} \sum_{x, y \in \V} d(x, y) \pi(x, y) = \sup_{f \in \operatorname{Lip}(1)} \sum_{x \in \V} f(x)(\nu(x) - \mu(x)),
    \end{equation}
    where $\Pi\left(\mu, \nu\right)$ denotes the coupling between probability measure $\mu$ and $\nu$ and $f \in \operatorname{Lip}(1)$ denotes that $f: \V \mapsto \R$ is 1-Lipschitz w.r.t. distance function $d$.
\end{theorem}

\begin{proof}
    We prove equation \ref{duality} by first showing that 
    \begin{equation}\label{duality lhs}
        \inf _{\pi \in \Pi\left(\mu, \nu\right)} \sum_{x, y \in \V} d(x, y) \pi(x, y) \geq \sup_{f \in \operatorname{Lip}(1)} \sum_{x \in \V} f(x)(\nu(x) - \mu(x)). 
    \end{equation}
    Then we provide a specific construction of 1-Lipschitz function $f: \V \mapsto \R$ showing the converse,
    \begin{equation}\label{duality rhs}
        \inf _{\pi \in \Pi\left(\mu, \nu\right)} \sum_{x, y \in \V} d(x, y) \pi(x, y) \leq \sup_{f \in \operatorname{Lip}(1)} \sum_{x \in \V} f(x)(\nu(x) - \mu(x)).
    \end{equation}
    
    Firstly, take arbitrary $\pi \in \Pi\left(\mu, \nu\right)$ and $f$ 1-Lipschitz, we have the following algebraic property:
    
    \begin{align*}
        \sum_{x \in \V} f(x)(\nu(x) - \mu(x)) &= \sum_{x, y \in \V} f(x)\pi(x, y) - \sum_{x, y \in \V} f(y)\pi(x, y) \\
        &= \sum_{x, y \in \V} [f(y)-f(x)]\pi(x, y) \\
        &\leq \sum_{x, y \in \V} d(x, y)\pi(x, y)
    \end{align*}
    Therefore, 
    \begin{align*}
        &\inf _{\pi \in \Pi\left(\mu, \nu\right)} \sum_{x, y \in \V} d(x, y) \pi(x, y) \geq \sum_{x \in \V} f(x)(\nu(x) - \mu(x)) \\
        \Rightarrow &\inf _{\pi \in \Pi\left(\mu, \nu\right)} \sum_{x, y \in \V} d(x, y) \pi(x, y) \geq \sup_{f \in \operatorname{Lip}(1)} \sum_{x \in \V} f(x)(\nu(x) - \mu(x)),
    \end{align*}
    which is exactly equation \ref{duality lhs}.

    Secondly, we construct a function with the help of \textit{$c$-convexity} introduced in definition \ref{c-convexity}. For a fixed $m \in \mathbb{N}$, we can pick a sequence $(x_i, y_i)_{i=0}^{m} \in \operatorname{supp}(\pi)$. Note that this choice of $m$ is finite and $m \leq |\V|^{2}$. We construct our function $f$ by
    \begin{equation}
        f(x) := \sup_{m \in \mathbb{N}}\sup_{(x_i, y_i)_{i=0}^{m}} \{\sum_{i = 0}^{m-1} [d(x_i, y_i) - d(x_{i+1}, y_i)] + [d(x_m, y_m) - d(x, y_m)]\}.
    \end{equation}
    We now want to show that $f^c(y) - f(x) = d(x, y)$ almost surely for $(x, y) \in \operatorname{supp}(\pi)$. Note that 
    \begin{align*}
        &f^c(y) = \inf_{x \in \mathcal{X}} [f(x) + d(x, y)] \\
        \Rightarrow &f^c(y) \leq f(x) + d(x, y) \\
        \Rightarrow &f^c(y) - f(x) \leq d(x, y).
    \end{align*}
    By lemma \ref{asymmetric distance}, we have that $f^c(y) = f(y)$, which implies that this choice of $f$ guarantees 1-Lipschitz. \\ 
    Suppose $(x, y) \in \operatorname{supp}(\pi)$, and in the choice of sequence $(x_i, y_i)_{i=0}^{m}$ we can let $(x_m, y_m) := (x, y)$. Therefore,
    \begin{align*}
        f(z) \geq &\sup_{m \in \mathbb{N}}\sup_{(x_i, y_i)_{i=0}^{m}} \{\sum_{i = 0}^{m-2} [d(x_i, y_i) - d(x_{i+1}, y_i)] + [d(x_{m-1}, y_{m-1}) - d(x, y_{m-1})]\\
        &+ [d(x, y) - d(z, y)]\} \\ 
        = & f(x) + d(x, y) - d(z, y).
    \end{align*}
    The last equality comes from the fact that in the definition of $f$, taking supremum over $m$ or $m-1$ does not matter. Hence,
    \begin{align*}
        &f(z) + d(z, y) \geq f(x) + d(x, y) \\
        \Rightarrow &\inf_{z \in \V} [f(z) + d(z, y)] \geq f(x) + d(x, y) \\ 
        \Rightarrow &f^c(y) \geq f(x) + d(x, y) \\ 
        \Rightarrow &f^c(y) - f(x) \geq d(x, y) \\ 
    \end{align*}
    Therefore, we have that $\forall (x, y) \in \operatorname{supp}(\pi), f^c(y) - f(x) = d(x, y)$ $\iff$ $f(y) - f(x) = d(x, y)$ by using lemma \ref{asymmetric distance} again. For clarity, we will denote this choice of $f$ to be $f^{*}$. Using this result, we have $\forall \pi \in \Pi\left(\mu, \nu\right),$
    \begin{align*}
        \sup_{f \in \operatorname{Lip}(1)} \sum_{x \in \V} f(x)(\nu(x) - \mu(x)) &\geq \sum_{x \in \V} f^{*}(x)(\nu(x) - \mu(x)) \\
        &= \sum_{x, y \in \V} [f^{*}(y)-f^{*}(x)]\pi(x, y) \\
        &= \sum_{x, y \in \V} d(x, y)\pi(x, y) \\ 
        &\geq \inf _{\pi \in \Pi\left(\mu, \nu\right)} \sum_{x, y \in \V} d(x, y) \pi(x, y)
    \end{align*}
    Therefore, both equations \ref{duality lhs} and \ref{duality rhs} hold, hence we have proven the Kantorovich--Rubinstein duality under the weaker assumption that distance function $d: \V \times \V \mapsto \R^{\geq0}$ is not necessarily symmetric.
\end{proof}

\begin{corollary}\label{linear programming}
    Suppose $\V$ to be a discrete finite set equipped with a possibly asymmetric distance function $d: \V \times \V \mapsto \R^{\geq 0}$ that satisfies triangle inequality. Suppose $\mu$ and $\nu$ to be two probability measure on $\V$ with support on $\left\{x_1, x_2, \ldots, x_n\right\}$ and $\left\{y_1, y_2, \ldots, y_m\right\}$ respectively. Then solving the Wasserstein distance $\mathcal{W}_1(\mu, \nu) = \inf _{\pi \in \Pi\left(\mu, \nu\right)} \sum_{x, y \in \V} d(x, y) \pi(x, y)$ is equivalent with
    \begin{equation}
        W_1(\mu, \nu)=\sup _{f \in \operatorname{Lip}(1)}\left\{\sum_i f\left(x_i\right) \nu\left(x_i\right)-\sum_j f\left(y_j\right) \mu\left(y_j\right)\right\},
    \end{equation}
    which is further equivalent to the following Linear programming problem
    \begin{equation}
        W_1(\mu, \nu)=\sup _{A f \preceq c} m^T f,
    \end{equation}
    with the following construction of matrix and vectors:
    \begin{align*} 
    &m :=(\mu\left(x_1\right), \ldots, \mu\left(x_n\right), \nu\left(y_1\right), \ldots, \nu\left(y_m\right))^T \in \mathbb{R}^{n+m}, \\ 
    &\phi:=(f\left(x_1\right), \ldots, f\left(x_n\right),-f\left(y_1\right), \ldots,-f\left(y_m\right))^T \in \mathbb{R}^{n+m}, \\ 
    &c :=(d\left(x_1, y_1\right), \ldots, d\left(x_1, y_m\right), d\left(x_2, y_1\right), \ldots, d\left(x_n, y_1\right), \ldots, d\left(x_n, y_m\right), \\
    &d\left(y_1, x_1\right), \ldots, d\left(y_m, x_1\right), d\left(y_1, x_2\right), \ldots, d\left(y_1, x_n\right), \ldots, d\left(y_m, x_n\right) )^T \in \mathbb{R}^{n m}, \\ 
    &A := \left(\begin{array}{cc}A_1 \\ A_2\end{array}\right), \text{where } A_1:=\left(\begin{array}{cc}a_1 & I_m \\ a_2 & I_m \\ \vdots & \vdots \\ a_n & I_m\end{array}\right), A_2:= -A_1, a_i \in \R^{m \times n} \text{ with all ones at \textit{i}-th column and zeros otherwise.}
    \end{align*}
\end{corollary}

\subsection{Analytical and Algebraic Properties of Continuous Unified Ricci Curvature}

To prove Proposition \ref{properties of CURC}, we perceive edge weights $\omega \in R^{n \times n}$ in its matrix form to keep in line with other linear algebra lemmas required in this section.

\begin{lemma}\label{perron-frobenius}s
    We say a matrix $A \in \R^{n \times n}$ is \textit{regular} if for some $k \geq 1$, $A^{k} > 0$. Or equivalently, matrix $A$ has non-negative entries and is strongly connected in our context. Then by \textit{Perron-Frobenius theorem}:
    \begin{itemize}
        \item There exists a unique positive unit left eigenvector $\textbf{v}_{pf}$ of $A$ called Perron-Frobenius left eigenvector, whose corresponding eigenvalue $\lambda_{pf}$ is real and has the largest norm among all eigenvalues.
        % \item Let $\lambda_{pf}$ be the corresponding eigenvalue of $\textbf{v}_{pf}$, for any other eigen
        \item $\lambda_{pf}$ is \textit{simple}, \textit{i.e.} has multiplicity one.
    \end{itemize}
\end{lemma}

\begin{proof}
    Perron-Frobenius theorem is well-known in the field of linear algebra, and has different forms on non-negative matrices, non-negative regular matrices and postivie matrices. We only need it for non-negative regular matrices.
\end{proof}

\begin{lemma}\label{perron eigenvector continuity}
    Let $A(t)$ be a differentiable matrix-valued function of $t$, $a(t)$ an eigenvalue of $A(t)$ of multiplicity one. Then we can choose an eigenvector $h(t)$ of $A(t)$ pertaining to the eigenvalue $a(t)$ to depend differentiably on $t$.
\end{lemma}

\begin{proof}
    For the purpose of our proof, we only need continuity of $h(t)$ on $t$, but we present this stronger statement, cf. Theorem 8, p130 in \citet{lax2007linear}.
\end{proof}

\begin{lemma}\label{perron measure continuity}
    Suppose arbitrary matrix $B \in \R^{n \times n}$ is a non-negative regular matrix. Then its Perron-Frobenius left eigenvector $\textbf{v}_{pf}$ depend continuously on the $B$ w.r.t. \textbf{small} perturbation $\varepsilon$ entry-wise, restricting to $B$ being non-negative and regular after the perturbation.
\end{lemma}

\begin{proof}
    Let $E_{ij}$ denotes a matrix with zero entries except for entry $(i, j)$. Let $A(\varepsilon):= B + E_{ij} \varepsilon$, which is obviously a matrix-valued function differentiable w.r.t. $\varepsilon$. Suppose $|\varepsilon|$ is small such that we are only dealing with non-negative regular $A(\varepsilon)$. Therefore by Perron-Frobenius theorem \ref{perron-frobenius}, there exists $\lambda_{pf}(\varepsilon)$ and $\textbf{v}_{pf}(\varepsilon)$ for $A(\varepsilon)$, which has multiplicity 1. Therefore, by lemma \ref{perron eigenvector continuity}, we have that $\textbf{v}_{pf}(\varepsilon)$ continuously depends on $\varepsilon$. Note that this eigenvector unnecessarily has unit length. But fortunately, the 2-norm of a positive continuous vector function is also continuous w.r.t. $\varepsilon$, we have that the Perron-measure $\mathfrak{m} := \frac{\textbf{v}_{pf}(\varepsilon)}{\| \textbf{v}_{pf}(\varepsilon) \|}$ is continuous w.r.t. $\varepsilon$ element-wise. 
    % Furthermore, by entry-wise continuity, we actually have Perron-measure $\mathfrak{m}$ as a vector-valued function is continuous w.r.t. matrix $\omega$. 
\end{proof}

\begin{lemma}\label{continuity of mean transition prob}
    The \textit{mean transition probability} $\mu_x$ for vertex $x$ defined in equation \ref{mean transition kernel} is continuous w.r.t. weight matrix $\omega$ entry-wise, restricting to $\omega$ being non-negative and regular after the perturbation.
\end{lemma}

\begin{proof}
    By lemma \ref{perron measure continuity}, we have that the Perron-measure $\mathfrak{m}$ is a continuous function w.r.t. $\omega$ entry-wise. Since $\forall$ $x \in \V$, $\mathfrak{m}(x) > 0$, we have $\forall x, y \in \V$, $\frac{\mathfrak{m}(y)}{\mathfrak{m}(x)}$ is continuous w.r.t. $\omega$ entry-wise. \\
    Now we consider normalized weight $W$. WLOG, we suppose a perturbation of $\delta$ on $\omega$ in entry $(i, j)$, which only influences the \textit{i}-th row of $W$. Denote this perturbed normalized weight matrix by $W^{*}$, and we have that
    \[ W^{*}(x, y) = 
    \begin{cases}
        \frac{W(x, y)}{1 + \delta}  & \text{if } x = i, y \neq j \\
        \frac{W(x, y) + \delta}{1 + \delta} & \text{if } x = i, y = j \\
        W(x, y) & \text{if } x \neq i
    \end{cases}
    \]
    Therefore if we choose to perturb the $(i, j)$ entry of $W$, the value of $W(x, y)$ is indifferent to this entry, hence continuous. For $W(i, y)$ where $y \neq j$, we pick $\varepsilon > 0$. By choosing $\delta \leq \frac{\varepsilon}{W(i, y) - \varepsilon}$, we ensure $W(x, y) - \frac{W(x, y)}{1 + \varepsilon} \leq \varepsilon$, hence $W(i, y)$ is continuous w.r.t. $\omega$ entry-wise. For $W(i, j)$, we also pick $\varepsilon > 0$, and we choose $\delta \leq \frac{\varepsilon}{1 - W(i, j) - \varepsilon}$ to get the entry-wise continuity. Therefore, $\forall x \in \V$, the mean transition measure $\mu_x$
    $$
    \mu_x(y):= \frac{1}{2}[W(x, y) + \frac{\mathfrak{m}(y)}{\mathfrak{m}(x)} W(y, x)],
    $$
    is a continuous function w.r.t. $\omega$ entry-wise, as summation and product of two continuous function is still continuous.
\end{proof}

\begin{lemma}\label{convex problem continuity}
    Consider the convex optimization problem with a valid solution:
    \begin{equation}\label{convex optimization}
        \mathcal{M} := \inf_{A f \preceq c(t)} m^T f,
    \end{equation}
    where $A \in \R^{n \times m}, f \in \R^{m \times 1}, m \in \R^{n \times 1}$, and c(t): $\R \mapsto \R^{n \times 1}$ being a vector-valued function that depends continuously on $t \in \R$. We claim that $\mathcal{M}$ also depends continuously on $t$.
\end{lemma}

\begin{proof}
    Note that equation \ref{convex optimization} is a convex optimization problem and admits a feasible solution. Therefore the optimization problem admits strong duality from the Slater's condition and is equivalent to the following dual problem:
    \begin{align*}
        &\text{maximize } \quad-c(t)^T g \\
        &\text{subject to } \quad A^T g+m=0, \quad g \succeq 0.
    \end{align*}
    In particular, the maximization problem has $\mathcal{M}$ as the optimal value. From the strong duality of convex optimization problem, we move the continuous function $c(t)$ from the constraint to the objective. Note that $-c(t)^Tg$ is a continuous function of $t$, therefore the supremum over $-c(t)^Tg$ is also a continuous function.
\end{proof}

\begin{lemma}\label{convex continuity extension}
    Consider the convex optimization problem with a valid solution:
    $$
    \mathcal{M} := \inf_{A f \preceq c(t)} m(t)^T f,
    $$
    where $A \in \R^{n \times m}, f \in \R^{m \times 1}$ and $m(t): \R \mapsto \R^{n \times 1}$ and $c(t): \R \mapsto \R^{n \times 1}$ being vector-valued functions that depend continuously on $t \in \R$. We claim that $\mathcal{M}$ depends continuously on $t$.
\end{lemma}

\begin{proof}
    We extend the lemma to the case where the objective is also a continuous function of $t$. Let $\delta > 0$ be a small perturbation, and $\mathcal{M}^{*} = \inf_{A f \preceq c(t+\delta)} m(t+\delta)^T f$. Therefore, we have that
    \begin{align*}
        |\mathcal{M} - \mathcal{M^{*}}| &= |\inf_{A f \preceq c} m^T f - \inf_{A f \preceq c^{*}} m^{*T} f| \\
        &= |\inf_{A f \preceq c} m^T f - \inf_{A f \preceq c^{*}} m^T f + \inf_{A f \preceq c^{*}} m^T f - \inf_{A f \preceq c^{*}} m^{*T} f| \\ 
        &\leq |\inf_{A f \preceq c} m^T f - \inf_{A f \preceq c^{*}} m^T f| + |\inf_{A f \preceq c^{*}} m^T f - \inf_{A f \preceq c^{*}} m^{*T} f|
    \end{align*}
    By lemma \ref{convex problem continuity} and continuity of supremum over continuous function, $\mathcal{M}$ depends continuously on variable $t$.
\end{proof}

The following is the proof of three important properties of CURC, the first one and the third one are natural results from the construction of $\kappa_\mathrm{CURC}$, but we stress that the second note on continuity of $\kappa_\mathrm{CURC}$ is non-trivial and new. 

\begin{proposition}\label{Curvature property}
We have the following properties for $\kappa_\mathrm{CURC}$: 
    \begin{itemize}
        \item (Unity) For connected unweighted-undirected graph $\gG=(\V, \E)$, for any pair of vertices $x, y \in \V$, we have $\kappa_\mathrm{CURC}(x, y) = \kappa_{\mathrm{OR}}(x, y).$
        \item (Continuity) If we perceive $\kappa_{\mathrm{CURC}}(x, y)$ as a function of $\omega$, then $\kappa_{\mathrm{CURC}}(x, y)$ is continuous w.r.t. $\omega$ entry-wise.
        \item (Scale Invariance) For strongly-connected weighted-directed graph $\gG=(\V, \E, \omega)$, when all edge weights $\omega$ are scaled by an arbitrary positive constant $\lambda$, the value of  $\kappa_\mathrm{CURC}(x, y)$ for any $x, y \in \V$ is unchanged. 
    \end{itemize}
\end{proposition}

\begin{proof} We prove the properties by sequential order.
    \begin{enumerate}
        \item \textbf{Proof of Unity} \\
        Suppose $A$ to be the adjacency matrix (binary) of unweighted-undirected graph $\gG = (\V, \E)$ and we are interested in $\kappa_{CURC}$ for $x, y \in \V$. Let $\mathcal{M}:=$ the length of the longest shortest path.
        Intuitively, by the definition of the \textit{$\varepsilon$-masked} CURC in Definition \ref{CURC}, we may pick $\varepsilon$ sufficiently small, such that none of the ``virtual edges" are masked, as edge length $\frac{1}{\varepsilon}$ will not be picked in calculating the weighted shortest distance. By picking $\epsilon < \frac{1}{\mathcal{M}}$, the ``virtual edges" have length even more than the longest distance in $\gG$, resulting $\forall x, y \in \V$, $d(x, y)$ is independent of these ``virtual edges".\\
        When $\gG$ is unweighted-undirected, the adjacency matrix $A$ is symmetric. Therefore the Perron-measure $\mathfrak{m}(x)$ for each vertex $x$ is proportional to the inverse degree $\frac{1}{d_x}$. By direct calculation, we have $W(x, y) = \frac{1}{d_x}$ and $W(y, x) = \frac{1}{d_y}$. Therefore, the mean transition distribution $$\mu_x(y) = \frac{1}{2}[\frac{1}{d_x} + \frac{d_y}{d_x \times d_y}] = \frac{1}{d_x},
        $$
        which is equivalent to the initial mass placement in the construction of Ollivier-Ricci Curvature $\kappa_{OR}$. \\
        Since the initial mass distribution according for $\kappa_{CURC}$ and $\kappa_{OR}$ is the same and we choose $\epsilon$ sufficiently small, namely when $\epsilon < \frac{1}{\mathcal{M}}$, the masked and unmasked Wasserstein distances are equal: $\mathcal{W}_1^{\varepsilon}(\mu_x, \mu_y) = \mathcal{W}_1(\mu_x, \mu_y)$. \\
        It is straightforward that the shortest weighted distance $d$ is positively related to $\varepsilon$. Hence $\kappa_{\operatorname{CURC}}^{\varepsilon}$ is a decreasing function in $\varepsilon$. But for a fixed graph $\gG$, $\kappa_{\operatorname{CURC}}^{\varepsilon}$ is invariant when $\varepsilon < \frac{1}{\mathcal{M}}$. Therefore the limit of $\kappa_{\operatorname{CURC}}^{\varepsilon}$ indeed tends to $\kappa_{\operatorname{OR}}$ as $\varepsilon \to 0$. Therefore, $\kappa_{CURC} = \kappa_{OR}$ on connnected unweighted-undirected graphs.
        
        \item \textbf{Proof of Continuity} \\
        To prove the entry-wise continuity of CURC w.r.t. weight matrix $\omega$ under the assumption that perturbation ensures $\omega$ to be non-negative and regular, we use the supremum form of Wasserstein distance from the Kantorovich--Rubinstein duality:
        \begin{equation}
            \mathcal{W}_1^{\varepsilon}\left(\mu_x, \mu_y\right)=\sup _{f \in \operatorname{Lip}(1)} \sum_{z \in \V} f(z)\left(\mu_y(z)-\mu_x(z)\right),
        \end{equation}
        where $\forall x, y \in V, f(y) - f(x) \leq d^{\varepsilon}(x, y)$ as $f \in \operatorname{Lip}(1).$
        We may exploit the limit definition of $\mathcal{W}_1^{\varepsilon}$ and take $\varepsilon$ sufficiently small so that $d^{\epsilon}$ and $\mathcal{W}_1^{\varepsilon}$ is independent of $\varepsilon$, hence we may abuse the notation, denote $d^{\varepsilon}(x, y)$ as $d(x, y)$ and $\mathcal{W}_1^{\varepsilon}(\mu_x, \mu_y)$ as $\mathcal{W}_1(\mu_x, \mu_y)$ for $x, y \in \V$. We first show that $\forall x \neq y \in \V$, $d(x, y)$ is continuous entry-wise w.r.t. $\omega$. Let $\mathcal{M}$ be the diameter w.r.t. edge length as inverse edge weights $\frac{1}{\omega}$, and for edge with $0$ weight we treat the edge length as $+\infty$. Suppose there is a small perturbation of $\delta$ on an arbitrary entry of $\omega$ after which the weight matrix is still non-negative regular, we denote the new weight matrix as $\omega^{\delta}$ and distance function as $d^{\delta}$ on this perturbed matrix. WLOG, we assume the perturbation is smaller than the smallest positive entry and we let $\omega^{min}$ to be this minimal positive entry of $\omega$. With this small perturbation, we have
        \begin{align*}
            |d(x, y) - d^{\delta}(x, y)| &\leq  \frac{1}{\omega^{min} - \delta} - \frac{1}{\omega^{min}}.
        \end{align*}
        Therefore, $\forall \varepsilon_0$, let $\delta \leq \frac{\omega^{2}\varepsilon_0}{1 + \omega \varepsilon_0}$, we have $|d(x, y) - d^{\delta}(x, y)| \leq \varepsilon_0$, hence $d(x, y)$ is continuous entry-wise w.r.t. $\omega$. \\        
        We will prove the continuity of
        \begin{align*}
            &\frac{\mathcal{W}_1(\mu_x, \mu_y)}{d(x, y)} = \sup _{f \in \operatorname{Lip}(1)} \sum_{z \in \V} f(z)\frac{\mu_y(z)-\mu_x(z)}{d(x, y)}, 
        \end{align*}
        which is sufficient for the overall continuity of CURC. By corollary \ref{linear programming}, we have that 
        \begin{align*}
            &\mathcal{W}_1(\mu_x, \mu_y) = \sup _{A f \preceq c} m^T f.
        \end{align*}
        By previous argument, $m$ as a function of mean transition measure $\mu$ and $c$ as a function of distance function $d$ are both continous w.r.t. $\omega$ entry-wise. Therefore, $\mathcal{W}_1(\mu_x, \mu_y)$ is indeed a continuous function w.r.t. $\omega$ entry-wise by lemma \ref{convex continuity extension}. Since $d(x, y)$ is positive and continuous entry-wise, we conclude that $\kappa_{CURC}$ is continuous entry-wise w.r.t. $\omega$.

        \item \textbf{Proof of Scale Invariance} \\
        The third property is straightforward from the construction of the \textbf{reciprocal edge weight} $\mathfrak{r}^{\varepsilon}$ in Definition \ref{reciprocal edge weight}. Suppose $\gG(\V, \E, \omega)$ to be the unscaled strongly-connected weighted-directed graph and  $\gG^{*}(\V, \E, \omega^{*})$ be the scaled one, where $\omega^{*} = \lambda \omega$. Let $\mathcal{M}$ and $\mathcal{M}^{*}$ be the diameter of $\gG$ and $\gG^{*}$ respectively w.r.t. the \textbf{reciprocal edge weight}. Note we are not considering the $\varepsilon$-masked edge weight for computing the diameter, in the sense that for non-exist edges, the edge length is $+\infty$. $\mathcal{M}$ and $\mathcal{M}^{*}$ are well-defined due to strongly-connectivity.\\
        After a scale wof $\lambda$, the eigenvalues of $\omega$ and $\omega^{*}$ differ by a factor of $\lambda$. By Perron-Frobenius theorem, the corresponding Perron-measure $\mathfrak{m}$ and $\mathfrak{m}^{*}$ are the same. Similarly, it is straightforward that the random walk matrices $W$ and $W^{*}$ are also the same. Therefore the initial measure $\mu$ of $\gG$ coincides with $\mu^{*}$ of $\gG^{*}$.\\
        Since $\lambda > 0$, we can find $\epsilon$ small enough, so that $\frac{1}{\epsilon} > \max\{\mathcal{M}, \mathcal{M}^{*}\}$, ensuring the distance function $d^{\varepsilon}$ and $d^{\varepsilon *}$ are independent of $\varepsilon$. Hence, for this choice of $\varepsilon$, $\forall x \neq y \in \V$, $d^{\varepsilon *}(x, y) = \lambda d^{\varepsilon}(x, y)$.
        By Kantorovich--Rubinstein duality \ref{KR duality}, we have that
        \begin{align*}
            \mathcal{W}_1^{\varepsilon *}(x, y) &= \inf _{\pi \in \Pi\left(\mu_x, \mu_y\right)} \sum_{x, y \in V} d^{*}(x, y) \pi(x, y) \\
            &= \inf _{\pi \in \Pi\left(\mu_x, \mu_y\right)} \sum_{x, y \in V} \lambda d(x, y) \pi(x, y) \\
            &= \lambda \mathcal{W}_1^{\varepsilon}(x, y).
        \end{align*}

        Therefore,
        \begin{align*}
            \kappa_{CURC}^{*} &= \lim _{\varepsilon \rightarrow 0} \kappa_{\mathrm{CURC}}^{\varepsilon *}(x, y) \\
            &= \lim_{\varepsilon \rightarrow 0} (1-\frac{\mathcal{W}_1^{\varepsilon *}\left(\mu_x, \mu_y\right)}{d^{\varepsilon *}(x, y)}) \\
            &= \lim_{\varepsilon \rightarrow 0} (1-\frac{\lambda \mathcal{W}_1^{\varepsilon}\left(\mu_x, \mu_y\right)}{\lambda d^{\varepsilon}(x, y)}) \\
            &= \lim_{\varepsilon \rightarrow 0} (1-\frac{ \mathcal{W}_1^{\varepsilon}\left(\mu_x, \mu_y\right)}{d^{\varepsilon}(x, y)}) \\
            &= \kappa_{CURC},
        \end{align*}
        as required. Therefore, CURC is scale invariant.
    \end{enumerate}
    
\end{proof}

\subsection{Continuous Unified Ricci Curvature and Bottlenecking}

To reveal the geometric connection between Continuous Unified Ricci Curvature and propagation graphs, we introduce the concept of \textit{Dirichlet isoperimetric constant} $\mathcal{I}_{\mathcal{V}}^D$, which is the extension of the well-known cheeger constant into strongly-connected weighted-directed graphs. We perceive $\mathcal{I}_{\mathcal{V}}^D$ as a measure of bottlenecking on weighted graphs and state that when CURC has a lower bound $K$, $\mathcal{I}_{\mathcal{V}}^D$ has a lower bound that is positively related to $K$. Our proof draws its foundation from the derivation presented in \citet{ozawa2020geometric}, where they prove the result on strongly-connected weighted-directed graph with unit edge length. Few minor modifications are required to adapt the result to our scenario concerning weighted edge lengths. We will identify and elucidate the key elements that require clarification, while also presenting the results that remain unchanged.

\begin{definition} \textbf{(Chung Laplacian)}
    Let $\gG = (\V, \E, \omega)$ be a finite strongly-connected weighted directd-graph and $f: \V \mapsto \R$. Suppose $\mu: \V \times \V \mapsto [0, 1]$ is a probability kernel satisfying $\sum_{y \in \V} \mu(x, y) = 1 \text{ for all } x \in \V$. 
    The \textit{Chung Laplacian} $\mathcal{L}$ on function $f$ associated with $\mu$ is defined as 
    $$
    \mathcal{L} f(x):=f(x)-\sum_{y \in \V} \mu(x, y) f(y).
    $$
    Let $d: \V \times \V \mapsto \R^{\geq 0}$ be a distance function (asymmetric). For each vertex $x \in \V$, the \textit{asymptotic mean curvature} $\mathcal{H}_x$ is defined by
    $$
    \mathcal{H}_x:=\mathcal{L} \rho_x(x),
    $$
    where $\rho_x: \V \rightarrow \mathbb{R}$ is the distance from $x$ defined as $\rho_x(y):=d(x, y)$. Note that with weighted shortest distance $d$, $\mathcal{H}_x \in (-\infty, 0).$\\
    For each vertex $x \in \V$, the $\operatorname{InRad}_x \V$ of $\V$ at vertex $x$ is defined by 
    $$\operatorname{InRad}_x \V:=\sup _{y \in \V} \rho_x(y),$$
    And for any $x \in \V$ and $R > 0$, we set $E_R(x):=\left\{y \in V \mid \rho_x(y) \geq R\right\}$.
\end{definition}

\begin{definition} \textbf{(Boundary Perron-measure)} \label{Inter Perron measure}
    For a non-empty $\Omega \subset \V$, its \textit{Boundary Perron-measure} is defined as
    $$\mathfrak{m}(\partial \Omega):=\sum_{y \in \Omega} \sum_{z \in V \backslash \Omega} \mathfrak{m}_{y z},$$
    where $\mathfrak{m}_{y z}:= \mathfrak{m}(y) \mu(y, z)$ and $\mathfrak{m}(\Omega) = \sum_{x \in \Omega} \mathfrak{m}(x)$.
\end{definition}

% \begin{definition} \textbf{(Extended Perron measure)} \label{Inter Perron measure}
%     Let $\gG = (\V, \E, \omega)$ be a finite strongly-connected weighted-directed graph, $\mu: \V \times \V \mapsto [0, 1]$ be the mean transition kernel and $\mathfrak{m}: \V \mapsto [0, 1]$ be the Perron-measure. Based on the Perron-measure $\mathfrak{m}$, we define the \textit{boundary Perron measure} by
%     $$\mathfrak{m}(x, y):=\frac{1}{2}(\mathfrak{m}(x) \mu(x, y)+\mathfrak{m}(y) \mu(y, x))=\mathfrak{m}(x) \mu(x, y),$$
%     where we abbreviate $\mathfrak{m}(x, y)$ as $\mathfrak{m}_{xy}$.
% \end{definition}

% Note that $\mathfrak{m}_{x y}=\mathfrak{m}_{y x}$ is symmetric and $\mu(x, y)=\frac{\mathfrak{m}_{x y}}{\mathfrak{m}(x)}$.

% \begin{definition}  \textbf{(Boundary measure)}
%     For a non-empty $\Omega \subset \V$, its boundary measure is defined as
%     $$\mathfrak{m}(\partial \Omega):=\sum_{y \in \Omega} \sum_{z \in V \backslash \Omega} \mathfrak{m}_{y z},$$
%     where $\mathfrak{m}_{y z}$ is defined in \ref{Inter Perron measure} and $\mathfrak{m}(\Omega) = \sum_{x \in \Omega} \mathfrak{m}(x)$.
% \end{definition}

\begin{definition} \textbf{(Dirichlet isoperimetric constant)}
    The \textit{Dirichlet isoperimetric constant} $\mathcal{I}_{\mathcal{V}}^D$ on $\V$ is defined by
    $$\mathcal{I}_{\mathcal{V}}^D:=\inf _{\Omega} \frac{\mathfrak{m}(\partial \Omega)}{\mathfrak{m}(\Omega)},$$
    which is analogous to cheeger constant on weighted-directed graph. 
\end{definition}

Exploiting definition \ref{CURC} on Continuous Unified Ricci Curvature , we introduce a limit version of \textbf{CURC} concerning idleness for theoretical derivation. 

\begin{definition}\textbf{($\alpha$-idle Mean Transition probability Kernel)} \label{idle kernel} 
    Define the  \textit{$\alpha$-idle Mean Transition probability Kernel} by
    \begin{equation}
    \mu_x^{\alpha}(y) = \mu^{\alpha}(x, y):= \begin{cases} \frac{1}{2}\alpha[W(x, y) + \frac{\mathfrak{m}(y)}{\mathfrak{m}(x)} W(y, x)] & \text{if $y \neq x$} \\ \left(1 - \alpha \right) &\text{if $y = x$}\end{cases}
    \end{equation}
\end{definition}

In the following, we use the same \textit{$\varepsilon$-masked reciprocal weighted edge length} for calulating Wasserstein distance.

\begin{definition} \textbf{(Idle-CURC)} \label{Limit version of CURC}
    Define the \textit{$\alpha$-idle $\varepsilon$-masked Continuous Unified Ricci Curvature} and \textit{$\alpha$-idle Continuous Unified Ricci Curvature} by
    \begin{align*}
        \kappa_{\mathrm{CURC}}^{\varepsilon \alpha}(x, y)&:=1-\frac{\mathcal{W}_1^{\varepsilon}\left(\mu^{\alpha}_x, \mu^{\alpha}_y\right)}{d^{\varepsilon}(x, y)},\\
        \kappa_{\mathrm{CURC}}^{\alpha}(x, y) &:=\lim _{\varepsilon \rightarrow 0} \frac{\kappa_{\mathrm{CURC}}^{\varepsilon \alpha}(x, y)}{\alpha}
    \end{align*}
    The \textit{idle-CURC} is defined by 
    $$
    \kappa_{\mathrm{CURC}}^{I}(x, y):=\lim _{\varepsilon \rightarrow 0} \lim_{\alpha \rightarrow 0} \frac{\kappa_{\mathrm{CURC}}^{\varepsilon \alpha}(x, y)}{\alpha}.
    $$
\end{definition}

\begin{theorem}\label{Concavity and geq}
    Let $\gG=(\V, \E, \omega)$ be a strongly-connected weighted-directed graph, we have that for all $x \neq y \in \V$,
    $$
        \kappa_{\mathrm{CURC}}^{I}(x, y) \geq \kappa_{\mathrm{CURC}}(x, y).
    $$
\end{theorem}

\begin{proof}
    Using the trick mentioned in the proof of \ref{Curvature property}, we choose $\varepsilon$ sufficiently small such that $\kappa_{\mathrm{CURC}}^{\varepsilon \alpha}(x, y)$ is independent of $\varepsilon$. WLOG, we abbreviate $\kappa_{\mathrm{CURC}}^{\varepsilon \alpha}(x, y)$ as $\kappa_{\mathrm{CURC}}^{\alpha}(x, y)$. By lemma 3.2 from \citet{ozawa2020geometric}, $\kappa_{\mathrm{CURC}}^{\alpha}(x, y)$ is concave in $\alpha \in [0, 1]$ and $\frac{\kappa_{\mathrm{CURC}}^{\alpha}(x, y)}{\alpha}$ is non-increasing in $\alpha \in (0, 1]$. Note that $\kappa_{\mathrm{CURC}}(x, y)$ is nothing but $\frac{\kappa_{\mathrm{CURC}}^{\alpha}(x, y)}{\alpha}$ taking $\alpha = 1$. Therefore by monotonicity, $\kappa_{\mathrm{CURC}}^{I}(x, y) \geq \kappa_{\mathrm{CURC}}(x, y)$. 
\end{proof}

\begin{proposition}\label{Green's theorem}
    Let $\Omega \subset \V$ be a non-empty subset. Then for all function $f_0, f_1: \V \rightarrow \R$,
    $$
    \begin{aligned}
    \sum_{x \in \Omega} \mathcal{L} f_0(x) f_1(x) \mathfrak{m}(x)= & \frac{1}{2} \sum_{x, y \in \Omega}\left(f_0(y)-f_0(x)\right)\left(f_1(y)-f_1(x)\right) \mathfrak{m}_{x y} \\
    & -\sum_{x \in \Omega} \sum_{y \in V \backslash \Omega}\left(f_0(y)-f_0(x)\right) f_1(x) \mathfrak{m}_{x y} .
    \end{aligned}
    $$
\end{proposition}

\begin{proof}
    The proof is a merely a calculation similar to integration by part. We stress that the result only depends on fact that $\mathcal{m}$ is symmetric. (c.f. Theorem 2.1 in \citet{grigor2018introduction})
\end{proof}

\begin{lemma}
Let $x, y \in \V$ with $x \neq y$. Then 
$$\quad \kappa_{\mathrm{CURC}}^{\alpha}(x, y)=\inf _{f \in \operatorname{Lip}_1(\V)}\left(\frac{1}{\alpha}\left(1-\nabla_{x y} f\right)+\nabla_{x y} \mathcal{L} f\right),$$
where $\nabla_{x y} f:=\frac{f(y)-f(x)}{d(x, y)}$.
\end{lemma}

\begin{proof}
    We refer to lemma 3.9 in \citet{ozawa2020geometric}, which is essentially similar. It is worth noticing that we are using a different weighted distance function, but there is no restriction on the distance function in the proof.
\end{proof}

\begin{proposition}\label{Laplacian with curvature}
    Let $x, y \in \V$ with $x \neq y$. Suppose $\mathcal{F}_{x y}:=\left\{f \in \operatorname{Lip}_1(\V) \mid \nabla_{x y} f=1\right\}$. Then we have
    $$
    \kappa_{\mathrm{CURC}}^{I}(x, y)=\inf _{f \in \mathcal{F}_{x y}} \nabla_{x y} \mathcal{L} f .
    $$
\end{proposition}

\begin{proof}
    We refer to theorem 3.10 in \citet{ozawa2020geometric}, where the only part worth mentioning is that we require 
    $$\operatorname{Lip}_{1, x}(V):=\left\{f \in \operatorname{Lip}_1(V) \mid f(x)=0\right\},$$
    to be compact w.r.t. the canonical topology on $\R^{n}$. Let $\omega^{*}:= \inf_{x, y \in \V} \omega(x, y) > 0$, then $d(x, y) \leq \frac{n}{\omega^{*}}$, which is bounded for fixed weight matrix $\omega$. As we restrict $f(x) = 0$, the 1-Lipshitz function $f$ w.r.t. the weighted shortest distance function $d$ is still bounded. Therefore, changing the distance function does not break compactness w.r.t. $\R^{n}$.
\end{proof}

\begin{theorem}\label{theorem 1.1}
    Let $x \in V$. For $K \in \mathbb{R}$ we assume $\inf _{y \in V \backslash\{x\}} \kappa_{\mathrm{CURC}}^{I}(x, y) \geq K$. For $\Lambda \in(-\infty, 0)$ we further assume $\mathcal{H}_x \geq \Lambda$. Then on $V \backslash\{x\}$, we have
    $$
    \mathcal{L} \rho_x \geq K \rho_x+\Lambda
    $$
\end{theorem} 

\begin{proof}
    Fix a vertex $x \in \V$. Note that the distance function $\rho_x \in \mathcal{F}_{x y}$ defined in proposition \ref{Laplacian with curvature}, as $\nabla_{x y} \rho_x=\frac{\rho_x(y) - \rho_x(x)}{d(x, y)}=1$. Hence, for all $y \in \V \backslash \{x\}$,
    $$
    K \leq \kappa_{CURC}^{I}(x, y) \leq \nabla_{x y} \mathcal{L} \rho_x=\frac{\mathcal{L} \rho_x(y)-\mathcal{L} \rho_x(x)}{d(x, y)} \leq \frac{\mathcal{L} \rho_x(y)-\Lambda}{d(x, y)}.
    $$
    When $y = x$, the result is direct from $\mathcal{H}_x \geq \Lambda$. Hence, $\mathcal{L} \rho_x \geq K \rho_x+\Lambda$.
\end{proof} 

\begin{theorem}\label{weighted cheeger constant proof}
Let $x \in V$. For $K \in \mathbb{R}$ we assume $\inf _{y \in V \backslash\{x\}} \kappa_{\mathrm{CURC}}(x, y) \geq K$. For $\Lambda \in$ $(-\infty, 0)$ we also assume $\mathcal{H}_x \geq \Lambda$. For $D>0$ we further assume $\operatorname{InRad}_x V \leq D$. Then for every $R > 0$ with $K R+\Lambda>0$, we have
$$
\mathcal{I}_{E_R(x)}^D \geq \frac{K R+\Lambda}{D}
$$
\end{theorem}

\begin{proof}
The proof is analogue to Proposition 9.6 in \citet{ozawa2020geometric}, while we include the proof utilizing previous results for completeness.
By theorem \ref{Concavity and geq}, $\kappa_{\mathrm{CURC}}(x, y) \geq K$ implies $\kappa_{\mathrm{CURC}}^{I}(x, y) \geq K$. Let $\Omega \subset E_R(x)$ be a non-empty vertex set. By Proposition \ref{Green's theorem}, we have
\begin{align*}
    -\sum_{y \in \Omega} \mathcal{L} \rho_x(y) \mathfrak{m}(y)&=\sum_{y \in \Omega} \sum_{z \in V \backslash \Omega}\left(\rho_x(z)-\rho_x(y)\right) \mathfrak{m}_{y z} \\
    &\geq -\sum_{y \in \Omega} \sum_{z \in V \backslash \Omega} \rho_x(y) \mathfrak{m}_{y z} \\ 
    &\geq-D \mathfrak{m}(\partial \Omega)
\end{align*}
By theorem \ref{theorem 1.1}, for all $y \in \Omega$,
$$\mathcal{L} \rho_x(y) \geq K \rho_x(y)+\Lambda \geq K R+\Lambda$$
Therefore,
\begin{align*}
    &\sum_{y \in \Omega} [\mathcal{L} \rho_x(y) - (K R+\Lambda)] \mathfrak{m}(y) \geq 0 \\
    \Rightarrow &\sum_{y \in \Omega} \mathcal{L} \rho_x(y) \mathfrak{m}(y) \geq \sum_{y \in \Omega} KR \mathfrak{m}(y) = (KR + \Lambda)\mathfrak{m}(\Omega).
\end{align*} 
Note that we have $\sum_{y \in \Omega} \mathcal{L} \rho_x(y) \mathfrak{m}(y) \leq D \mathfrak{m}(\partial \Omega)$, which combined together yields
\begin{align*}
    &D \mathfrak{m}(\partial \Omega) \geq (KR + \Lambda)\mathfrak{m}(\Omega) \\
    \Rightarrow &\frac{\mathfrak{m}(\partial \Omega)}{\mathfrak{m}(\Omega)} \geq \frac{(KR + \Lambda)}{D} \\
    \Rightarrow &\mathcal{I}_{E_R(x)}^D \geq \frac{K R+\Lambda}{D}
\end{align*}
\end{proof}

Before we end the discussion of properties of Continuous Unified Ricci Curvature, we point out that computing optimal-transportation based graph curvature can be computational intensive from solving a Linear programming problem similar to corollary \ref{linear programming}. In \citet{topping2022UnderstandingOversquashingBottlenecks} they provide a lower bound estimation for the Ollivier-ricci graph curvature. Likewise, we provide a universal lower bound for \text{CURC} using Kantorovich-Rubinstein duality and a tighter lower bound for \text{CURC} under stronger assumptions.

\begin{proposition}\label{lower bound 1}
    Based on the construction of CURC, we choose $\varepsilon$ sufficiently small s.t. distance function $d^{\varepsilon}$ is independent of $\varepsilon$ and denote it as $d$. Let $\mathcal{D}(x, y) := \max\{d(x, y), d(y, x)\}$ be the largest weighted distance between vertices $x$ and $y$. For distinct vertices $x, y \in \V$, we have
    \begin{align*}
        \kappa_{\mathrm{CURC}}(x, y) &\geq -\frac{2 \mathcal{D}(x, y)}{d(x, y)}(1-\mu(x, y)-\mu(y, x))_{+}+\frac{1}{d(x, y)}(d(x, y)+\mathcal{D}(x, y)-\mathcal{H}(y, x)) \\
        &-\frac{\mathcal{D}(x, y)-d(y, x)}{d(x, y)}(\mu(x, y)+\mu(y, x)),
    \end{align*}
    where $\mathcal{H}(x, y)$ is defined by 
    \begin{align*}
        \widetilde{\mathcal{H}}(x) &:=  -\sum_{y \in V} \mu(x, y) d(y, x) \\
        \mathcal{H}(x) &:=  -\sum_{y \in V} \mu(x, y) d(x, y) \\
        \mathcal{H}(x, y) &:= - \sum_{y \in \V} \mu(x, y) d(x, y) - \sum_{y \in \V} \mu(x, y) d(y, x) = \widetilde{\mathcal{H}}(x) + \mathcal{H}(x).
    \end{align*}
\end{proposition}

\begin{proof}
    The proof is a simple calculation of the Kantorovich-Rubinstein duality and we present the proof along the line of Proposition 6.1 in \citet{ozawa2020geometric}.
    By theorem \ref{KR duality}, we have
    \begin{align*}
        \mathcal{W}_1(\mu_x, \mu_y) &= \sup_{f \in \operatorname{Lip}(1)} \sum_{z \in \V} f(x)(\mu_x(z) - \mu_y(z)) \\
        &= \sup_{f \in \operatorname{Lip}(1)} \Bigg\{\left(\sum_{z \in V \backslash\{x\}}(f(z)-f(y))\mu(y, z)\right) - \left(\sum_{z \in V \backslash\{y\}}(f(z)-f(x)) \mu(x, z)\right) \\
        &+ \left(f(y)-f(x)\right)\left(1-\mu(x, y)-\mu(y, x)\right)\Bigg\}.
    \end{align*}    
    For an arbitrary function $f \in \operatorname{Lip}(1)$ w.r.t. $d$, we have that
    $$
        f(z) - f(y) \leq d(y, z), \quad f(z) - f(x) \geq -d(z, x), \quad |f(y) - f(x)| \leq \mathcal{D}(x, y), 
    $$
    Therefore, 
    \begin{align*}
    \mathcal{W}_1\left(\mu_x, \mu_y \right) \leq & \sum_{z \in V \backslash\{x\}} d(y, z) \mu(y, z)+\sum_{z \in V \backslash\{y\}} d(z, x) \mu(x, z) \\
    & +\mathcal{D}(x, y)|1-\mu(x, y)-\mu(y, x)| \\
    = & \left(-\mathcal{H}_y-d(y, x) \mu(y, x)\right)+\left(-\overleftarrow{\mathcal{H}}_x-d(y, x) \mu(x, y)\right) \\
    & +\mathcal{D}(x, y)\left(2(1-\mu(x, y)-\mu(y, x))_{+}-(1-\mu(x, y)-\mu(y, x))\right) \\
    = & \mathcal{H}(y, x)-d(y, x)(\mu(x, y)+\mu(y, x)) \\
    & +\mathcal{D}(x, y)\left(2(1-\mu(x, y)-\mu(y, x))_{+}-(1-\mu(x, y)-\mu(y, x))\right) \\
    = & 2 \mathcal{D}(x, y)(1-\mu(x, y)-\mu(y, x))_{+}-(\mathcal{D}(x, y)-\mathcal{H}(y, x)) \\
    & +(\mathcal{D}(x, y)-d(y, x))(\mu(x, y)+\mu(y, x)).
    \end{align*}
    Hence, 
    \begin{align*}
        \kappa_{CURC}(x, y) &= 1 - \frac{\mathcal{W}_1(\mu_x, \mu_y)}{d(x, y)} \\
        & \geq 1 - 2 \frac{\mathcal{D}(x, y)}{d(x, y)}(1-\mu(x, y)-\mu(y, x))_{+}-\frac{1}{d(x, y)}(\mathcal{D}(x, y)-\mathcal{H}(y, x)) \\
        & + \frac{(\mathcal{D}(x, y)-d(y, x))}{d(x, y)}(\mu(x, y)+\mu(y, x)) \\
        &= -\frac{2 \mathcal{D}(x, y)}{d(x, y)}(1-\mu(x, y)-\mu(y, x))_{+}+\frac{1}{d(x, y)}(d(x, y)+\mathcal{D}(x, y)-\mathcal{H}(y, x)) \\
        &-\frac{\mathcal{D}(x, y)-d(y, x)}{d(x, y)}(\mu(x, y)+\mu(y, x)).
    \end{align*} 
\end{proof}

\begin{remark}
    To calculate this lower bound, after pre-processing distance function $d$ and the mean transition probability, the asymptotic complexity is $\mathcal{O}(n)$ for each vertex pair $x \neq y$. The pre-processing inclues a Floyd-Washall algorithm for shortest path which is $\mathcal{O}(n^3)$ and a power method for computing perron eigenvectors which is conventionally $\mathcal{O}(n^2)$. Therefore, the overall computational complexity is $\mathcal{O}(n^3)$
\end{remark}

When strongly-connected $\gG=(\V, \E, \omega)$ satisfies $(x, y) \in \E \iff (y, x) \in \E$, and we use edge length $1$ for computing the shortest distance, CURC degenerates to the curvature definition in \citet{ozawa2020geometric}. Under this stronger assumption, we can achieve a tighter bound for $\kappa_{CURC}$ with a specific transportation plan utilizing the topology of local neighborhoods.

\begin{definition}
    For $x \sim y$, we define
    \begin{itemize}
        \item $\overrightarrow{\mathcal{N}}_x:=\{y \in V \mid x \rightarrow y\}, \overleftarrow{\mathcal{N}}_x:=\{y \in V \mid y \rightarrow x\}, \mathcal{N}_x:=\overrightarrow{N}_x \cup \overleftarrow{\mathcal{N}}_x$, which are \textit{inner neighborhood}, \textit{outer neighborhood} and \textit{neighborhood} respectively. If $\forall x, y \in \V$, $x \to y$ implies $y \to x$, then $\overrightarrow{\mathcal{N}}_x = \overleftarrow{\mathcal{N}}_x = \mathcal{N}_x$. Hence we use notation $\mathcal{N}_x$ to denote neighborhood for $x$.
        \item ${\Delta}(x, y):=\mathcal{N}_x \cap \mathcal{N}_y$ denotes set of common neighbors of vertices $x$ and $y$.
        \item $\square(x, y):=\left\{z \in \mathcal{N}_x \backslash \mathcal{N}_y, z \neq y: \exists w \in\left(\mathcal{N}_z \cap \mathcal{N}_y\right) \backslash \mathcal{N}_x \right\}$, which denotes the neighbors of $x$ forming 4-cycle based at $x \sim y$ without diagonals inside.
        \item ${\square}^{\mathfrak{m}}(x, y):=\max \left\{|U|: U \subseteq \square(x, y), \exists \varphi: U \rightarrow \square(y, x), \varphi \in \mathcal{D}(U)\right\}$, and we use $\varphi^{\mathfrak{m}}$ to denote one such optimal pairing between $\square(x, y)$ and $\square(y, x)$.
    \end{itemize}
\end{definition}

\begin{proposition}\label{lower bound 2}
On a strongly-connected weighted-directed locally graph $G$=$(\V, \E, \omega)$, where $\forall x, y \in \V$, if $x \to y \iff y \to x$, we have that if $x \sim y$, then 
    \begin{align*} 
    \kappa_{CURC}(x, y) \geq & -\left(1-\mu(x, y)-\mu(y, x)-\sum_{z \in \Delta(x, y)} \mu(x, z) \vee \mu(y, z) - \sum_{z \in \square(x, y)} \mu(x, z) \wedge \mu(y, {\varphi}^{\mathfrak{m}}(z)) \right)_{+} \\ 
    & -\left(1-\mu(x, y)-\mu(y, x)-\sum_{z \in \Delta(x, y)} \mu(x, z) \wedge \mu(y, z) - \sum_{z \in \square(x, y)} \mu(x, z) \wedge \mu(y, {\varphi}^{\mathfrak{m}}(z)) \right)_{+} \\
    &+\sum_{z \in \Delta(x, y)} \mu(x, z) \wedge \mu(y, z).
    \end{align*}
\end{proposition}

\begin{remark}
    When $\gG = (\V, \E, \omega)$ satisfies $x \to y \iff y \to x$, optimal transportation based curvature of $x \sim y$ is only dependent up to cycles of size at most 5. In theorem 6 of \citet{jost2014ollivier}, they give a bound concerning the influence of triangles and in theorem 2 of \citet{topping2022UnderstandingOversquashingBottlenecks}, they extend the result concerning cycles of size 4. The proof of this lower bound is in line with these two theorems and we give a tighter lower bound for $\kappa_{CURC}$ concerning the influence of 4-cycles compared to proposition \ref{lower bound 1} under this stronger assumption. The computational cost for computing this lower bound is at most $\mathcal{O}(n^4)$. We stress that under specific user case for CURC which requires lower computational cost, the lower bound estimations for $\kappa_{CURC}$ from proposition \ref{lower bound 1} and proposition \ref{lower bound 2} become handy.
\end{remark}

\clearpage
\clearpage
\section{Expressiveness of GPNNs}
% \todo[inline]{Double check notation and typo}
\subsection{Color refinement(CR)} 
\label{CR}
% Given a graph, $\gG=(\V,\E)$, the 1-dimensional Weisfeiler-Lehman algorithm (1-WL), also known as the color-refinement algorithm, iteratively computes a color mapping $\gX_{\gG}: \V \mapsto \gC$, where $\gC$ is the color set, which assigns each node a color $\gX_{\gG}(v) \in \gC$. Initially, every vertex is assigned the same color. In each subsequent iteration, the 1-WL algorithm updates the color of each vertex by amalgamating its current color with the colors of its neighboring vertices, employing a hash function. This process is reiterated for a sufficiently large number of iterations $T$, typically $T=|\V|$.

The 1-dimensional Weisfeiler-Lehman algorithm (1-WL), also referred to as the color-refinement algorithm, operates iteratively to determine a color mapping \( \mathcal{X}_{\mathcal{G}}: \mathcal{V} \mapsto \mathcal{C} \) for a given graph \( \mathcal{G} = (\mathcal{V}, \mathcal{E}) \), where \( \mathcal{C} \) represents the set of colors. Every vertex is initially assigned an identical color. During each ensuing iteration, a hash function is utilized by the 1-WL algorithm to amalgamate the current color of each vertex with the colors of its adjacent vertices, thereby updating the vertex's color. The algorithm persists in this process for a substantial number of iterations \( T \), typically set to \( T = |\mathcal{V}| \).

\subsection{Prototypical GPNNs}

It is well known that most of the MPNNs have an expressiveness upper bound of 1-WL \cite{xu2019HowPowerfulAre, morris2019WeisfeilerLemanGo}. For GPNNs, the situation is a bit different due to the fact that we have the propagation depends on adjacency function $F(\mathbf{A})$, which is any permutation equivariant function.

Here we set two prototypical GPNNs to analyze their expressiveness.  First, we consider the GPNNs with homogeneous adjacency features across all heads and layers and refer to them as \textbf{Static GPNNs}.
\begin{equation}
    F = F^{l}
\end{equation}
For generality, we also consider the case when multiple adjacency function $F$ is used in the model. An extra prototype model is defined with recurrence as a special case of \textbf{Dynamic GPNNs}
\begin{equation}
\begin{aligned}
    F^l &= F^{l+p}
\end{aligned}
\end{equation}
In order to reach the upper-bounded expressiveness, the model would be assumed to have a sufficient number of heads and two MLPs with a sufficient layer and width for the connectivity function and update function.

\begin{lemma}\label{xulemma} \textbf{\cite{xu2019HowPowerfulAre}, Lemma 5} If we assume that the set \( \gX \) is countable, a function \( f : \gX \mapsto \R^n \) can be established such that each bounded-size multiset \( \hat{\mathcal{X}} \subset \mathcal{X} \) has a unique corresponding function \( h(\hat{\mathcal{X}}) := \sum_{x \in \hat{\mathcal{X}}} f(x) \). Additionally, a decomposition of any multiset function g can be represented as \( g(\hat{\mathcal{X}}) = \phi\left( \sum_{x \in \hat{\mathcal{X}}} f(x) \right) \) for some function \( \phi \).
\end{lemma}

\begin{proposition} (\textbf{Static GPNNs}) \label{homogeneous-gpnn}
Suppose GPNN model M has a a fixed adjacency feature $[f_{uv}]_{u,v\in \cV}$, with sufficient heads and layers, the expressiveness of M is upper-bounded by the iterative color-refinement
\begin{equation}\label{GDWL_to_proof}
    \mathcal{X}^{t+1}_{\gG}(v) = \text{hash} \{\{ (\mathcal{X}^t_{\gG}(u), f_{vu}): u \in \V \}\}
\end{equation}\end{proposition}
\begin{proof} \label{homogeneous-gpnn-proof}
First, we prove that for any $f_{uv}$ there exists a function $\pi_\textit{base}$ which is invective from each entry of  $f_{uv}=[F(A)]_{uv}$, to $\R$. Here we define the set of all possible values of 
\begin{equation}
    F_n:=\{[F(A)]_{uv}: A=\text{adj}(\gG),\gG=(\V, \E),|\V|\leq n, (v, u) \in \V^2\}
\end{equation}
For all graphs with no more than $n$ nodes, the total number of possible values of $f_{vu} \in \R^d$ is finite and depends on $n$ and $F$, denoted as $|F_n|=N$.  Given arbitrary bijection $\operatorname{id}: F_n \mapsto [N]$,
By the Stone–Weierstrass Theorem applied to the algebra of continuous functions $C(\R^d, \R)$ there exists a polynomial $\pi_\textit{base}$ so that 
\begin{equation}
    \pi_{\textit{base}}(\text{f}) = \operatorname{id}(\text{f}),~\text{for any } \text{f} \in F_n
\end{equation}
Now, we are ready to construct the $\pi^h$ using $\pi_\textit{base}$ combined with the indicator function $\pi_1^h(d):=\mathbb{I}\left(d=h\right)$ . For each head, we have $\pi^h=\pi^h_1 \circ \pi_\textit{base}$. By multiplying $\mP$ with $\phi(\mX)$, we can recover the color-refinement of
\begin{equation} \label{GPWL}
    \begin{aligned}
        \begin{aligned}
\chi_\gG^l(v) & :=\operatorname{hash}\left(\left(\chi_\gG^{l, 1}(v), \chi_\gG^{l, 2}(v), \cdots, \chi_\gG^{l,\left|F_n\right|}(v)\right)\right), \\
\text { where } \chi_\gG^{l, h}(v) & :=\left\{\left\{\chi_\gG^{l-1}(u): u \in \mathcal{V}, \operatorname{id}(f_{vu})=h \right\}\right\} .
\end{aligned}
    \end{aligned}
\end{equation}
In detail: by applying \textbf{Lemma}~\ref{xulemma} to matrix multiplication, we fulfill the injective multiset function in \ref{GPWL}.
\begin{equation}
    \mX^{l,h}_v = \sum_{u \in \{\pi^h(f(A)_{v,u}) = 1 \}} \phi(\mX^l_u)
\end{equation}
By concatenating all the heads (injective) and passing them to an \textbf{MLP} update function, we can fulfill the $\text{hash}$ function in \ref{GPWL}.
Before conclusion, we refer to the universal approximation theorem of \textbf{MLPs} \cite{hornik1989multilayer} to validate the use of \textbf{MLPs} to approximate constructed functions.
Finally, it's easy to see that the color refinement in \ref{GPWL} is identical to \ref{GDWL_to_proof}
\end{proof}

\begin{proposition} (\textbf{Dynamic GPNNs})
Suppose GPNN model M has a layer-dependent  propagation function $F^l(\mathbf{A})$, repeats every $p$ layer: $F^l = F^{l+p}$. With sufficient heads and layers, by stacking repetitions of such repetition, the expressiveness of M is upper-bounded by the iterative color refinement
\begin{equation} \label{combined_coloring}
    \mathcal{X}^{t+1}_{\gG}(v) = \text{hash} \{\{  \left(\mathcal{X}^t_{\gG}(u), \left(f^1_{vu}, f^2_{vu}, \ldots, f^p_{vu} \right) \right): u \in \V \}\}
\end{equation}
\end{proposition}
In order to prove this, we would like to introduce two concepts for general CR algorithms, the stable colormap and the partition of a colormap. 
For any CR algorithm, at each iteration, the color mapping $\chi_{\gG}^t$ induces a partition of the vertex set $\V$ with an equivalence relation $\sim_{\chi_\gG^t}$ defined to be $u \sim_{\chi_\gG^t} v \Leftrightarrow \chi_\gG^t(u) = \chi_\gG^t(v)$ for $u, v \in V$. We call each equivalence class a color class with an associated color $c \in C$, denoted as $(\chi_\gG^t)^{-1}(c) := \{v \in V : \chi_\gG^t(v) = c\}$. Formally, we define the partition of any color mapping $\chi_\gG$
\begin{definition}
    \textbf{(Partition)} The partition corresponding to $\chi_\gG$ is the set $P(\chi_\gG) = \{\chi_\gG^{-1}(c): c\in \gC_\gG\}$, where $\gC_\gG := \{\chi_\gG : v \in V\}$. More specifically, if any element in $P(\chi_\gG^1)$ is a subset of some element in $P(\chi_\gG^2)$, we say that $P(\chi_\gG^1)$ is at least as fine as $P(\chi_\gG^2)$
\end{definition}
It's easy to see due to the $\operatorname{hash}$ function, any color refinement iteration refines the partition $P(\chi^t_\gG)$ to a finer partition $P(\chi^{t+1}_\gG)$. Since the number of vertices $|V| \leq n$, there must exist an iteration $T < |V|$ such that $P(\chi^{T}_\gG) = P(\chi^{T+1}_\gG)$.  Formally, we define the stable color mapping and stable partition
\begin{definition}
    \textbf{(Stable partition and stable color mapping)} Given a graph $\gG=(\V, \E)$, and an CR refinement $C \in \operatorname{End}(\operatorname{Hom}(\V, \gC))$. Starting from $\chi^0_\gG=c_0$ (a constant initial color mapping), $\chi^{t+1}_\gG = C(\chi^{t}_\gG)$. There exist an iteration $T<|\V|$ , such that $P(\chi^{T}_\gG) = P(\chi^{T+1}_\gG)$. Such $P(\chi^{T}_\gG)$ is called a stable partition denoted as $P_\text{stable}(C)$. Furthermore, we use $\chi_\gG(C)$ to represent one of the many $\chi^{T'}_\gG$ with $T'\geq T$, namely the stable color mapping. 
\end{definition}

CR algorithms decide if the graph pair $(\gG, \gH)$ is isomorphic by comparing the color mapping $\chi^{T}_\gG$ and $\chi^{T}_\gH$. If the stable partition of CR iteration $C^1$ is finer than $C^2$ for any graphs with finite nodes, we can conclude that $C^1$ is more powerful than $C^2$.  We refer to \cite{zhang2023rethinking} for more detail. 

\begin{proof} \label{layer-recurrent-gpnn-proof}
Given \textbf{Proposition}~\ref{homogeneous-gpnn}, we know that each GPNN layer $l \in [L]$ with $\hat{l} := l \bmod p$ can (under sufficient layers and width conditions) fulfill the coloring process of
\begin{equation}
    \mathcal{X}^{l+1}_{\gG}(v) = \text{hash} \{\{ (\mathcal{X}^l_{\gG}(u), f^{\hat{l}}_{vu}): u \in \V \}\}
\end{equation}
We simplify the notation of this color refinement iteration by $C^{\hat{l}} \in \operatorname{End}(\operatorname{Hom}(\V, \gC))$
\begin{equation}
\gX^{l+1}_{\gG} = C^{\hat{l}}(\gX^{l}_{\gG})    
\end{equation}
Now, we define the color refinement iteration for a full recurrent period $p$ for $l=kp, k\in\mathbb{N}$, 
\begin{equation} \label{period_coloring}
    \mathcal{X}^{l+p}_{\gG} = C^p \circ \cdots \circ C^1 \circ C^0(\mathcal{X}^{l}_{\gG})
\end{equation}
Our goal is to show that the combined color refinement $C_{\text{comb}}=C^p \circ \cdots \circ C^1 \circ C^0$ is as powerful as the color refinement in \ref{combined_coloring} denoted as $C_\text{concat}$. In order to achieve that, we will compare the stable partition $P(C_\text{comb})$ and $P(C_\text{concat})$ on an arbitrary graph $\gG=(\V, \E)$.
For stable coloring $\chi_\gG(C_\text{comb})$ and $\chi_{\gG}(C_\text{concat})$. For $v_1, v_2 \in \V$, we will prove:
\begin{equation} \label{stable_partition}
    \chi_\gG(C_\text{comb})(v_1) = \chi_\gG(C_\text{comb})(v_2) \Leftrightarrow \chi_\gG(C_\text{concat})(v_1) = \chi_\gG(C_\text{concat})(v_2)
\end{equation}
From the left to right, since the stable partition is unique and denoted as $P_\gG(\chi_\gG(C_\text{comb}))$.  We have: \begin{equation}
    \chi_\gG(C_\text{comb})(v_1) = \chi_\gG(C_\text{comb})(v_2) \Leftrightarrow \exists S \in P_\gG(\chi_\gG(C_\text{comb})),~\text{s.t.}~v_1, v_2 \in S
\end{equation}
Thus we have$P_\gG(\chi_\gG(C_\text{comb})) = P_\gG(C^1\circ \chi_\gG(C_\text{comb})) = \cdots = P_\gG(C^p \circ \cdots \circ C^1\circ\chi_\gG(C_\text{comb})), \text{with}~ v_1, v_2 \in S ~\text{an element of each of the partitions.}$
Which is equivalent to $C^i(\chi_\gG(C_\text{comb}))(v_1) = C^i(\chi_\gG(C_\text{comb}))(v_2), \text{for}~ i \in [p].$
Write it with $\operatorname{hash}$ function notation we have $\operatorname{hash}\{\{\left(\chi_\gG(C_\text{comb})(u), F^i(A_{\gG})_{v_1u}\right), u \in \V\}\} = \operatorname{hash}\{\{\left(\chi_\gG(C_\text{comb})(u), F^i(\mathbf{A}_{\gG})_{v_2u}\right), u \in \V\}\}, \text{for}~ i \in [p].$
Thus we can infer that
\begin{align*}
\operatorname{hash}\{\{\left(\chi_\gG(C_\text{comb})(u), (F^0(\mathbf{A}_{\gG})_{v_1u}, F^1(\mathbf{A}_\gG)_{v_1u}, \ldots, F^p(\mathbf{A}_\gG)_{v_1u} )\right), u \in \V\}\} \\= \operatorname{hash}\{\{\left(\chi_\gG(C_\text{comb})(u), (F^0(\mathbf{A}_{\gG})_{v_2u}, F^1(\mathbf{A}_\gG)_{v_2u}, \ldots, F^p(\mathbf{A}_\gG)_{v_2u} )\right), u \in \V\}\}
\end{align*}
Recall on the right-hand side, that the $\operatorname{hash}$ notation of $C_\text{concat}$ is
\begin{align*}
\operatorname{hash}\{\{\left(\chi_\gG(C_\text{concat})(u), (F^0(\mathbf{A}_{\gG})_{v_1u}, F^1(\mathbf{A}_\gG)_{v_1u}, \ldots, F^p(\mathbf{A}_\gG)_{v_1u} )\right), u \in \V\}\} \\= \operatorname{hash}\{\{\left(\chi_\gG(C_\text{concat})(u), (F^0(\mathbf{A}_{\gG})_{v_2u}, F^1(\mathbf{A}_\gG)_{v_2u}, \ldots, F^p(\mathbf{A}_\gG)_{v_2u} )\right), u \in \V\}\}
\end{align*}
Thus $P(C_\text{comb})$ is at least as fine as $P(C_\text{concat})$.
By which we prove the $\chi_\gG(C_\text{comb})(v_1) = \chi_\gG(C_\text{comb})(v_2) \Rightarrow \chi_\gG(C_\text{concat})(v_1) = \chi_\gG(C_\text{concat})(v_2)$

It is straightforward from left to right since the $\operatorname{hash}$ notation of $C_\text{concat}$ implies each of the $C^i$ iterations holds. We have $\chi_\gG(C_\text{comb})(v_1) = \chi_\gG(C_\text{comb})(v_2)  \Leftrightarrow \chi_\gG(C_\text{concat})(v_1) = \chi_\gG(C_\text{concat})(v_2)$.
By proving \ref{stable_partition}, we conclude that $C_\text{comb}$ is at least as fine as $C_\text{concat}$ and vice versa, thus $C_\text{comb}$ is as powerful as $C_\text{concat}$.

Substituting $[F(A\gG)]_uv$ with fun, we finish our proof.

\end{proof}

\clearpage
\clearpage
\section{Background}

\subsection{MPNNs} \label{appendix: background-MPNN}
The message-passing framework ~\cite{gilmer2017NeuralMessagePassing} encapsulates a family of models by defining the message function and update functions. An $l$-th layer performs the following update:
\begin{equation} \label{eq:mpnns}
    \begin{aligned}
    m_v^{l+1} & =\sum_{u \in N(v)} M_l\left(h_v^l, h_u^l, e_{v u}\right) \\
    h_v^{l+1} & =U_t\left(h_v^l, m_v^{l+1}\right)
    \end{aligned}   
\end{equation}
An additional readout function is applied in the final layer, $h_v^{L}$. The defining component of the framework is the summation over set $N(v)$, denoting the neighbors of vertex $v$. The original MPNN ~\cite{gilmer2017neural} intended to be broad and thus defines the message function as any function depending on the hidden vertices and edge feature. Despite that broader MPNN family recently extends beyond Eq. \ref{eq:mpnns} \cite{velickovic2018GraphAttentionNetworks, bresson2018ResidualGatedGraph}, it can still be handled by adding additional terms or normalization factors. In MPNNs, the message is passed according to the connectivity of the input graph so that the structural information of the graph will be collected.

\subsection{Graph Transformers}
Instead of being restricted by the input graph, transformers propagate information by attending to the vertex features. Graph transformers extend self-attention formulated as, 
\begin{equation}
\operatorname{Attn}(\mathbf{X}):=\operatorname{softmax}\left(\frac{\mathbf{Q K}^T}{\sqrt{d_{\text {out }}}}\right) \mathbf{V} \in \mathbb{R}^{n \times d_{\text {out }}}
\end{equation}
Where softmax is applied on each row of the normalized similarity matrix.  $\mathbf{V}$, $\mathbf{Q}$, and $\mathbf{K}$ represent the value, target, and source vertex features, respectively. One possible limitation in applying self-attention to graphs is the lack of structural information \cite{dwivedi2021GeneralizationTransformerNetworks}. Various approaches have been proposed to address this issue, including combining adjacency matrix with the attention \cite{ying2021TransformersReallyPerform,kreuzer2021rethinking, rampasek2022RecipeGeneralPowerful}, using graph kernel over neighborhood graph of each node instead of softmax over node features \cite{chen2022StructureAwareTransformerGraph} or including powerful positional and structural encoding \cite{ma2023graph}. Thus, a defining formalism of graph transformers has yet to be proposed to the best of our knowledge. As a result, GTs, in contrast to MPNNs, have no uniform expressiveness upper-bound defined for the entire family. Expressiveness is usually discussed case by case in terms of models ~\cite{zhang2023rethinking,cai2023ConnectionMPNNGraph}.

\subsection{Expressiveness}
The expressiveness of Graph Neural Networks (GNNs) indicates the upper bound of their ability to discriminate between different graphs ~\cite{xu2019HowPowerfulAre}. It has been noted that most MPNNs' expressiveness is bounded by 1-dimensional Weisfeiler Leman (1-WL) graph isomorphism test ~\cite{weisfeiler1968ReductionGraphCanonical, morris2019weisfeiler}. 1-WL is also referred to as the color-refinement algorithm, formulated as
\begin{equation}
\chi_\mathcal{G}^t(v):=\operatorname{hash}\left(\chi_\mathcal{G}^{t-1}(v),\left\{\left\{\chi_\mathcal{G}^{t-1}(u): u \in \mathcal{N}_\mathcal{G}(v)\right\}\right\}\right).
\end{equation}
The ``color map'' of graph $\mathcal{G}$, denoted as $\chi_\mathcal{G}$ maps from $\mathcal{V}$ to a set of colors $\mathcal{C}$. As an iterative algorithm, for each time step $t$, the color for each vertex is updated by a hash function over a multi-set consisting of all the colors of its neighbors. Even though 1-WL is a powerful test known to distinguish a broad class of graphs, it fails to discriminate some simple graph pairs ~\cite{BabaiKucera1979}. To obtain expressive GNNs, methods are proposed based on higher-order WL tests ~\cite{maron2019ProvablyPowerfulGraph,morris2022speqnets}, subgraph isomorphism/homomorphic counting ~\cite{frasca2022understanding,bevilacqua2022EquivariantSubgraphAggregation,welke2023expectation}, equivalent polynomial ~\cite{puny2023equivariant}.

% \subsection{\wl{Ollivier Ricci Curvature}}
% To analyze the over-squashing and bottlenecks in MPNNs, prior work \cite{topping2022UnderstandingOversquashingBottlenecks} introduces the Balanced Forman curvature for measuring information propagation in observed graphs. However, such curvature is limited to unweighted-undirected graphs, hindering our understanding of weighted-directed propagation graphs in the GPNN framework. 

\subsection{Ollivier-Ricci Curvature}
% 3. Olivier-Ricci Curvature definition [TODO: modify]
% 1 Why do we care about Ricci curvature? what does it mean?

In differential geometry, Ricci curvature is a fundamental concept related to  volume growth  and allowing to classify the local characteristic of the space (roughly, whether it is sphere- or hyperboloid-like). Curvature also determines the behavior of parallel lines (whether they converge or diverge), known as geodesic dispersion.  
Discrete curvatures are analogous constructions for graphs (or more generally, metric spaces) trying to mimic some properties of the continuous  curvature. 

%Curvature is an important mathematical object from the field of differential geometry and Riemannian geometry, which determines the \textcolor{red}{geodesic dispersion}. \textit{Discrete curvature} extends the notion of curvature to metric spaces, which becomes handy dealing with spaces that do not have a smooth, differentiable structure. Among all discrete curvature formulations, 
\citet{ollivier2009ricci} introduced a notion of curvature for metric spaces that measures the Wasserstein distance between Markov chains, i.e. random walks, defined on two nodes.
Let $\gG$ be a graph with a distance metric $d_\gG$, and $\mu_{v}$ be a probability measure on $\gG$ for node $v \in \V$. The Ollivier–Ricci curvature of any pair $\{(i,j)|(x,y)\in \V^2, x\neq y \}$ is defined as

\begin{equation}
\kappa_{\mathrm{OR}}(x, y):=1-\frac{1}{d_G(x, y)} W_1\left(\mu_x, \mu_y\right),
\end{equation}

where $W_1$ refers to the first Wasserstein distance between $\mu_i$ and $\mu_j$.

% 2. Discrete ricci curvature has different forms we pick OR curvature because... ?
Ollivier-ricci curvature is the most prominent discrete curvature on metric spaces which quantifies how the geometry of the manifold deviates from flat (Euclidean) space in terms of the metric structure. Other choices of discrete curvatures include the \textbf{Forman–Ricci Curvature}
$$\kappa_{\mathrm{FR}}(x, y):=4-d_x-d_y+3\left|\#_{\Delta}\right|$$ 
and the \textbf{Resistance Curvature}
$$\kappa_{\mathrm{R}}(x, y):=\frac{2\left(p_x+p_y\right)}{R_{x y}}.$$
Specifically, we build on our the framework of OR-curvature, due to its intrinsic relation with MPNN and its geometric and spectral properties relating to graph structure.

\subsection{Over-squashing and over-smoothing Trade-off }
During message-passing, information of distant vertex is cached in vertex features along the path. Some of the vertices end up passing messages growing exponentially with the distance, leading to a potential loss of information known as {\em over-squashing} ~\cite{alon2020bottleneck}. ~\cite{topping2022UnderstandingOversquashingBottlenecks} demonstrates the connection between over-squashing and high negative Ollivire-ricci curvature ~\cite{ollivier2009ricci} for undirected graphs.
On the other hand, stacking multiple massage-passing layers can lead to {\em over-smoothing}, wherein the node embedding from different clusters mixed up, thereby adversely affecting performance \cite{zhao2020PairNormTacklingOversmoothing,wenkel2022OvercomingOversmoothnessGraph}.

{\em Trade-off} between over-smoothing and over-squashing  has been discussed  ~\cite{giraldo2022understanding} . The authors show that message-passing converges to a stationary distribution exponentially according to the spectral gap $\lambda_2$. Finally, cheeger inequality provides a well-established connection between $\lambda_2$, and the Cheeger constant – a measure of bottleneck. As a result, reducing over-smoothing may lead to over-squashing, and vice versa.

Graph transformers aggregate information based on attention. Over-squashing is on GTs are less discussed: {\em 1.} attention is dynamic across layers and is learned from data {\em 2.} analysing tool like Ollivire-ricci curvature could not be applied to asymmetric attention which is adopted by some of recent works.

\subsection{Motivation for Unified Framework}
The exploration of the intricate relationship between Message Passing (MP) and Attention mechanisms highlights a significant yet underexplored avenue in computational research. A notable absence of detailed inspection within existing works prompts the need for a comprehensive approach. By introducing a unified framework, researchers can embark on a systematic expressiveness assessment routine for models encapsulated within this framework, paving the way for a deeper understanding of their capabilities. This framework not only sheds light on the empirical superiority of Graph Transformers (GTs) over Message Passing Neural Networks (MPNNs) but also serves as a bridge to quantify the improvements each design choice in GTs contributes over MPNNs. Furthermore, this approach offers a higher-level taxonomy of existing methods, fostering an environment conducive to the discovery of superior design choices and models. In essence, this logical progression from identifying a foundational connection to enhancing model design and expressiveness underlines the transformative potential of a unified analytical framework in advancing the field.

\clearpage
%In this subsection, we review priors that are related to our proposed GPNN and CURC in different categories: %\\
\section{Related Works}
\label{sec:related_work}
\textbf{Graph Rewiring} has been introduced to combat the over-squashing phenomenon in MPNN.  \citet{topping2022UnderstandingOversquashingBottlenecks} proposed an iterative graph rewiring algorithm based on Balanced Forman Ricci curvature to mitigate the effect of negatively-curved edges on bottlenecks. \citet{gutteridge2023drew} introduced a delayed message passing mechanism, which dynamically performs rewiring on graphs in the form of $k$-hop skip connections. \citet{bruel2022rewiring} employed the strategy to rewire the node to all the other nodes in a receptive field and use positional encoding to describe the original graph structure.

\textbf{Graph Positional Encodings} have been studied to enhance  MPNNs~\cite{zhang2021EigenGNNGraphStructure, lim2022SignBasisInvariant, wang2022EquivariantStablePositional, dwivedi2021GraphNeuralNetworks, bouritsas2022improving, velingker2022AffinityAwareGraphNetworks, you2019PositionawareGraphNeural, li2020DistanceEncodingDesign},
and graph transformers~\cite{dwivedi2021GeneralizationTransformerNetworks, kreuzer2021rethinking, rampasek2022RecipeGeneralPowerful, ying2021TransformersReallyPerform, zhang2023rethinking, ma2023graph} by injecting various graph features, such as spectral information, affinity-measure and geodesic distance.

% \citet{welke2023expectation} utilize \text {Lovász' characterisation} of graph isomorphism to embedding that can theoretically distinguish all graphs with sufficient computing resources. \citet{ma2023graph} introduced a positional embedding based on random-walk matrix at different time step. \cite{bevilacqua2022EquivariantSubgraphAggregation} and \cite{frasca2022understanding} aggregates the encoding based on subgraphs to construct a combined overall graph encoding. 

\textbf{Expressivity} is an eternal topic of graph neural networks (GNNs) research. \citet{xu2019HowPowerfulAre} first points out the expressiveness limitation of MPNNs bounded by 1-WL algorithm on the graph isomorphism test.
Follow-up works have attempted to breakthrough via higher-order-GNNs~\cite{morris2019WeisfeilerLemanGo, maron2019ProvablyPowerfulGraph, bodnar2021WeisfeilerLehmanGo}. However, those methods usually lead to higher order of complexity. Other approaches includ strong structural and positional encoding~\cite{bouritsas2022improving, zhang2023rethinking} and subgraph aggregation~\cite{bevilacqua2022EquivariantSubgraphAggregation, zhou2023relational}. Given various methods stating better than 1-WL expressivity, recent methods have started to rethink the WL hierarchy as the default measurement of the expressiveness \cite{puny2023equivariant, zhang2023rethinking, morris2022speqnets}. %Graph equivariant polynormials \cite{puny2023equivariant} looking in to the space of matrix equivariant polynomials.  

% \cite{bouritsas2022improving} employs a subgraph counting mechanism to the model, \cite{bodnar2021WeisfeilerLehmanGo} performs MPNN on regular Cell Complexes instead of Simplicial Complexes, \cite{bevilacqua2022EquivariantSubgraphAggregation} utilizes the subgraph structure and introduces DSS-WL, where all previously mentioned works proposed model structure that exceeds the expressiveness of WL-test theoretically. 
    
\textbf{Ollivier-Ricci curvature.} There are various works investigating the sole mathematical properties of OR-curvatures proposed by \citet{ollivier2009ricci}. \citet{jost2014ollivier} reveals OR-curvature's connection with local clustering coefficient on undirected graphs where \citet{topping2022UnderstandingOversquashingBottlenecks} utilized to obtain a lower bound estimation for performing graph rewiring. Without restricting it to an unweighted-undirected graph, \citet{bai2020ollivier} extends OR-curvature to weighted graphs by using the weighted graph Laplacian and carries the discussion to continuous-time Ollivier-Ricci flow. \citet{ozawa2020geometric} extends OR-curvature to strongly-connected weighted-directed graph using Perron measure with canonical shortest distance function, and we extend the curvature to weighted distance function to obtain CURC. 

\clearpage
\clearpage
\section{Generality of GPNN framework}
\label{appendix:gpnn}

\subsection{Taxonomy of Existing Graph Models in GPNNs} \label{appendix: taxonomy}

\newcommand{\mpnn}[1]{\textcolor{black}{#1}}
\newcommand{\rewire}[1]{\textcolor{blue}{#1}}
\newcommand{\polyspect}[1]{\textcolor{red}{#1}}
\newcommand{\diff}[1]{\textcolor{violet}{#1}}
\newcommand{\gt}[1]{\textcolor{purple}{#1}}
\newcommand{\spect}[1]{\textcolor{green}{#1}}
\newcommand{\approxgt}[1]{\textcolor{orange}{#1}}

\begin{table}[h]
    \footnotesize
    \centering
    \caption{Demonstration of three different graph model families to showcase the generality of the GPNN framework.}
\scalebox{1.0}{
    \begin{tabular}{@{}lc|c|c@{}}%{lc|c|c}
    % \headrule
    \toprule
         & \makecell[c]{MPNNs
         % \\\cite{xu2019HowPowerfulAre}
         }
         & \makecell[c]{ Rewiring
         % \\ \cite{gutteridge2023drew}
         }
         &\makecell[c]{ GraphTransformer
         % \\\cite{ying2021TransformersReallyPerform}
         }
          \\ %&\\
        \midrule
        & \includegraphics[width=0.26\linewidth]{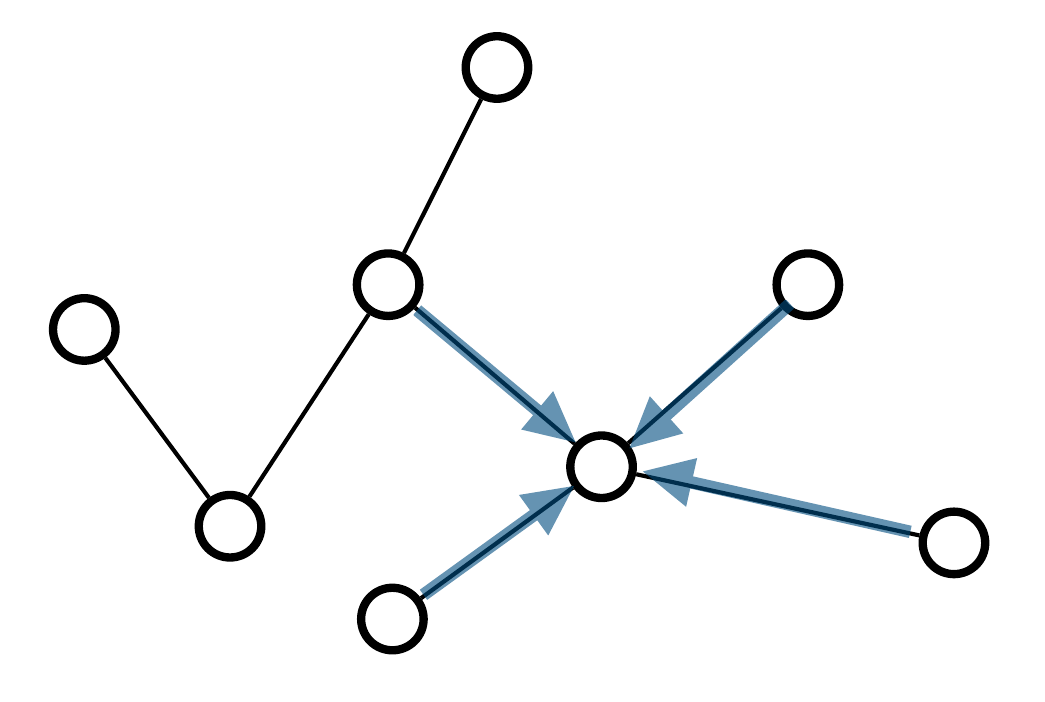} & 
                   \includegraphics[width=0.26\linewidth]{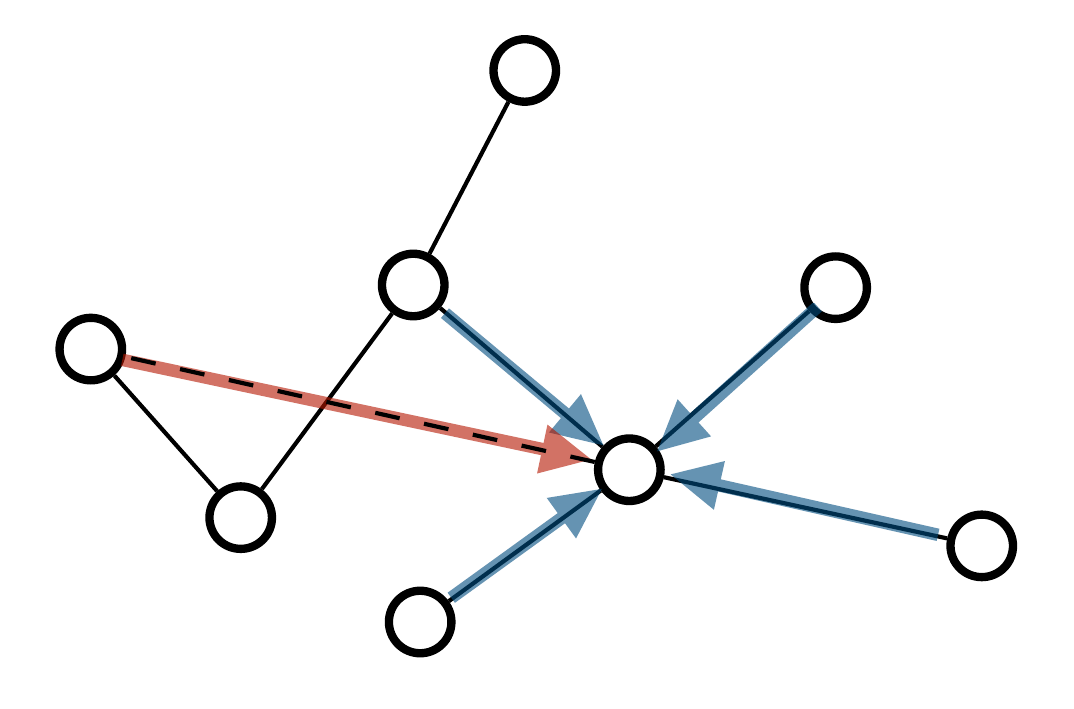} & 
                   \includegraphics[width=0.26\linewidth]{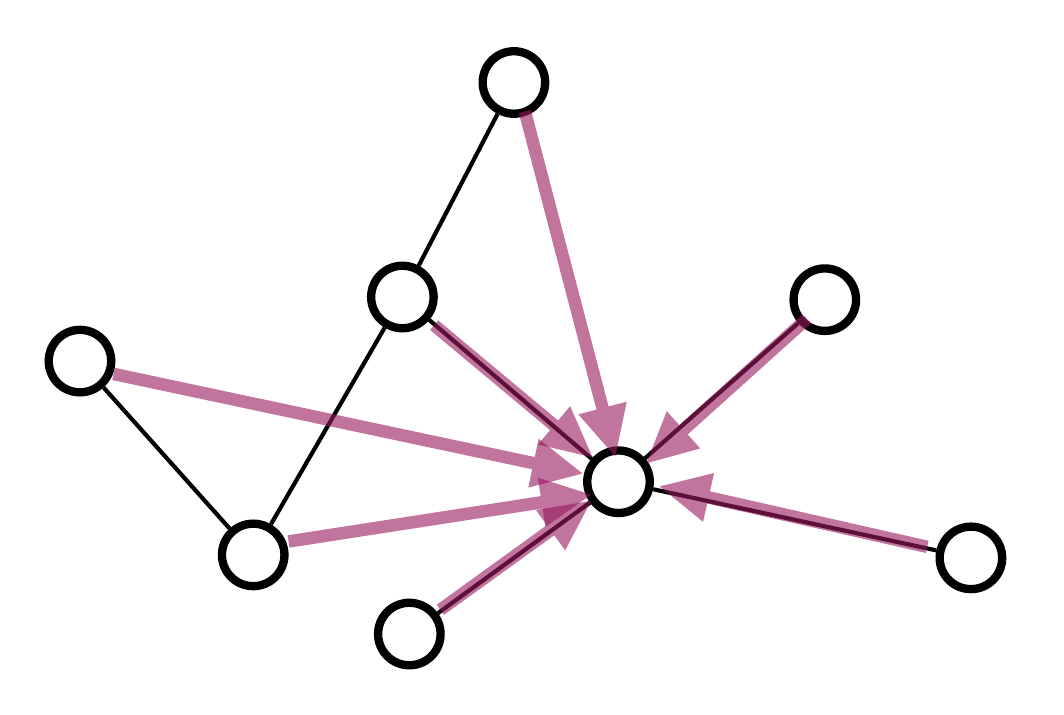}  \\
                   %& \includegraphics[width=0.2\linewidth]{example-image-c}\\
                   \midrule
                   &\multicolumn{3}{c}{Taxonomy of Propagation Graphs}
                   \\
                   \midrule
                   & Local  & Non-Local & Global \\
                   & Static/Dynamic & Static/Dynamic & Dynamic  \\
                  & with/without Feature & 
                   without Feature & 
                   with Feature \\
                  &  Feat. Dep./Ind. & 
                   Feat. Ind.
                   & 
                   Feat. Dep.\\
        \bottomrule
        % $\omega(\mX_u, \mX_v, f(\mA))$ & $f(\mA)_{u,v}$ & $\idt_{\{k\}}(d_\gG(\mA)) (\diag(\mA \bs{1})^{-1/2}\; {\mA}\; \diag(\bs{1}^\intercal {\mA} )^{-1/2})$ & $exp\left((\mW_1 \mX_\elem{u})^\intercal (\mW_2\mX_\elem{v}) + f(\mA)_\elem{u,v}\right)$  \\
        % $f(\mA)$ & $\mA$  &  $\idt_{\{k\}}(d_\gG(\mA)) (\diag(\mA \bs{1})^{-1/2}\; {\mA}\; \diag(\bs{1}^\intercal {\mA} )^{-1/2})$ & $\operatorname{SP}(\mA)$  \\
        % $\rho(\mP)$ & $\diag(\mP \bs{1})^{-1/2}\; {\mP}\; \diag(\bs{1}^\intercal {\mP} )^{-1/2}$& $\mP$ & $\diag(\mP \bs{1})^{-1}\; {\mP}$ \\
    \end{tabular}
}
    
    % \caption{We pick three methods from existing literature coming from three different model families to showcase the generality of the GPNN framework. We show that by defining different adjacency function $f$ and entry-wise function $\omega$ (with the help of $\rho$ to rescale) we can recover GIN, DREW, and Graphormer. $\operatorname{SP}(\mA)$ denotes the shortest path distance of graph.}
    \label{tab:casting}
\end{table}

\begin{table*}[h!]
    \centering
    \begin{tabular}{c|c|c}
    \toprule
    Propagation Graph  &  Static  &  Dynamic   \\
    \midrule    Local   &  \makecell[c]{\mpnn{GCN~\cite{kipf2017SemiSupervisedClassificationGraph}}, \\
    \mpnn{GraphSAGE~\cite{hamilton2017InductiveRepresentationLearning}},  \\ \mpnn{GIN~\cite{xu2019HowPowerfulAre}}
    } & \makecell[c]{\mpnn{MoNet~\cite{monti2017GeometricDeepLearning}}, \\
    \mpnn{GAT*~\cite{velickovic2018GraphAttentionNetworks}}, \\ 
    \mpnn{GatedGCN*}~\cite{bresson2018ResidualGatedGraph}
    }            \\ \midrule
    Non-Local & \makecell[c]{\rewire{SDRF~\cite{topping2022UnderstandingOversquashingBottlenecks}}  \\
                             \diff{GDC~\cite{gasteiger2019DiffusionImprovesGraph}}
                            }
                            & \makecell[c]{\rewire{Drew~\cite{gutteridge2023drew}} \\ \rewire{LASER~\cite{barbero2023locality}} \\
                            \polyspect{ChebNet~\cite{defferrard2017ConvolutionalNeuralNetworks}},  \\
                            \polyspect{BernNet~\cite{barbero2022SheafNeuralNetworks}}, \\
                            \polyspect{JacobConv~\cite{wang2022HowPowerfulAre}} \\
                            \diff{SIGN~\cite{frasca2020SIGNScalableInception}},
                            \diff{ADC~\cite{zhao2021AdaptiveDiffusionGraph}} \\
                            \approxgt{Exphormer*~\cite{shirzad2023ExphormerScalingGraph}}, \\
                            \approxgt{MPNN+VN*~\cite{cai2023ConnectionMPNNGraph}}, \\  \approxgt{GPS-BigBird*~\cite{rampasek2022RecipeGeneralPowerful}}
                            }   
                            \\ \midrule 
    Global &  DeepSet~\cite{zaheer2017DeepSets} &  \makecell[c]{\spect{Spectral GCN~\cite{bruna2014SpectralNetworksLocally}}, \\ \spect{Specformer~\cite{bo2023SpecformerSpectralGraph}}, \\
                        \gt{SANs*~\cite{kreuzer2021rethinking}},\\
                        \gt{Graphormer*~\cite{ying2021TransformersReallyPerform}},\\
                        \gt{GraphGPS-Full*~\cite{rampasek2022RecipeGeneralPowerful}}, \\ \gt{EGT*~\cite{hussain2022GlobalSelfAttentionReplacement}},\\
                        \gt{GRIT*~\cite{ma2023graph}}}    
    \\ \bottomrule
    % \bottomrule
    \end{tabular}
    \caption{Taxonomy in GPNN Framework} 
    \makecell[l]{*: denotes that the propagation matrices are conditional on the node feature.
    We disregard the impact of residual connections.
    \\
    \mpnn{MPNNs}, \rewire{Graph Rewiring}, \diff{Diffusion Enhanced GNNs}, \polyspect{Polynomical Spectral GNN}, \\ \spect{Full Spectral GNNs},  \gt{Graph Transformers}, \approxgt{Approximated/Efficient Graph Transformers}} 
    \label{tab:taxonomy}
\end{table*}

\subsection{Existing Models as GPNNs} \label{appendix: casting}
\paragraph{Notation}
\begin{itemize}
    \item $\mathbf{A} := [a_{ij}]_{i, j \in \cV}$  adjacency matrix
    % \item $\mathbf{P} := [\omega_{ij}]_{i, j \in \cV}$  propagation matrix
    \item $\mathbf{D} := \operatorname{diag}[d_{ii}]_{i \in \cV}$  degree (diagonal) matrix
    \item $\tilde{\mathbf{A}}$, $\tilde{\mathbf{D}}$ denote the corresponding matrices with self-loops
\end{itemize}

We demonstrate how different graph models can be cast into the GPNN framework by defining adjacency functions and connectivity functions.
We provide the examples with scalar-valued feature and one propagation graph with further specification, 
which can be directly extended vector-valued features via multi-head 
(i.e., multiple propagation matrices) and/or convolution architecture.

We ignore other optional components (e.g., normalization layers, residual connection) in the networks.
Without further specification,
we assume that $h_i^{l+1} := m_i^{l+1}$,  and the superscript of layer index on weight matrices/vectors are dropped for simplicity.

% \subsection{Structural+Local}
% \mlh{To unified the notation; to double check the usage of $u$}

\paragraph{GCN~\cite{kipf2017SemiSupervisedClassificationGraph}} generates the propagation matrix by symmetric normalized adjacency matrix with self-loops,
\begin{equation}
   \omega^{l+1}_{ij}  :=  \tilde{a}_{ij} \tilde{d}_i^{-1/2} \tilde{d}_j^{-1/2}
\end{equation}

\paragraph{MoNet~\cite{monti2017GeometricDeepLearning}} 
proposes to generate the propagation matrix as a Gaussian kernel based on ``pseudo-coordinate'' between two adjacent nodes, which is $[d_i^{-1}, d_j^{-1}], \forall (v,u) \in \cE$,
\begin{equation}
    \begin{aligned}
       \omega^{l+1}_{ij}  & = \tilde{a}_{ij} \cdot \exp \left(-\frac{1}{2} (\tilde{\mathbf{u}}_{ij} - \boldsymbol{\mu})^\intercal \boldsymbol{\Sigma}^{-1} (\tilde{\mathbf{u}}_{ij} - \boldsymbol{\mu}) \right) \\
       & \text{where } \tilde{\mathbf{u}}_{ij} = \mathbf{W}\cdot [d_i^{-1}, d_j^{-1}]^\intercal + \mathbf{b}
    \end{aligned}
\end{equation}
and $\mathbf{W} \in \RR^{r \times 2}$, $\mathbf{b} \in \RR^{r}$, $\boldsymbol{\mu} \in \RR^{r}$ and $\boldsymbol{\Sigma} \in \RR^{r \times r}$ are learnable parameters.

\paragraph{GraphSAGE-GCN~\cite{hamilton2017InductiveRepresentationLearning}} generates the propagation matrix via row-normalized adjacency matrix with self-loops,
\begin{equation}
   \omega^{l+1}_{ij}  = \tilde{a}_{ij} \tilde{d}_i^{-1}  \, .
\end{equation}

\newcommand{\bW}{\mathbf{W}}

\paragraph{GraphSAGE-Average~\cite{hamilton2017InductiveRepresentationLearning}} 
generates the propagation matrix, similar to GraphSAGE-GCN, by the row-normalized adjacency matrix without self-loops and self-loops with extra learnable parameters, 
\begin{equation}
    \begin{aligned}
   \omega^{l+1}_{ij}  &=  a_{ij} d_i^{-1}  \\
    h_i^{l+1}  &= \mathbf{W}_1 h_i^{l} + \mathbf{W}_2 m_j^{l+1}
    \end{aligned}
 \end{equation}
 where $\bW_1, \bW_2 \in \RR^{c \times c}$ are learnable matrices.

\paragraph{GIN~\cite{xu2019HowPowerfulAre}}
generates the propagation matrix based on the unnormalized adjacency matrix and an extra learnable parameter for self-loops, 
\begin{equation}
    \begin{aligned}
   \omega^{l+1}_{ij}  &=   a_{ij}   \\
    h_v^{l+1}  &= (1 - \theta_0) h_v^{l} + \theta_1 m_v^{l+1}
    \end{aligned}
 \end{equation}
where $\theta_0, \theta_1 \in \RR$ are learnable parameters.

\paragraph{GAT~\cite{velickovic2018GraphAttentionNetworks}}
generate the propagation matrix based on the attention mechanism on the observed edges,
\begin{equation}
    \begin{aligned}
        \omega^{l+1}_{ij} &= \tilde{a}_{ij} \operatorname{Softmax}_{j' \in \{j' |  \tilde{a}_{ij'} = 1\}} \left(\operatorname{LeakyReLU} \left( \mathbf{w}_1^\intercal \mathbf{W} \mathbf{x}_i + \mathbf{w}_2^\intercal \mathbf{W} \mathbf{x}_j  \right) \right)  \\
    \end{aligned}
\end{equation}
where node representation $\mathbf{x}_i \in \RR^d$,
 
\paragraph{GatedGCN~\cite{bresson2018ResidualGatedGraph, dwivedi2020BenchmarkingGraphNeural}},
similar to GAT, 
include semantic information into the propagation matrix via the (normalized) gating mechanism,
\begin{equation}
    \begin{aligned}
        \omega^{l+1}_{ij} &=  \tilde{a}_{ij} \cdot \frac{u_{ij}}{\sum_{j' \in \{j' |  \tilde{a}_{ij'} = 1\}} u_{ij'}}  \\
        & \text{where } u_{ij} = \sigma(\mathbf{w}_1^\intercal \mathbf{x}_i + \mathbf{w}_2^\intercal \mathbf{x}_j) \\
        h_i^{l+1}  &= \theta_1 h_i^{l} + \theta_2 m_j^{l+1}
    \end{aligned}
\end{equation}
where $\sigma$ denotes the Sigmoid operation; $\theta_0, \theta_1 \in \RR$ are learnable scalar; $\mathbf{w}_1, \mathbf{w}_2 \in \RR^{r}$ are weight vectors; $\mathbf{x}_i \in \RR^d$ stands for the node representation.

\paragraph{(Graph Rewiring) Drew~\cite{gutteridge2023drew}}
proposes to generate the propagation matrix by rewiring to nodes in top-$L$-hop neighborhood for $L$-th layer,
\begin{equation}
    \begin{aligned}
    \omega^{l+1}_{ij} & = \left\{ 
    \begin{array}{ll}
       \frac{\theta_k}{|\{j'|d_\gG(i,j')=k\}|} & \text{if } d_\gG(i,j)=k.   \\
       0 & \text{otherwise} 
    \end{array} \right. \\ & k=1, \dots, L ,  
    \\
        h_i^{l+1}  &= \theta_0 h_i^{l} + m_j^{j} 
    \end{aligned}
\end{equation}
where $d_\gG: \cV \times \cV \to \mathbf{Z}_{\geq 0}$ denotes the shortest-path distance on graphs; $\theta_1, \cdots, \theta_L \in \RR$ are learnable weights.

\paragraph{(Polynomial Spectral GNN) ChebNet~\cite{defferrard2017ConvolutionalNeuralNetworks}}
generate the propagation matrix as the approximation to Laplacian eigenvectors via the Chebyshev polynomial, which can be viewed as the top-$K$ power of the normalized $\mathbf{L}=\mathbf{D}-\mathbf{A}$,

\begin{equation}
    \begin{aligned}
        \omega^{l+1}_{ij} &=
        \sum_{k=0}^{K-1} \theta_k \tau_{ij}^k 
    \end{aligned}
\end{equation}
where $\tau_{ij}^k$ is the $i,j$-element of the $T_k(\tilde{L})$, in which, $\tilde{L}:= (\mathbf{D} - \mathbf{A})/\lambda_\text{max} - \mathbf{I})$ is the noramlized Laplacian matrix by the largest eigenvalue and $T_k(x):= 2x T_{k-1}(x) - T_{k-2}(x)$ is the Chebyshev polynomial of order $k$.

\paragraph{SANs~\cite{kreuzer2021RethinkingGraphTransformers}}

\begin{equation}
    \begin{aligned}
    \omega^{l+1}_{ij} =
    \operatorname{Softmax}_{j \in \cV} \left( (\mathbf{W}_1 (\mathbf{x}_i + \mathbf{e}_i))^\intercal 
    (\mathbf{W}_2 (\mathbf{x}_{j}+\mathbf{e}_j)) 
    \right)
    \end{aligned}
\end{equation}
where $\mathbf{W}_1, \mathbf{W}_2 \in \RR^{r \times d}$ are learnable weight matrices;
$\mathbf{e}_{i} \in \RR^d$ stands for the positional encoding for node $i$.
Note that, the scaling factor in scaled dot-product is ignored for simplicity, same for the following.

\paragraph{Graphormer ~\cite{ying2021TransformersReallyPerform}}

\begin{equation}
    \begin{aligned}
    \omega^{l+1}_{ij} =
    \operatorname{Softmax}_{j \in \cV} \left( (\mathbf{W}_1 \mathbf{x}_i)^\intercal (\mathbf{W}_2\mathbf{x}_{j}) + \mathbf{w}^\intercal \mathbf{e}_{ij}
    \right)
    \end{aligned}
\end{equation}
where $\mathbf{W}_1, \mathbf{W}_2 \in \RR^{r \times d}$ and $\mathbf{w} \in \RR^d$ are learnable weight matrices/vectors;
$\mathbf{e}_{ij} \in \RR^d$ stands for the positional encoding and/or edge-attributes between node $i$ and node $j$.

\paragraph{GRIT~\cite{ma2023graph}}

\begin{equation}
    \begin{aligned}
        \omega^{l+1}_{ij} = 
        \mathbf{w}^\intercal \text{ReLU}(\rho((\mathbf{W}_1 \mathbf{x}_i + \mathbf{W}_2\mathbf{x}_j) \odot \mathbf{W}_3 \mathbf{e}_{ij}) + \mathbf{W}_4 \mathbf{e}_{ij})) 
    \end{aligned}
\end{equation}
where $\mathbf{W}_1, \mathbf{W}_2, \mathbf{W}_3, \mathbf{W}_4 \in \RR^{r \times d}$ and $\mathbf{w} \in \RR^r$ are learnable weight matrices/vectors;
$\mathbf{e}_{ij} \in \RR^d$ stands for the positional encoding and/or edge-attributes between node $i$ and node $j$.

\clearpage
\clearpage
\section{Experimental Details}

\subsection{Details of GPNN-PE}
\label{appendix:gpnn-pe}

\subsubsection{Model Architecture}

To verify our theoretical findings, we build up  a purely structural-based GPNN, called GPNN-PE, based on the SOTA GTs - GRIT~\cite{ma2023graph}, by removing the \textit{query-key architecture} which models the token similarity on node representations.

\newcommand{\x}{\mathbf{x}}
\newcommand{\e}{\mathbf{e}}
\renewcommand{\P}{\mathbf{P}}
\renewcommand{\RR}{\mathbb{R}}
\newcommand{\W}{\mathbf{W}}

In each layer, we update node representations $\x_u \forall u \in \mathcal{V}$ and node-pair representations $\e_{u, v}, \forall u,v \in \mathcal{V}$.
Similar to GRIT, we initialize these using the initial node features and our RRWP positional encodings: $\x_i  = [\mathbf{x}'_i \| \P_{i,i}] \in \RR^{d_h + K}$ and $\e_{i,j} = [\e'_{i,j} \| \P_{i,j}] \in \RR^{d_e + K}$, where $\x'_i \in \RR^{d_h}$ and $\e'_{i,j} \in \RR^{d_e}$ are observed node and edge attributes, respectively; $\P_{i,j}$ is the relative positional encoding for graphs.
Note that, if node/edge attributes are not present in the data, we can set $\x'_{i}$/$\e'_{i,j}$ as zero-vectors $\mathbf{0} \in \RR^d$. 
We set $\e'_{i,j} = \mathbf{0}$ if there is no observed edge from $i$ to $j$ in the original graph.

We replace the original attention computation in GRIT with a multi-layer perceptron (MLP):
\begin{equation}
   \begin{aligned}
    & \hat{\e}_{i,j} = \sigma(  \W_{1}\e_{i,j}) \in \mathbb{R}^{d'}, \\
    &\alpha_{ij} = \text{Softmax}_{j \in \mathcal{V}} (\W_{2} \hat{\e}_{i,j}) \in \mathbb{R},\\
    & \hat{\x}_i =  \sum_{j \in \mathcal{V}} \alpha_{ij} \cdot ( \W_3 \x_j + \W_{4} \hat{\e}_{i,j}) \in \mathbb{R}^{d''},\\
   \end{aligned}
\end{equation}
where $\sigma$ is a non-linear activation (ReLU by default); 
$\W_1 \in \mathbb{R}^{d' \times d}$, $\W_2 \in \mathbb{R}^{1 \times d'}$, $\W_3 \in \mathbb{R}^{d'' \times d}$ and $\W_4 \in \mathbb{R}^{d'' \times d'}$ are learnable weight matrices.

Following GRIT, we retain the update of edges and the multiple heads (say, $N_h$ heads) without further specification:
\begin{equation}\label{eq:attention_out}
   \begin{aligned}
    \x^\text{out}_i =  \sum_{h=1}^{N_h} \W^h_\text{O}\hat{\x}^h_i \in \mathbb{R}^d\,,  \\
    \e^\text{out}_{ij} =  \sum_{h=1}^{N_h} \W^h_\text{Eo}\hat{\e}^h_{ij}  \in \mathbb{R}^d \,, \\
   \end{aligned}
\end{equation}
where $\W^h_\text{O}, \W^h_\text{Eo}  \in \mathbb{R}^{d \times d''}$ are output weight matrices for each head $h$.

%for stabilizing the training.
% We also include biases in our implementation, but they are omitted here for simplicity.
% Note that we update the pair representations $\e_{i,j}$, so our Transformer is capable of updating the positional encodings. In particular, our Transformer is capable of applying an elementwise $\mlp$ to $\P$, as we showed was useful in Proposition~\ref{prop:mlp_rrwp}.

% Similarly to other self-attention mechanisms, our proposed attention mechanism can be extended to multiple heads (say, $N_h$ heads) by assigning different weight matrices for different heads. We perform the above computations for different heads $h \in \{1, \ldots, N_h\}$ to get representations $\hat{\x}^h_i$ and $\hat{\e}^h_{i,j}$, then combine the different heads as follows:
% \begin{equation}\label{eq:attention_out}
%    \begin{aligned}
%     \x^\text{out}_i =  \sum_{h=1}^{N_h} \W^h_\text{O}\hat{\x}^h_i \in \mathbb{R}^d\,,  \\
%     \e^\text{out}_{ij} =  \sum_{h=1}^{N_h} \W^h_\text{Eo}\hat{\e}^h_{ij}  \in \mathbb{R}^d \,, \\
%    \end{aligned}
% \end{equation}
% where $\W^h_\text{O}, \W^h_\text{Eo}  \in \mathbb{R}^{d \times d''}$ are output weight matrices for each head $h$. 

\subsubsection{Positional Encoding}
\label{rrwp}
\renewcommand{\rw}{\mathbf{M}}

% In contrast with positional encoding, which embed each of node to a euclidean space, relative positional encoding embed a pair of node regarding their relative position in the graph. 
In this work, we apply Relative Random walk positional encoding utilized in GRIT~\cite{ma2023graph}, which is one of the most expressive graph positional encoding.
Let $\mathbf{A} \in \RR^{n \times n}$ be the adjacency matrix of a graph $(\V, \E)$ with $n$ nodes, and let $\mathbf{D}$ be the diagonal degree matrix. 
Define $\rw := \mathbf{D}^{-1}\mathbf{A}$, and note that $\rw_{ij}$ is the probability that $i$ hops to $j$ in one step of a simple random walk.
The proposed relative random walk probabilities (RRWP) initial positional encoding is defined for each pair of nodes $i, j \in \V$ as follows:
%\todo{change the notation of RW-matrix for clarification}
\begin{equation}
    \mathbf{P}_{i,j} = [\mathbf{I}, \rw, \rw^2, \dots, \rw^{k-1}]_{i,j} \in \mathbb{R}^k,
\end{equation}
in which $\mathbf{I}$ is the identity matrix.
In other words, in GPNN-PE, $\psi: \R^{n \times n} \to \R^{n \times n \times k}$ is defined as
\begin{equation}
    \psi(\mathbf{A})_{i,j}:= [\mathbf{I}, \mathbf{D}^{-1}\mathbf{A}, (\mathbf{D}^{-1}\mathbf{A})^2, \dots, (\mathbf{D}^{-1}\mathbf{A})^{k-1}]_{i,j}  
\end{equation}
% RRWP can be viewed as an extension of RWSE~\cite{dwivedi2021GraphNeuralNetworks} to all pairs of nodes (from absolute PE to Relative PE) and thus encode more positional information.

% For any node $i \in \V$, the diagonal  $\P_{i,i}$ can additionally be utilized as an initial node-level structural encoding, which is the same as the Random Walk Structural Encodings (RWSE) used in past work~\citep{dwivedi2021GraphNeuralNetworks, rampasek2022RecipeGeneralPowerful}. 
% The parameter $K \in \mathbb{N}$ controls the maximum length of random walks considered.

\subsection{More about Experimental}
\label{appendix:experiment_details}

\subsubsection{Baselines}
We primarily compare our methods with the SOTA graph transformer, GRIT~\cite{ma2023graph},
as well as a number of prevalent graph-learning models:
% \ars{GAT-POS should be included here}
popular message-passing neural networks (GCN~\cite{kipf2017SemiSupervisedClassificationGraph}, GIN~~\cite{xu2019HowPowerfulAre} and its variant with edge-features~\cite{hu2020StrategiesPretrainingGraph}, GAT~\cite{velickovic2018GraphAttentionNetworks}, GatedGCN~\cite{bresson2018ResidualGatedGraph}, GatedGCN-LSPE~\cite{dwivedi2021GraphNeuralNetworks},
PNA~\cite{corso2020PrincipalNeighbourhoodAggregation}); 
Graph Transformers (Graphormer~\cite{ying2021TransformersReallyPerform}, 
K-Subgraph SAT~\cite{chen2022StructureAwareTransformerGraph}, EGT~\cite{hussain2022GlobalSelfAttentionReplacement}, SAN~\cite{kreuzer2021RethinkingGraphTransformers}, 
Graphormer-URPE~\cite{luo2022your},
Graphormer-GD~\cite{zhang2023rethinking}, GraphGPS~\cite{rampasek2022RecipeGeneralPowerful}); and other recent Graph Neural Networks with SOTA performance (DGN~\cite{beani2021DirectionalGraphNetworks},
GSN~\cite{bouritsas2022ImprovingGraphNeural}, 
CIN~\cite{bodnar2021WeisfeilerLehmanGo},
CRaW1~\cite{toenshoff2021graph},
GIN-AK+~\cite{zhao2021stars}).

\subsubsection{Extended Experimental Results}
We also evaluate GPNN-PE on 
on five datasets from the Benchmarking GNNs work ~\cite{dwivedi2020BenchmarkingGraphNeural} (as shown in Table~\ref{tab:exp_bmgnn}).
These datasets are among the most widely used graph benchmarks and cover diverse graph learning tasks,
including node classification, graph classification, and graph regression, with a focus on graph structure and long-range dependencies.

\begin{table*}[h!]
    \centering
    \caption{
    Test performance in five benchmarks from \cite{dwivedi2020BenchmarkingGraphNeural}. 
    Shown is the mean $\pm$ s.d. of 4 runs with different random seeds. Highlighted are the top \first{first}, \second{second}, and \third{third} results. 
    \# Param $\sim500K$ for ZINC, PATTERN, CLUSTER and $\sim 100K$ for MNIST and CIFAR10.
    % ${}^*$ indicates statistically significant difference against the second-best result from the two-sample one-tailed t-test.
    }
    \vskip 0.12in
    {\scriptsize
    % \resizebox{2*\columnwidth}{!}{
    \begin{tabular}{lccccc}
    \toprule
       \textbf{Model}  &  \textbf{ZINC} & \textbf{MNIST} & \textbf{CIFAR10} & \textbf{PATTERN} & \textbf{ CLUSTER} \\
       \cmidrule{2-6} 
       & \textbf{MAE}$\downarrow$  &  \textbf{Accuracy}$\uparrow$ & \textbf{Accuracy}$\uparrow$ & \textbf{Accuracy}$\uparrow$ & \textbf{Accuracy}$\uparrow$ \\
       \midrule
       GCN  & $0.367 \pm 0.011$ & $90.705 \pm 0.218$ & $55.710 \pm 0.381$ & $71.892 \pm 0.334$ & $68.498 \pm 0.976$ \\
GIN  & $0.526 \pm 0.051$ & $96.485 \pm 0.252$ & $55.255 \pm 1.527$ & $85.387 \pm 0.136$ & $64.716 \pm 1.553$ \\
GAT & $0.384 \pm 0.007$ & $95.535 \pm 0.205$ & $64.223 \pm 0.455$ & $78.271 \pm 0.186$ & $70.587 \pm 0.447$ \\
GatedGCN & $0.282 \pm 0.015$ & $97.340 \pm 0.143$ & $67.312 \pm 0.311$ & $85.568 \pm 0.088$ & $73.840 \pm 0.326$ \\
GatedGCN-LSPE & $0.090 \pm 0.001$ & $-$ & $-$ & $-$ & $-$ \\
% PNA & $0.188 \pm 0.004$ & $97.94 \pm 0.12$ & $70.35 \pm 0.63$ & $-$ & $-$ \\
% DGN & $0.168 \pm 0.003$ & $-$ &  {${72.838 \pm 0.417}$} & $86.680 \pm 0.034$ & $-$ \\
% GSN  & $0.101 \pm 0.010$ & $-$ & $-$ & $-$ & $-$ \\
% \midrule CIN & {${0.079} \pm {0.006}$} & $-$ & $-$ & $-$ & $-$ \\
% CRaW1 & $0.085 \pm 0.004$ & ${97.944} \pm {0.050}$ & $69.013 \pm 0.259$ & $-$ & $-$ \\
% GIN-AK+ & ${0 . 0 8 0} \pm {0 . 0 0 1}$ & $-$ & $72.19 \pm 0.13$ & {${86.850 \pm 0.057}$} & $-$ \\
\midrule SAN & $0.139 \pm 0.006$ & $-$ & $-$ & $86.581 \pm 0.037$ & $76.691 \pm 0.65$ \\
Graphormer & $0.122 \pm 0.006$ & $-$ & $-$ & $-$ & $-$ \\
% K-Subgraph SAT & $0.094 \pm 0.008$ & $-$ & $-$ & {${86.848 \pm 0.037}$} & $77.856 \pm 0.104$ \\
% EGT & $0.108 \pm 0.009$ & \first{$\mathbf{98.173 \pm 0.087}$} & $68.702 \pm 0.409$ & $86.821 \pm 0.020$ &  \second{$\mathbf{79.232 \pm 0.348}$} \\
% Graphormer-URPE & $0.086 \pm 0.007$ & $-$ & $-$ & $-$ & $-$  \\
Graphormer-GD & $0.081 \pm 0.009$ & $-$ & $-$ & $-$ & $-$ \\
 GPS & {${0.070} \pm {0.004}$} & {${98.051 \pm 0.126}$} & {${72.298 \pm 0.356}$} & $86.685 \pm 0.059$ & {${78.016 \pm 0.180}$} \\
GRIT & \first{$\mathbf{0.059 \pm 0.002}$}&  \third{$\mathbf{98.108  \pm 0.111}$} & \first{$\mathbf{76.468 \pm 0.881}$} & \first{$\mathbf{87.196 \pm 0.076}$}  & \first{$\mathbf{80.026 \pm 0.277}$} \\
\midrule
GPNN-PE &  \second{$\mathbf{0.060 \pm 0.003}$} &   \second{$\mathbf{98.165 \pm 0.077}$}&   \second{$\mathbf{75.505 \pm 0.642
}$}& \second{$\mathbf{87.083 \pm 0.035}$}  & \third{$\mathbf{78.878 \pm 0.152}$}        \\
\quad + static & \third{$\mathbf{0.064 \pm 0.002}$} &  {$98.018 \pm 0.024$} & \third{$\mathbf{75.050, 0.282}$} & \third{$\mathbf{87.045 \pm 0.032}$}  & {$
78.830 \pm 0.127$} \\
\quad + 1-head & {$0.066 \pm 0.005$} &  {$97.560 \pm 0.090$} & {$72.042 \pm 0.714$} & {$86.965 \pm 0.043$}  & {$78.373 \pm 0.212$} \\
       \bottomrule
    \end{tabular}
   % }
   }
\\
    \label{tab:exp_bmgnn}
\end{table*}

\subsubsection{Descriptions of Datasets}

A summary of the statistics and characteristics of datasets is shown in Table.~\ref{tab:dataset}. The first five datasets are from \citet{dwivedi2020BenchmarkingGraphNeural} and the last two are from \citet{dwivedi2022LongRangeGraph}.
Readers are referred to \citet{dwivedi2020BenchmarkingGraphNeural} and \citet{dwivedi2022LongRangeGraph} for more details about the datasets.

% \dl{should cite original dataset sources in appendix}

\begin{table}[h!]
    \centering
    \caption{Overview of the graph learning datasets involved in this work~\cite{dwivedi2020BenchmarkingGraphNeural, dwivedi2022LongRangeGraph, irwin2012ZINCFreeTool, hu2021ogblsc} .}
    \scriptsize
    \setlength{\tabcolsep}{1.6pt}
    \begin{tabular}{l|ccccccc}
    \toprule
       \textbf{Dataset} & \textbf{\# Graphs} & \textbf{Avg. \# nodes} & \textbf{Avg. \# edges}  &  \textbf{Directed} 
 & \textbf{Prediction level} & \textbf{Prediction task} & \textbf{Metric}\\
 \midrule
        ZINC & 12,000 & 23.2 & 24.9 & No &  graph & regression & Mean Abs. Error \\
        MNIST &  70,000  &70.6  & 564.5 & Yes  & graph & 10-class classif. &  Accuracy \\
        CIFAR10 & 60,000 & 117.6 & 941.1 & Yes  & graph & 10-class classif. & Accuracy \\
        PATTERN & 14,000 & 118.9 & 3,039.3  & No & inductive node & binary classif. & Weighted Accuracy \\
        CLUSTER & 12,000 & 117.2 & 2,150.9 &  No & inductive node &  6-class classif. & Accuracy \\ 
        \midrule
        Peptides-func & 15,535 & 150.9 & 307.3 & No & graph & 10-task classif. &  Avg. Precision \\
        Peptides-struct &  15,535  & 150.9 & 307.3 & No & graph & 11-task regression &  Mean Abs. Error  \\
        % \midrule
        % PCQM4Mv2 & 3,746,620 & 14.1 & 14.6 & No & graph & regression & Mean Abs. Error \\
        \bottomrule
    \end{tabular}
    \label{tab:dataset}
\end{table}

\subsubsection{Dataset splits and random seed}
Our experiments are conducted on the standard train/validation/test splits of the evaluated benchmarks.
For each dataset, we execute 4 runs with different random seeds (0,1,2,3) and report the mean performance and standard deviation.

\subsubsection{Hyperparameters}

Due to the limited time and computational resources, we did not perform an exhaustive search or a grid search on the hyperparameters.
We mainly follow the hyperparameter setting of GRIT~\cite{ma2023graph} and make slight changes to fit the number of parameters into the commonly used parameter budgets.
For the benchmarks from \cite{dwivedi2020BenchmarkingGraphNeural, dwivedi2022LongRangeGraph}, we follow the most commonly used parameter budgets: up to 500k parameters for ZINC, PATTERN, CLUSTER, Peptides-func and Peptides-struct; and $\sim$100k parameters for MNIST and CIFAR10.

The final hyperparameters are presented in Tables.~\ref{tab:bmgnn_hparam} and Tables.~\ref{tab:lrgb_hparam}.

\begin{table}[h!]
    \centering
    \caption{Hyperparameters for five datasets from BenchmarkingGNNs \cite{dwivedi2020BenchmarkingGraphNeural}.
    }
    \label{tab:bmgnn_hparam}
\begin{tabular}{lccccc}
\toprule
Hyperparameter & ZINC & MNIST & CIFAR10 & PATTERN & CLUSTER \\
\midrule
\# Transformer Layers & 10 & 4 & 4 & 10 & 16 \\
Hidden dim & 64 & 52 & 52 & 64 & 48 \\
\# Heads & 8 & 4 & 4 & 8 & 8 \\
Dropout & 0 & 0 & 0 & 0 & $0.01$ \\
Attention dropout & $0.2$ & $0.5$ & $0.5$ & $0.2$ & $0.5$ \\
Graph pooling & sum & mean & mean & $-$ & $-$ \\
\midrule
PE dim (RW-steps) & 21 & 18 & 18 & 21 & 32 \\
PE encoder & linear & linear & linear & linear & linear \\
\midrule
Batch size & 32/256 & 16 & 16 & 32 & 16 \\
Learning Rate & $0.001$ & $0.001$ & $0.001$ & $0.0005$ & $0.0005$ \\
\# Epochs & 2000 & 200 & 200 & 100 & 100 \\
\# Warmup epochs & 50 & 5 & 5 & 5 & 5 \\
Weight decay & $1 \mathrm{e}-5$ & $1 \mathrm{e}-5$ & $1 \mathrm{e}-5$ & $1 \mathrm{e}-5$ & $1 \mathrm{e}-5$ \\
\midrule
\# Parameters & 417,877 & 100,754, & 98,238 & 353, 877 & 319,670 \\
\# Param. (1 Head) & 385,237 & 108,866 & 106,350 & 389,717 & 351,926 \\
\# Param. (Share Attn) & 311,381 & 92,330 & 89,814 & 315,861 &  283,670 \\
\bottomrule
\end{tabular}
\end{table}

\begin{table}[h!]
    \centering
        \caption{Hyperparameters for two datasets from the Long-range Graph Benchmark~\cite{dwivedi2022LongRangeGraph}}
    \label{tab:lrgb_hparam}
    \begin{tabular}{lcc}
\midrule Hyperparameter & Peptides-func & Peptides-struct  \\
\midrule \# Transformer Layers & 4 & 4   \\
Hidden dim &  96 & 96   \\
\# Heads & 4 & 8   \\
Dropout & 0 & 0 \\
Attention dropout &  0.2 & 0.2  \\
Graph pooling &  mean & mean  \\
\midrule
PE dim (walk-step) & 17 & 24   \\
PE encoder & linear & linear  \\
\midrule
Batch size & 32 & 32  \\
Learning Rate &  0.0003 & 0.0003 \\
\# Epochs & 200 & 200  \\
\# Warmup epochs &  5 & 5  \\
Weight decay &  0 & 0  \\
\midrule
\# Parameters & 332,142 & 338,315 \\
\# Param. (1 Head)& 370,571 & 360,574 \\
\# Param. (Share Attn)& 310,091 & 304,702 \\
\bottomrule
\end{tabular}
\end{table}

\subsection{Visualization of Curvature}

\begin{figure}[h!]
    \centering
    \includegraphics[width=.6\textwidth]{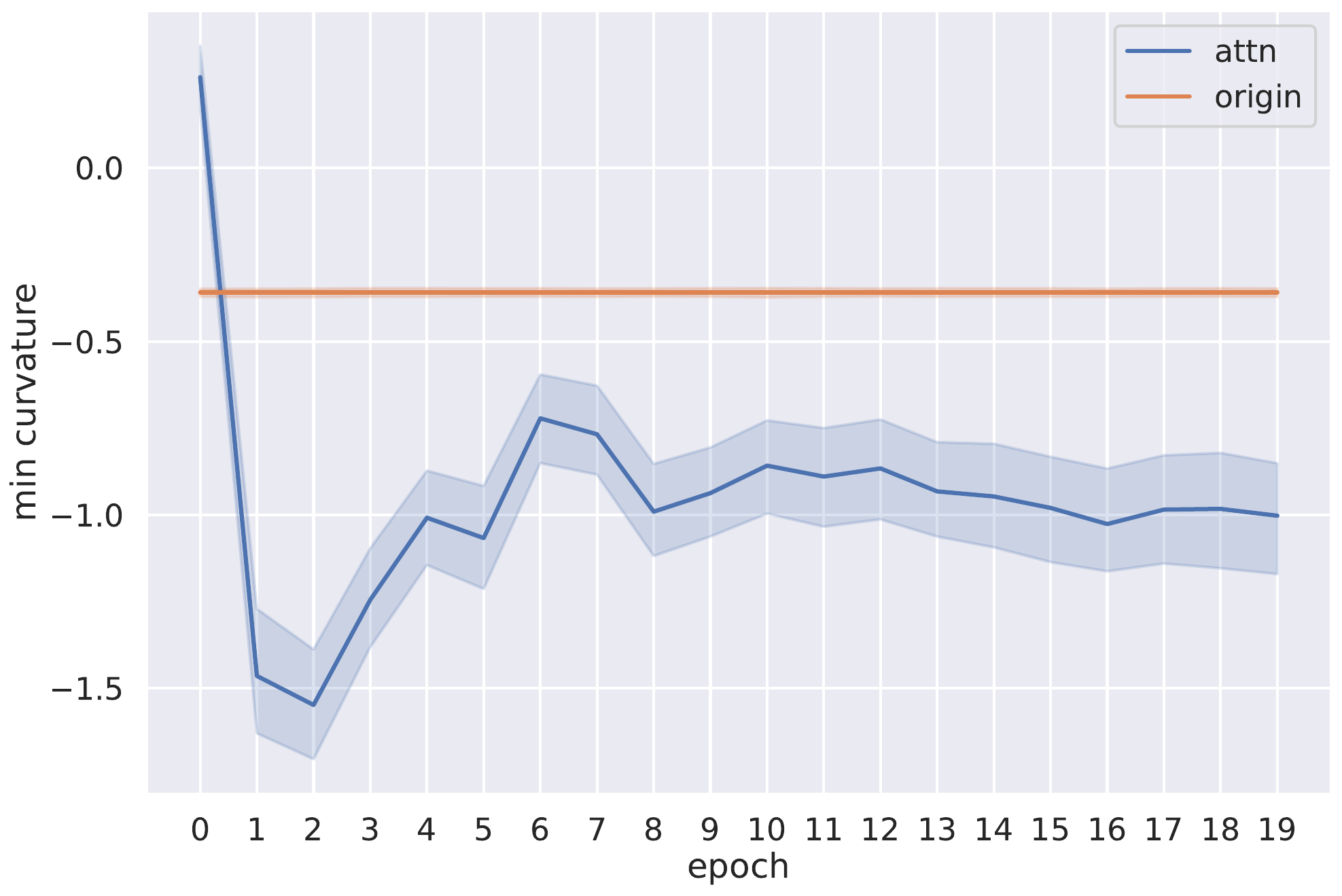}
    \caption{The trend of Minimum CURC for the first 32 test graphs in ZINC. 
    Shade is the Confidence Interval at the $95\%$ confidence level. }
    \label{fig:min_cur_trend}
\end{figure}

\begin{figure}
    \centering
    \includegraphics[width=.6\textwidth]{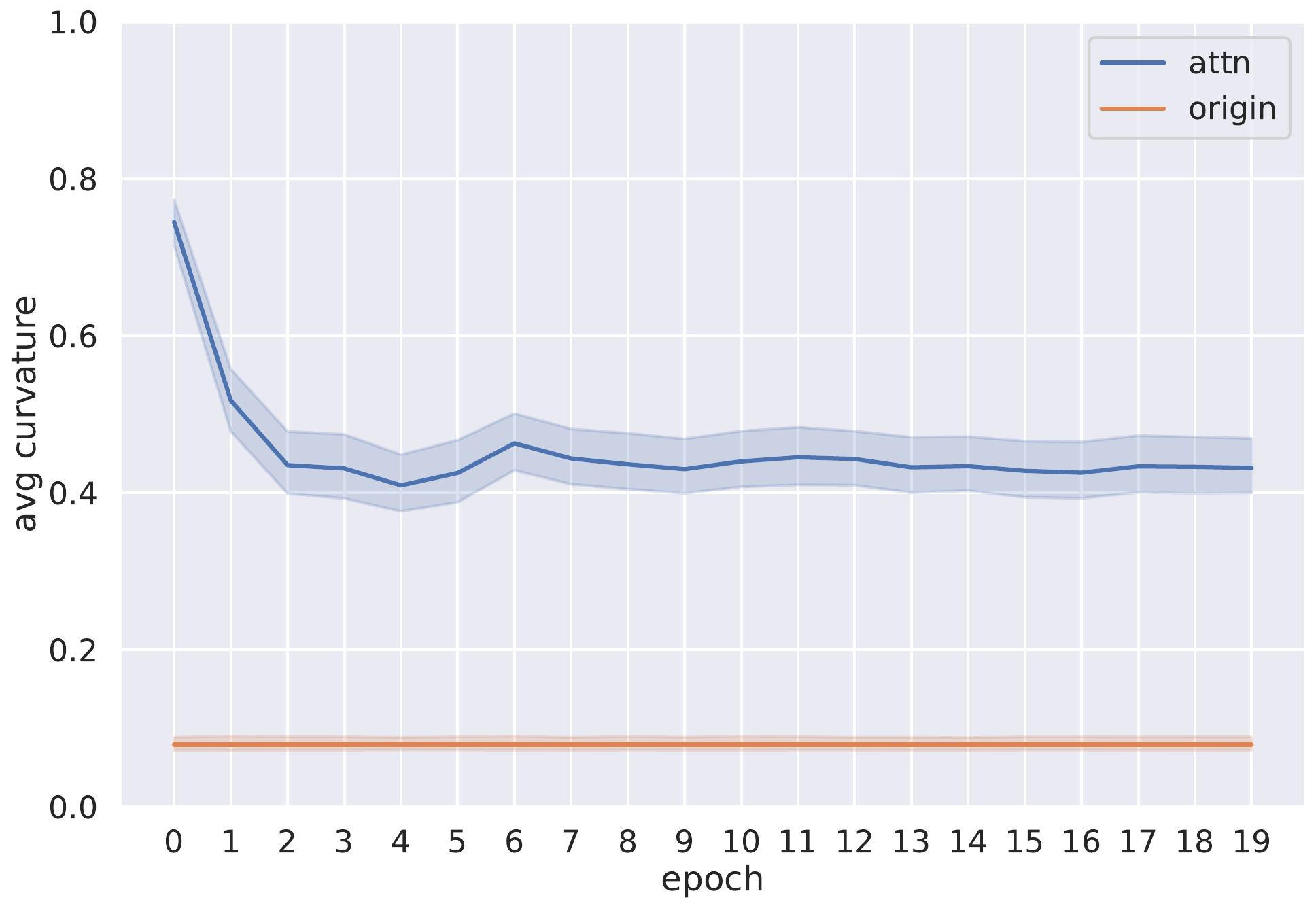}
   \caption{The trend of Average CURC for the first 32 test graphs in ZINC.
    Shade is the Confidence Interval at the $95\%$ confidence level. }

    \label{fig:avg_cur_trend}
\end{figure}

We visualize the trend curves of minimum/average CURC given the attention matrices across the first 32 test graphs in ZINC, 
as shown in Fig.~\ref{fig:min_cur_trend} and Fig.~\ref{fig:avg_cur_trend}.

% You can have as much text here as you want. The main body must be at mos  $8$ pages long.
% For the final version, one more page can be added.
% If you want, you can use an appendix like this one.  

% The $\mathtt{\backslash onecolumn}$ command above can be kept in place if you prefer a one-column appendix, or can be removed if you prefer a two-column appendix.  Apart from this possible change, the style (font s e, spacing, margins, page numbering, etc.) should be kept the same as the main body.
%%%%%%%%%%%%%%%%%%%%%%%%%%%%%%%%%%%%%%%%%%%%%%%%%%%%%%%%%%%%%%%%%%%%%%%%%%%%%%%
%%%%%%%%%%%%%%%%%%%%%%%%%%%%%%%%%%%%%%%%%%%%%%%%%%%%%%%%%%%%%%%%%%%%%%%%%%%%%%%

\end{document}